\documentclass{article}

\usepackage[letterpaper, left=1in, right=1in, top=1in, bottom=1in]{geometry} %

\usepackage[square,numbers]{natbib}
\renewcommand{\bibname}{References}
\renewcommand{\bibsection}{\subsubsection*{\bibname}}

\usepackage{graphicx}

\usepackage{amsmath,amsthm,amssymb}

\usepackage{verbatim}
\newcommand{\cmmnt}[1]{}
\usepackage{enumitem}
\usepackage{xspace} %

\newtheorem{theorem}{Theorem}

\usepackage{amsmath,amsfonts,bm}

\def\eqref#1{equation~\ref{#1}}

\def\floor#1{\lfloor #1 \rfloor}
\def\1{\bm{1}}

\DeclareMathAlphabet{\mathsfit}{\encodingdefault}{\sfdefault}{m}{sl}
\SetMathAlphabet{\mathsfit}{bold}{\encodingdefault}{\sfdefault}{bx}{n}

\DeclareMathOperator*{\argmax}{arg\,max}
\DeclareMathOperator*{\argmin}{arg\,min}

\DeclareMathOperator{\sign}{sign}

\usepackage{booktabs} 
\usepackage{multirow}
\usepackage{amsmath}
\usepackage{mathrsfs}
\usepackage{bm}
\usepackage{amssymb}

\usepackage{tensor}
\usepackage{caption}
\usepackage{subcaption}
\usepackage{algorithm}
\usepackage{algpseudocode}
\usepackage{wrapfig}

\usepackage[utf8]{inputenc} %
\usepackage[T1]{fontenc}    %
\usepackage{hyperref}       %
\usepackage{url}            %
\usepackage{booktabs}       %
\usepackage{amsfonts}       %
\usepackage{nicefrac}       %
\usepackage{microtype}      %
\usepackage{xcolor}         %

\definecolor{maroon}{cmyk}{0, 0.87, 0.68, 0.32}
\definecolor{halfgray}{gray}{0.55}
\definecolor{ipython_frame}{RGB}{207, 207, 207}
\definecolor{ipython_bg}{RGB}{247, 247, 247}
\definecolor{ipython_red}{RGB}{186, 33, 33}
\definecolor{ipython_green}{RGB}{0, 128, 0}
\definecolor{ipython_cyan}{RGB}{64, 128, 128}
\definecolor{ipython_purple}{RGB}{170, 34, 255}

\usepackage{listings}
\lstdefinelanguage{iPython}{
    morekeywords={access,and,break,class,continue,def,del,elif,else,except,exec,finally,for,from,global,if,import,in,is,lambda,not,or,pass,print,raise,return,try,while},%
    morekeywords=[2]{abs,all,any,basestring,bin,bool,bytearray,callable,chr,classmethod,cmp,compile,complex,delattr,dict,dir,divmod,enumerate,eval,execfile,file,filter,float,format,frozenset,getattr,globals,hasattr,hash,help,hex,id,input,int,isinstance,issubclass,iter,len,list,locals,long,map,max,memoryview,min,next,object,oct,open,ord,pow,property,range,raw_input,reduce,reload,repr,reversed,round,set,setattr,slice,sorted,staticmethod,str,sum,super,tuple,type,unichr,unicode,vars,xrange,zip,apply,buffer,coerce,intern},%
    sensitive=true,%
    morecomment=[l]\#,%
    morestring=[b]',%
    morestring=[b]",%
    morestring=[s]{'''}{'''},%
    morestring=[s]{"""}{"""},%
    morestring=[s]{r'}{'},%
    morestring=[s]{r"}{"},%
    morestring=[s]{r'''}{'''},%
    morestring=[s]{r"""}{"""},%
    morestring=[s]{u'}{'},%
    morestring=[s]{u"}{"},%
    morestring=[s]{u'''}{'''},%
    morestring=[s]{u"""}{"""},%
    literate=
    {á}{{\'a}}1 {é}{{\'e}}1 {í}{{\'i}}1 {ó}{{\'o}}1 {ú}{{\'u}}1
    {Á}{{\'A}}1 {É}{{\'E}}1 {Í}{{\'I}}1 {Ó}{{\'O}}1 {Ú}{{\'U}}1
    {à}{{\`a}}1 {è}{{\`e}}1 {ì}{{\`i}}1 {ò}{{\`o}}1 {ù}{{\`u}}1
    {À}{{\`A}}1 {È}{{\'E}}1 {Ì}{{\`I}}1 {Ò}{{\`O}}1 {Ù}{{\`U}}1
    {ä}{{\"a}}1 {ë}{{\"e}}1 {ï}{{\"i}}1 {ö}{{\"o}}1 {ü}{{\"u}}1
    {Ä}{{\"A}}1 {Ë}{{\"E}}1 {Ï}{{\"I}}1 {Ö}{{\"O}}1 {Ü}{{\"U}}1
    {â}{{\^a}}1 {ê}{{\^e}}1 {î}{{\^i}}1 {ô}{{\^o}}1 {û}{{\^u}}1
    {Â}{{\^A}}1 {Ê}{{\^E}}1 {Î}{{\^I}}1 {Ô}{{\^O}}1 {Û}{{\^U}}1
    {œ}{{\oe}}1 {Œ}{{\OE}}1 {æ}{{\ae}}1 {Æ}{{\AE}}1 {ß}{{\ss}}1
    {ç}{{\c c}}1 {Ç}{{\c C}}1 {ø}{{\o}}1 {å}{{\r a}}1 {Å}{{\r A}}1
    {€}{{\EUR}}1 {£}{{\pounds}}1
    {^}{{{\color{ipython_purple}\^{}}}}1
    {=}{{{\color{ipython_purple}=}}}1
    {+}{{{\color{ipython_purple}+}}}1
    {*}{{{\color{ipython_purple}$^\ast$}}}1
    {/}{{{\color{ipython_purple}/}}}1
    {+=}{{{+=}}}1
    {-=}{{{-=}}}1
    {*=}{{{$^\ast$=}}}1
    {/=}{{{/=}}}1,
    literate=
    *{-}{{{\color{ipython_purple}-}}}1
     {?}{{{\color{ipython_purple}?}}}1,
    identifierstyle=\color{black}\ttfamily,
    commentstyle=\color{ipython_cyan}\ttfamily,
    stringstyle=\color{ipython_red}\ttfamily,
    keepspaces=true,
    showspaces=false,
    showstringspaces=false,
    rulecolor=\color{ipython_frame},
    frame=single,
    frameround={t}{t}{t}{t},
    framexleftmargin=6mm,
    numbers=left,
    numberstyle=\tiny\color{halfgray},
    backgroundcolor=\color{ipython_bg},
    basicstyle=\scriptsize,
    keywordstyle=\color{ipython_green}\ttfamily,
}

\title{The Manifold Hypothesis for Gradient-Based Explanations}

\author{%
  Sebastian Bordt, Uddeshya Upadhyay, Zeynep Akata, Ulrike von Luxburg \\[4pt]
  University of Tübingen, Tübingen AI Center, Germany\\
}

\begin{document}

\maketitle

\begin{abstract}
   When do gradient-based explanation algorithms provide perceptually-aligned explanations? We propose a criterion:  the feature attributions need to be aligned with the tangent space of the data manifold. To provide evidence for this hypothesis, we introduce a framework based on variational autoencoders that allows to estimate and generate image manifolds. Through experiments across a range of different datasets 
    -- MNIST, EMNIST, CIFAR10, X-ray pneumonia and Diabetic Retinopathy detection --
   we demonstrate that the more a feature attribution is aligned with the tangent space of the data, the more perceptually-aligned it tends to be. We then show that the attributions provided by popular post-hoc methods such as Integrated Gradients and SmoothGrad are more strongly aligned with the data manifold than the raw gradient. Adversarial training also improves the alignment of model gradients with the data manifold. As a consequence, we suggest that explanation algorithms should actively strive to align their explanations with the data manifold.
   This is an extended version of a CVPR Workshop  \href{https://openaccess.thecvf.com/content/CVPR2023W/XAI4CV/papers/Bordt_The_Manifold_Hypothesis_for_Gradient-Based_Explanations_CVPRW_2023_paper.pdf}{paper}. Code is available on \href{https://github.com/tml-tuebingen/explanations-manifold}{github}.
\end{abstract}

\section{Introduction}
\label{sec:intro}

Post-hoc explanation algorithms for image classification often rely on the gradient with respect to the input~\cite{smilkov2017smoothgrad,sundararajan2017axiomatic,agarwal2021towards}. In many cases, however, model gradients and post-hoc explanations \cite{simonyan2014deep,10.1371/journal.pone.0130140,shrikumar2017learning,ancona2017towards, lim2021building} posses little visual structure that can be interpreted by humans \cite{kaur2019perceptually}. This makes image classification with neural networks one of the most challenging applications of explainable machine learning. In addition, it has demonstrated that popular post-hoc explanation algorithms fail various sanity checks \citep{adebayo2018sanity, kindermans2019reliability, arun2020assessing}. 

Recently, a number of different papers have observed conditions that lead to {\it perceptually aligned gradients} (PAGs) \cite{ganzperceptually}. In particular, it has been shown that adversarial training, as well as other forms of robust training, lead to PAGs \cite{tsipras2019robustness,kaur2019perceptually,kim2019bridging,shah2021input}. However, it remains unclear what exactly makes a feature attribution perceptually-aligned.

In this work, we try to understand when a feature attribution is perceptually-aligned (that is, aligned with human perception). We propose and investigate the following hypothesis:
\begin{center}
    \hspace{-1.2em}\fbox{\begin{minipage}{45em}
{\bf Hypothesis:} Feature attributions are more perceptually-aligned the more they are aligned with the tangent space of the image manifold.
\end{minipage}}\\[1pt]
\end{center}
{\bf To understand the motivation behind the hypothesis}, note that it is widely believed that natural image data concentrates around a low-dimensional image manifold \citep[Section 5.11.3]{Goodfellow-et-al-2016}. This image manifold captures the geometric structure of the data. In particular, the tangent space of an image captures all components of the image that can be slightly changed while still staying within the realm of natural images. If an attribution approximately lies in this tangent space, this means that it highlights structurally meaningful components of the image that contribute to the prediction. If an attribution lies orthogonal to the tangent space, this means that it points in some direction that would not lead to realistic images, and a human would have a hard time understanding its meaning. Random noise, in particular, lies with high probability orthogonal to the image manifold.  

\begin{figure}
    \centering
    \includegraphics[width=0.99\linewidth]{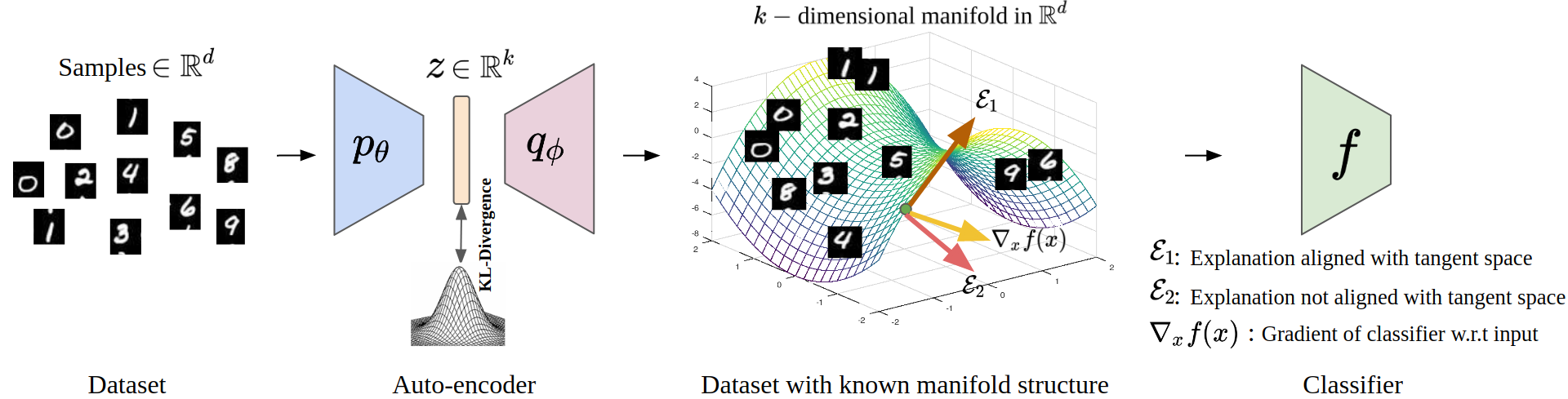}
    \caption{Conceptual overview of our approach. We first estimate the data manifold of an existing dataset with a variational autoencoder, then use the decoder as a generative model. On the generated data, we train a classifier $f$. For this classifier, we evaluate whether different gradient based explanations $\mathcal{E}_i$ align with the tangent space of the data manifold. Moving along an explanation aligned with the tangent space keeps us in the manifold, whereas moving along an orthogonal explanation takes us out of manifold. Our hypothesis is that the latter does not lead to perceptually-aligned explanations because it describes changes that lead to unnatural images.}
    \label{fig:concept}
\end{figure}

{\bf To provide empirical evidence for the hypothesis}, we employ autoencoders and estimate the image manifolds of five different datasets:  MNIST, EMNIST, CIFAR10, X-ray pneumonia, and diabetic retinopathy detection. By projecting different feature attributions into the tangent space, we then provide qualitative evidence that tangent-space components are perceptually-aligned, whereas orthogonal components visually resemble random noise (Sec. \ref{sec:results_projection}). As depicted in Figure \ref{fig:concept}, we also use variational autoencoders as generative models. This allows us to generate image datasets with {\it completely known manifold structure}. 

We then show that popular post-hoc methods such as SmoothGrad, Integrated Gradients and Input $\times$ Gradient improve the alignment of attributions with the data manifold (Sec \ref{sec:post-hoc-alignment}). The same is true for $l_2$-adversarial training, which significantly aligns model gradients with the data manifold (Sec. \ref{sec:adversarial}). These results hold consistently across all the different datasets. We also show that some form of adjustment to the model architecture or training algorithm is necessary: generalization of neural networks alone does not imply the alignment of model gradients with the data manifold (Sec. \ref{sec:theory}).

A user study demonstrates that humans perceive feature attributions as better aligned with perception if they are better aligned with the tangent space (Sec. \ref{sec:user_study}). Finally, we relate our measure of alignment with the data manifold to other measures such as the remove-and-retrain benchmark \citep{hooker2019benchmark} (Sec. \ref{sec:roar}) and sanity checks for explanations \citep{adebayo2018sanity} (Sec. \ref{sec:model_and_data}).

{\bf Apart from the intuitive and empirical plausibility of the hypothesis}, its main appeal is that it provides a clear perspective on why explaining image classifiers is difficult: While our empirical investigations show that post-hoc methods and adversarial training improve the alignment of attributions with the data manifold, in many cases there remains much room for improvement. Overall, the proposed manifold hypothesis is an important step toward understanding when feature attributions are explanations.

\section{Related Work}
\label{sec:related}

{\bf Explanation algorithms.} Many different approaches aim to explain the predictions of deep neural networks \citep{Singla2020Explanation}. Some are based on the gradient with respect to the input \citep{simonyan2014deep,smilkov2017smoothgrad,sundararajan2017axiomatic,agarwal2021towards}. Others explain the prediction in terms of the activations of intermediate layers \citep{selvaraju2017grad,leino2018influence,dhamdhere2018important} or via modifications to the backpropagation algorithm \citep{springenberg2014striving}. Other approaches are related to feature perturbation, concepts learned by the network, function approximation, counterfactuals, causality and generative modeling \cite{covert2021explaining,kim2018interpretability,ribeiro2016should,pawelczyk2022exploring,schwab2019cxplain,yeh2020completeness,koh2020concept,chang2018explaining}. This is already a very extensive literature and we do not aim to give a comprehensive overview. A number of recent works have begun to highlight connections between different explanation algorithms \cite{covert2021explaining,agarwal2021towards} and subjected them to theoretical analysis \cite{garreau2021limeforimages}.

{\bf Projections on the image manifold.} The long-standing hypothesis that natural image data concentrates around a low-dimensional image manifold is supported by a number of empirical studies \citep{weinberger2006unsupervised, fefferman2016testing}. However, the exact properties of these manifolds remain unknown \citep{aamari2019nonasymptotic}. Shao et al. \citep{shao2018riemannian} investigate the properties of manifolds generated by deep generative models and find that they have mostly low curvature.

Many different papers employ techniques where data points or model gradients are being projected on the data manifold \cite{stutz2019disentangling,dombrowski2021diffeomorphic}. In explainable machine learning, it has been shown that explanations can be manipulated by modifying the model outside of the image manifold, and that one can defend against such attacks by projecting the explanations back on the manifold \cite{dombrowski2019explanations}. 

{\bf Evaluating explanations.} The unavailability of ground-truth explanations and fact that explanations may be susceptible to adversarial attacks ~\citep{heo2019fooling,dombrowski2019explanations} makes it difficult to evaluate them ~\citep{samek2016evaluating,samek2021explaining}. A recent literature on {\it sanity checks} has shown that these principal difficulties non-withstanding, many explanations fail even the most basic tests such as parameter randomization \citep{adebayo2018sanity,adebayo2020debugging,kindermans2019reliability, arun2020assessing}. Another approach to assess attributions is to evaluate whether they are able to highlight discriminatory features \cite{hooker2019benchmark,shah2021input}. In applications, it is important to assess the effect of explanations on different human decision makers \citep{narayanan2018humans,miller2019explanation}. 

{\bf Alignment of the implicit density model with the ground truth class-conditional density model.} Srinivas et al. \citep{srinivas2020rethinking} have proposed that gradient-based explanations are more interpretable the more the density model that is implicit in the classifier $f$ is aligned with the ground truth class-conditional density model. While this criterion is much more explicit than the manifold hypothesis (it specifies what explanations should be) and also broader since it applies whether or not the data lies on a manifold, it is closely connection to the manifold hypothesis. If the data concentrates uniformly around a low-dimensional manifold, then alignment of the implicit density model with the ground truth class-conditional density model implies that the model gradient is aligned with the tangent space of the data manifold. We formally prove and discuss this connection in appendix \ref{apx:connetion_srinivas}.

\section{Overview of our approach: Measuring alignment with the image manifold}
\label{sec:framework}

We want to evaluate the following hypothesis: A gradient-based feature attribution $E\in\mathbb{R}^d$ at a point $x\in\mathcal{M}$ is more perceptually-aligned the more it is aligned with the tangent space of the image manifold at $x$. In order to do this, we have to measure the alignment of attributions with the tangent space of the image manifold. Below, we first give a background on manifolds, tangent spaces, and explanation algorithms; then, we detail our evaluation approach.

\subsection{Background}
\label{sec:model_gradients}

\paragraph{Data manifolds and tangent spaces.}
A $k$-dimensional differentiable manifold $\mathcal{M}\subset\mathbb{R}^d$ is a subset of a $d$-dimensional space that locally resembles $\mathbb{R}^k$. At every point $x\in\mathcal{M}$, the tangent space $\mathcal{T}_x$ is a $k$-dimensional subspace of $\mathbb{R}^d$. The tangent space $\mathcal{T}_x$ consists of all directions $v$ such that $x+v$, for $\left\lVert v\right\rVert$ small, is again close to the manifold. Manifolds and tangent spaces are the subject of differential geometry.

\paragraph{Model gradients and explanation algorithms.} 

We consider DNNs that learn  differentiable functions $f:\mathbb{R}^d\to\mathbb{R}^C$. Here $C$ is the number of classes and the model prediction is given by $\argmax_{i} f(x)_i$. The gradient of class $i$ at point $x$ with respect to the input is given by $\text{grad}_{i}(x)=\frac{\partial (f(x)_{i})}{\partial x}$. Note that the gradient is considered with respect to the predicted class $i$ and before the softmax is being applied. In addition to the gradient itself  \citep{simonyan2014deep}, we consider three gradient-based feature attribution methods: Integrated Gradients \citep{sundararajan2017axiomatic}, Input $\times$ Gradient \citep{ancona2017towards}, and SmoothGrad \citep{smilkov2017smoothgrad}. All methods provide explanations as vectors in $E\in\mathbb{R}^d$. We restrict ourselves to these four methods because they are directly related to the gradient with respect to the input, which is our main object of investigation.

\subsection{How do we know the image manifold?}
\label{sec:approach}

To estimate the image manifold we make use of two related approaches. In the {\it generative approach} (appendix Algorithm 1), we first train a variational autoencoder \citep{kingma2013auto, higgins2016beta} on some existing dataset. After training, we pass the entire dataset through the autoencoder. Then we train an auxiliary classifier to reproduce the original labels from latent codes and reconstructed images. Equipped with this labeling function, we sample from the prior and use decoder and labeling function to generate a new dataset with {\it completely known manifold structure}: the tangent space at each datapoint can be computed from the decoder via backpropagation \citep{shao2018riemannian, anders2020fairwashing}. 

The main limitation of the generative approach is that we might not be able to obtain high-quality samples with reasonably small latent spaces. While there have been great advances in generative modeling, state-of-the-art models like hierarchical variational autoencoders \citep{vahdat2020nvae}  require very large latent spaces, i.e. $k\approx d$. For our analysis it is however critical that $\sqrt{k/d}$ is small -- with $k=d$, the fraction of even a random vector in tangent space is always 1 (see discussion below). To evaluate our hypothesis on real-world high-dimensional image data where it is difficult to obtain realistic samples with not-too-large latent spaces, we have to rely on {\it estimating} the tangent space. In this {\it reconstructive approach}, we simply pass the original dataset through an autoencoder and take the reconstucted images with the original labels as our new dataset.

\subsection{Measuring alignment with the image manifold}
\label{sec:measurement}

To measure how well an explanation $E\in\mathbb{R}^n$ is aligned with the data manifold, we first project it into the tangent space -- denoted by $\texttt{proj}_{\mathcal{T}_x}\,E$ -- and then compute the
\begin{equation} \label{eq:fraction_in_tangent_space}
    \text{Fraction of the Explanation in Tangent Space} =\left\lVert \texttt{proj}_{\mathcal{T}_x}\,E\right\rVert_2 \big/  \left\lVert E\right\rVert_2\,\in\, [0,1].
\end{equation}
The projection into the tangent space uniquely decomposes an attribution into a part that lies in the tangent space and a part that is orthogonal to it. If the attribution completely lies in tangent space, we have $\texttt{proj}_{\mathcal{T}_x}\,E=E$ and our measure is 1. If the attribution is completely orthogonal to the tangent space, we have $\texttt{proj}_{\mathcal{T}_x}\,E=0$ and our measure is 0. 

When we quantitatively evaluate (\ref{eq:fraction_in_tangent_space}), we need to account for the fact that even a random vector has a non-zero fraction in tangent space. A random vector is by definition completely unrelated to the structure of the data manifold. The expected fraction of a random vector that lies in any $k$-dimensional subspace is approximately $\sqrt{k/d}$. In our MNIST32 task, for example, $d=1024$, $k=10$ and $\sqrt{10/1024}\approx 0.1$. Thus, we could only say that an explanation is systematically related to the data manifold if, on average, its fraction in tangent space is significantly larger than $0.1$.

\section{Experimental Results}
\label{sec:results}

\subsection{Experimental Setup}
\label{sec:setup}

Given a dataset obtained with the generative or reconstructive approach, we first train a neural network to minimize the test error. To the resulting prediction function, we then apply explanation algorithms and evaluate how the feature attributions relate to the data manifold. The core idea is to show that (1) the part of an attribution that lies in tangent space is perceptually-aligned, whereas the part that lies orthogonal to the tangent space is not; (2) among different feature attributions for the same image, attributions that have a larger fraction in tangent space are more perceptually-aligned.

\paragraph{Datasets.} 

We evaluate the hypothesis on six datasets. This includes (i)~MNIST32 and (ii)~MNIST256, two variants of the MNIST dataset \citep{lecun1998gradient} with 10 classes, 60000 grayscale training images and 10000 grayscale test images of size $32\times32$ and $256\times256$, respectively. The MNIST32 dataset was obtained from MNIST with the generative approach, using a $\beta$-TCVAE \citep{chen2018isolating}. It lies on a completely known 10-dimensional image manifold in a 1024-dimensional space. The (iii)~EMNIST128 dataset is a variant of the EMNIST dataset \citep{cohen2017emnist} that extends MNIST with handwritten letters and has over 60 classes. EMNIST128 and MNIST256 serves as examples of high-dimensional problems. The (iv)~CIFAR10 dataset was created from CIFAR10 \citep{krizhevsky2009learning} with the reconstructive approach, using a convolutional autoencoder with a latend dimension of $k=144$. We also evaluate the hypothesis on two high dimensional medical imaging datasets: (v)~X-ray Pneumonia~\citep{KERMANY20181122} and (vi)~Diabetic Retinopathy Detection (\href{https://www.kaggle.com/c/diabetic-retinopathy-detection}{https://www.kaggle.com/c/diabetic-retinopathy-detection}). These two datasets have been used before to study the properties of post-hoc explanation methods \citep{rajaraman2019visualizing,app10082908,AMYAR2020104037,arun2020assessing, chetoui2020explainable, van2020systematic}. Details on the creation of all datasets and the trained models are in appendix \ref{apx:experiments}. 

\begin{figure}[!t]
    \centering
    \includegraphics[width=\textwidth]{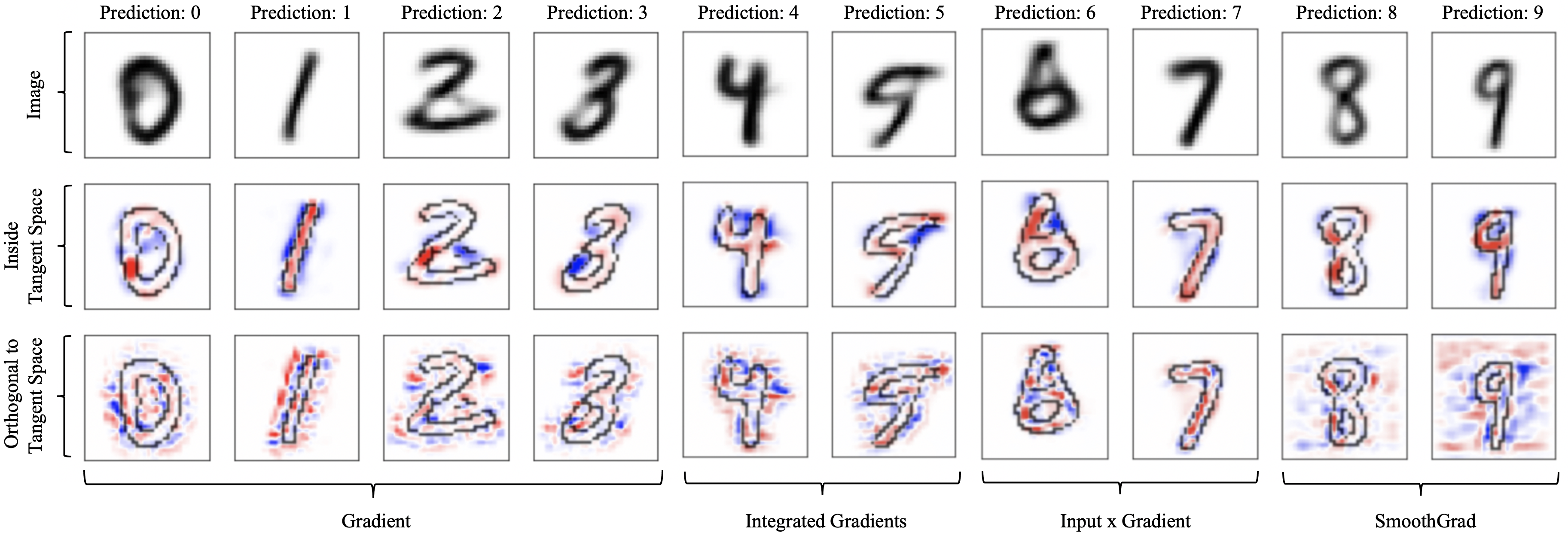}
    \caption{The part of an attribution that lies in the tangent space is perceptually-aligned, whereas the part that is orthogonal to the tangent space is not. (First row) Images from the test set of MNIST32. (Second row) The part of the attribution that lies in tangent space. (Third row) The part of attribution that is orthogonal to the tangent space. Red corresponds to positive, blue to negative attribution (best viewed in digital format). Additional attributions for more images are depicted in appendix Figure \ref{fig:apx_mnist_32_additional_attributions}.}
    \label{fig:mnist_gradients}
\end{figure}

\subsection{The part of an attribution that lies in tangent space is perceptually-aligned}
\label{sec:results_projection}

We first demonstrate on MNIST32 that the part of an attribution that lies in tangent space is perceptually-aligned (and, often, also related to the class of the image) whereas the part of the attribution that is orthogonal to the tangent space is not. Figure \ref{fig:mnist_gradients} depicts the gradient, Integrated Gradients, Input $\times$ Gradient and SmoothGrad attributions for a neural network with two convolutional and two fully connected layers that achieves a test accuracy $>99\%$. In the figure, the attributions are decomposed into the part that lies in tangent space (second row) and the part that is orthogonal to the tangent space (third row). It is quite clear that the parts that lie in the tangent space are perceptually-aligned, whereas the parts that are orthogonal to it are not. In fact, the parts of the attributions that are orthogonal to the tangent space consist of seemingly unrelated spots of positive and negative attribution. For most images, the part that lies in the tangent space is also explanatory in the sense that it highlights regions that are  plausibly important for the classification decision. For example, in case of the number 3 (fourth column of Figure \ref{fig:mnist_gradients}), regions that would complete an 8 have negative attribution.  Note that while the part of an attribution that lies in tangent space will always be perceptually-aligned, it will not necessarily be explanatory (many directions in the tangent space might not correspond to regions that are salient for the classification decision - compare the examples of random attributions in appendix Figure \ref{fig:apx_mnist_32_random}). Empirically, however, we find that attributions with a sufficiently large fraction in tangent space are often explanatory.

In conclusion, projecting attributions into the tangent space of the data manifold provides some first intuitive evidence for our manifold hypothesis.

\subsection{Post-hoc methods align attributions with the data manifold}
\label{sec:post-hoc-alignment}

\begin{figure}[!t]
    \centering
    \includegraphics[width=0.31\textwidth]{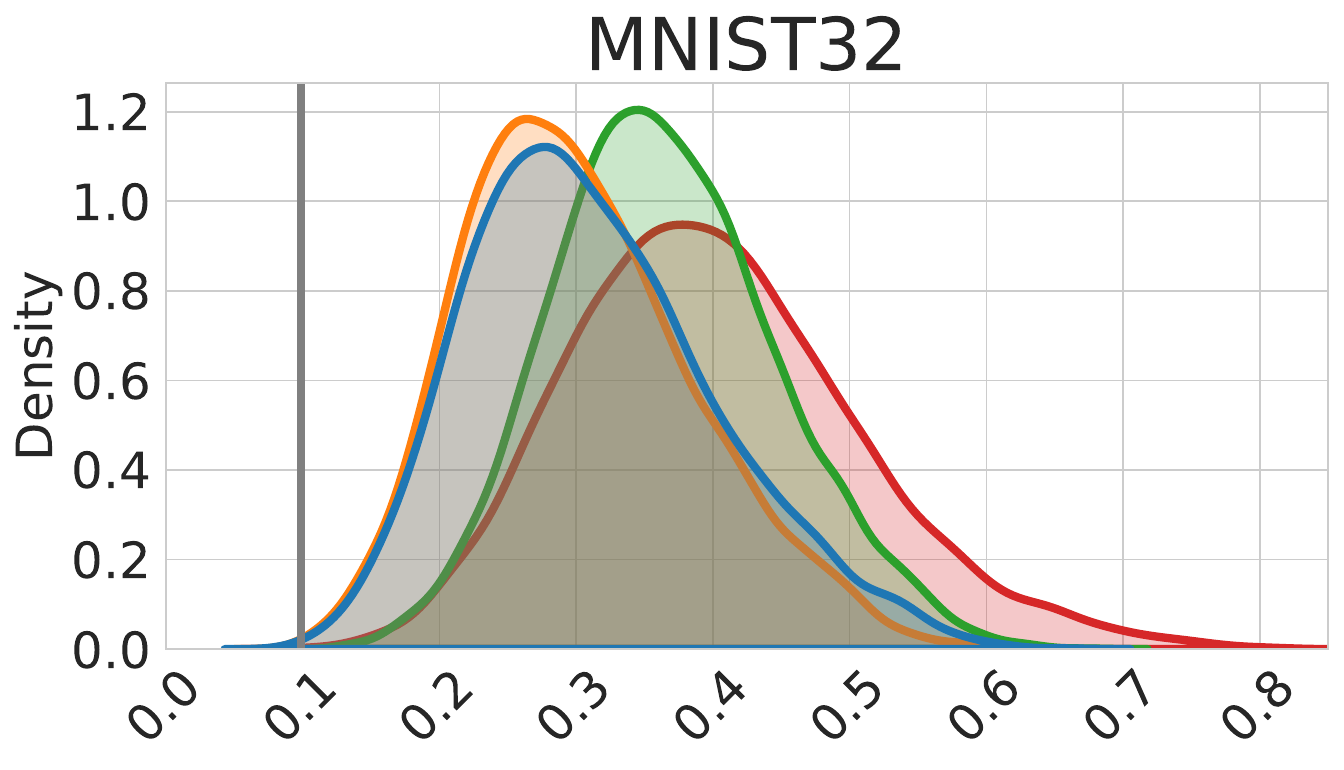}
    \hfill
    \includegraphics[width=0.31\textwidth]{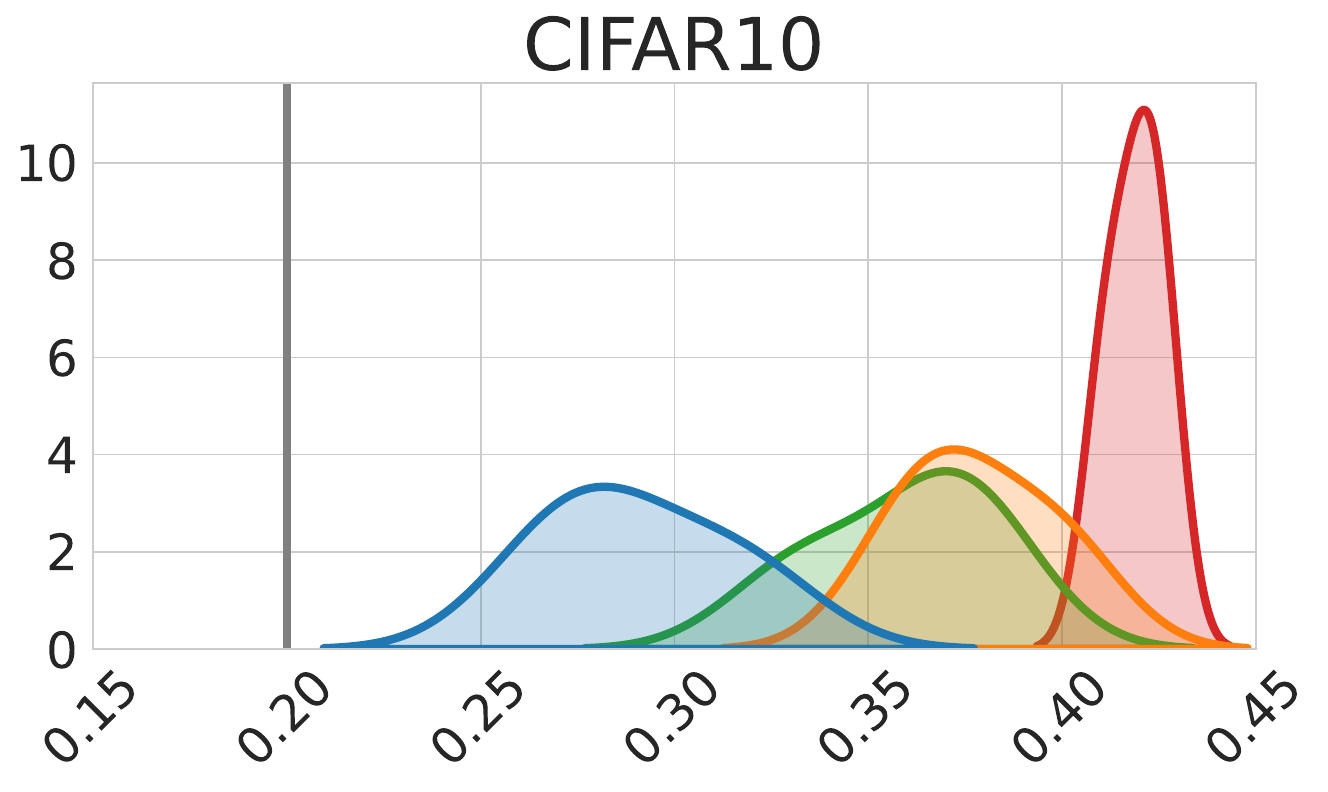}
    \hfill
    \includegraphics[width=0.29\textwidth]{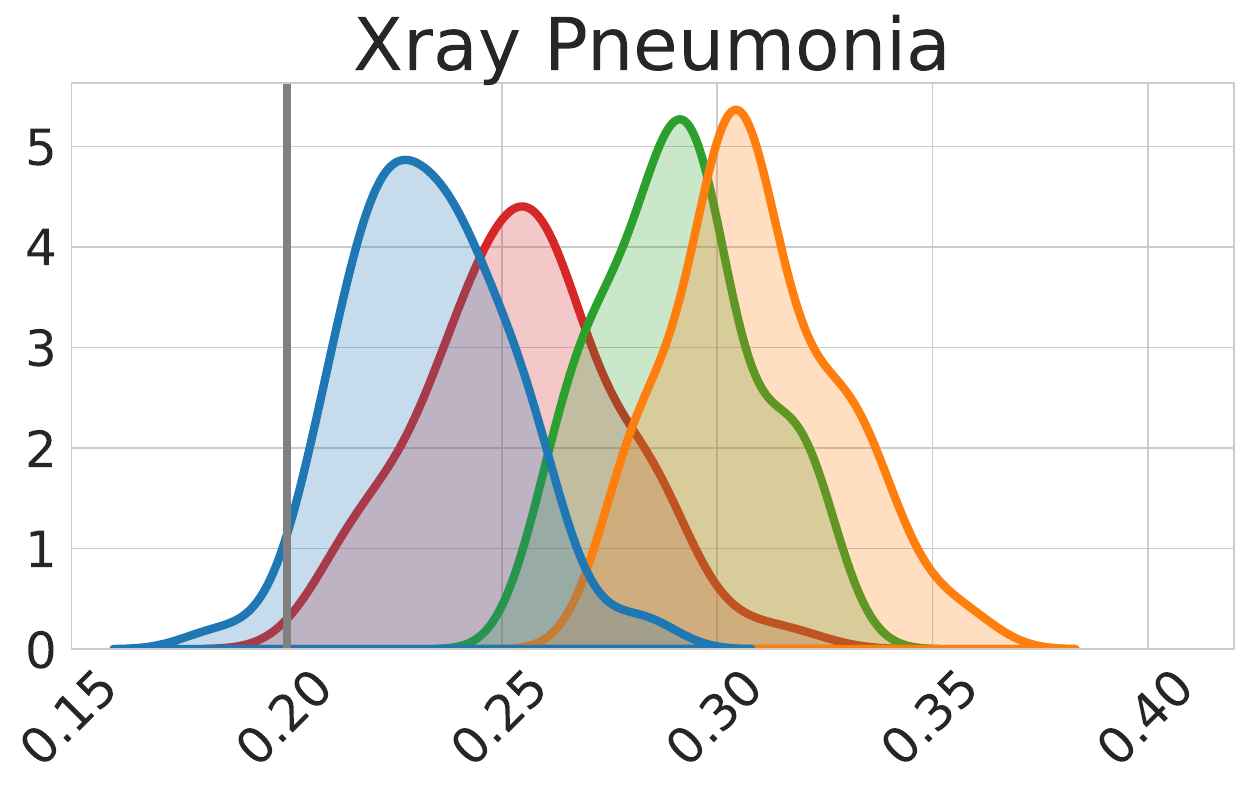}
    \includegraphics[width=0.31\textwidth]{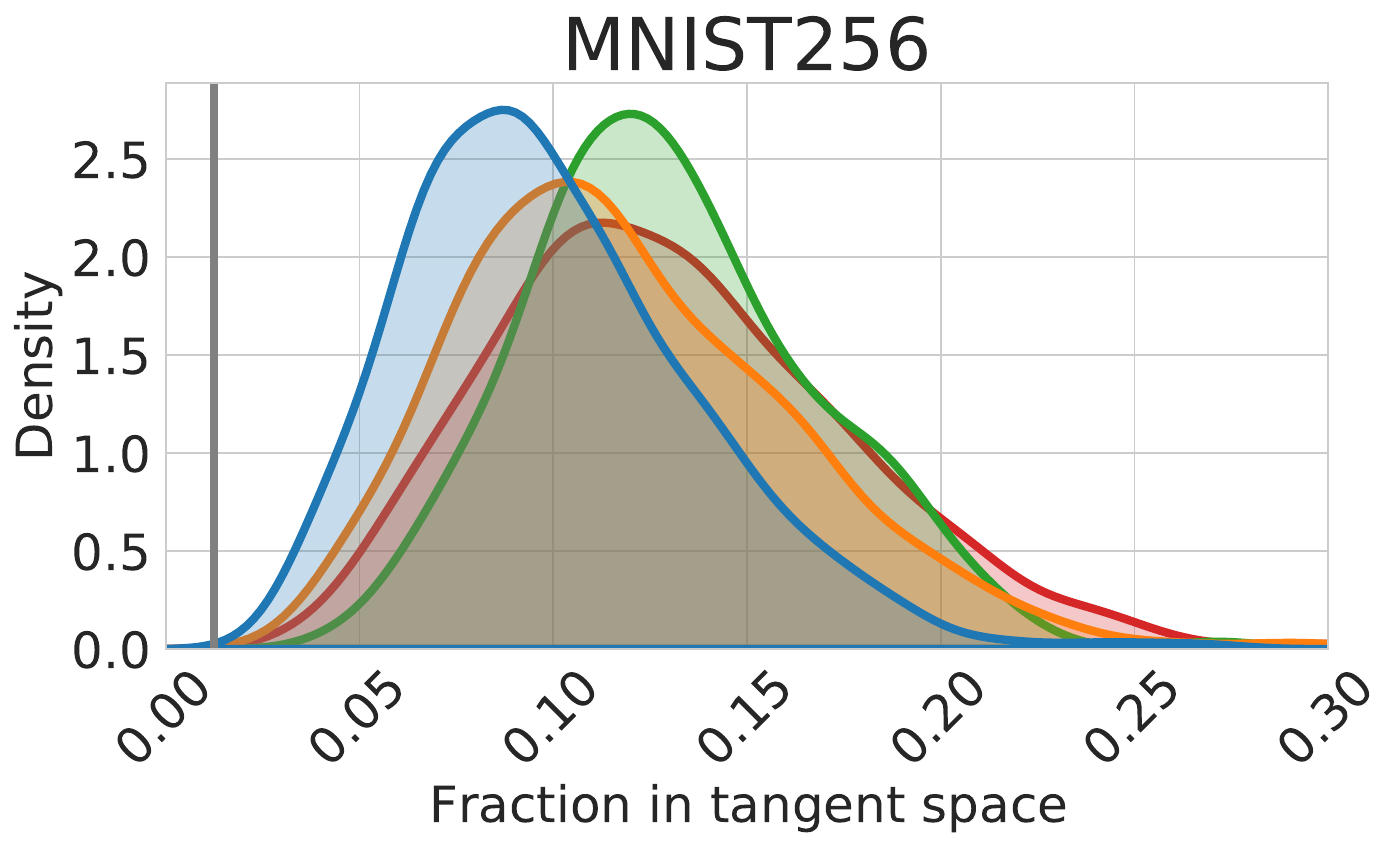}
    \hfill
    \includegraphics[width=0.31\textwidth]{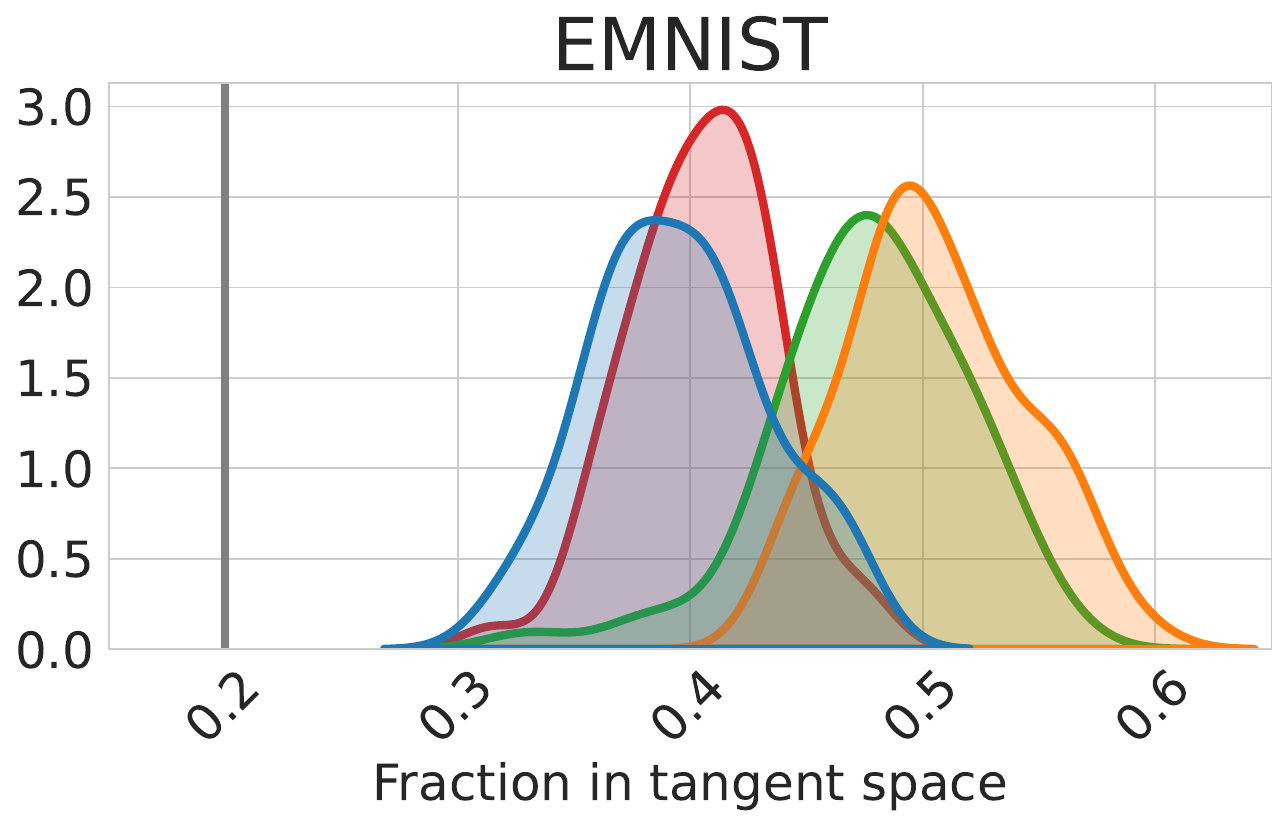}
    \hfill
    \includegraphics[width=0.31\textwidth]{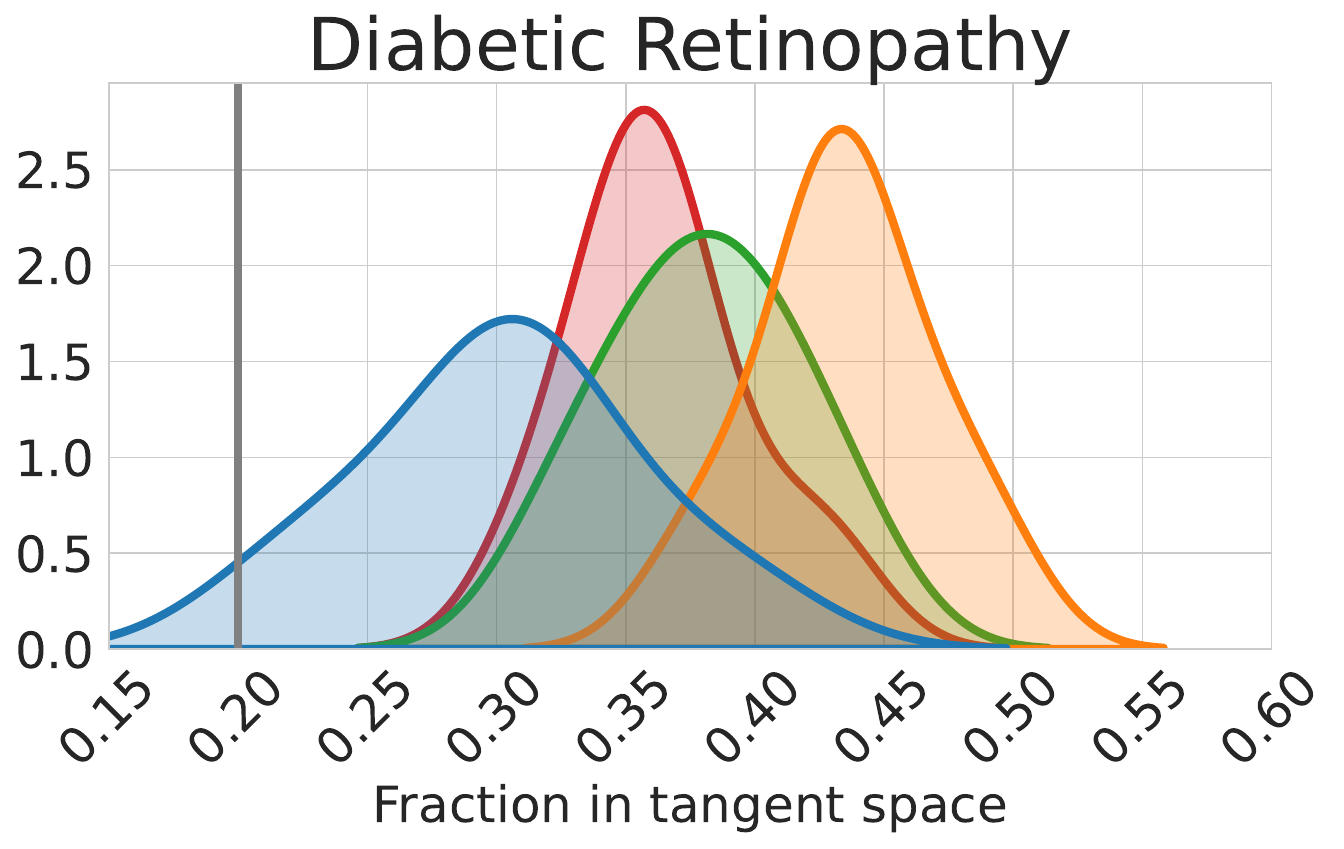}\\
    \vspace{3pt}
    \includegraphics[width=0.8\textwidth]{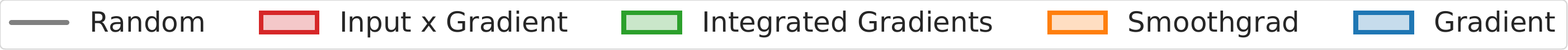}
    \vspace{-1pt}
    \caption{Post-hoc explanation methods improve the alignment of model gradients with the data manifold. Figure shows the fraction of four different explanation methods in tangent space on six different datasets. Gray line indicates the random baseline $\sqrt{k/d}$ (compare Sec. \ref{sec:measurement}).}
    \label{fig:methods_fractions}
\end{figure}

We now demonstrate that the attributions provided by post-hoc methods are more aligned with the tangent space than the gradient. Figure \ref{fig:methods_fractions} depicts the fraction in tangent space (\ref{eq:fraction_in_tangent_space}) of model gradients, SmoothGrad, Integrated Gradients and Input $\times$ Gradient on six different datasets. All attributions have a fraction in tangent space that is considerably larger than random. In particular, the mean fraction of the raw gradient in tangent space is significantly larger than random on all datastes. However, even if the relation between the gradient and the data manifold is better than random, the gradient nevertheless is the method with the weakest connection to the data manifold. Integrated Gradients, Input $\times$ Gradient and SmoothGrad improve upon the gradient on every single dataset. 

While the overall improvement of post-hoc methods over the gradient is consistent across all datasets, the relative ordering of the different post-hoc methods is not. On MNIST32 and CIFAR10, Input $\times$ Gradient is most aligned with the data manifold. On EMNIST128, Pneumonia and Diabetic Retinopathy it is SmoothGrad. To verify that the relative ordering of the different explanation methods is not just a random artifact, we replicated the results for the MNIST32 dataset 10 times with different random seeds for the autoencoder, the sampling of the dataset and the training of the model. It turns out that Input $\times$ Gradient is most strongly aligned with the data manifold also across these replications (appendix Figure \ref{fig:apx_mnist_replication}). Thus, the relative ordering between the different explanation methods must be driven by other factors such as the structure of the image manifold and the dimension of the problem. As an experiment,  we artificially upsampled the MNIST32 dataset to $256\times256$ by bilinear upsampling. This preserves the structure of the data manifold while increasing the dimension of the ambient space (on the resulting MNIST256 dataset, the ratio $\sqrt{k/d}$ is only $0.012$). As can be seen in the bottom left part of Figure \ref{fig:methods_fractions}, SmoothGrad improves upon the gradient on the high-dimensional problem, unlike on the original MNIST32 dataset. This suggests that the relative ordering of SmoothGrad across the different problems is indeed related to the dimension of the problem. In addition, this experiment reduced the overall fraction in tangent space of all explanation methods. We conjecture that holding everything else fixed, aligning model gradients with the data manifold is harder as the ratio $\sqrt{k/d}$ decreases.

In conclusion, post-hoc methods consistently improve our measure of alignment with the data manifold, providing evidence of our hypothesis. In the next section, we show that the attributions of the post-hoc methods are indeed more perceptually-aligned.

\subsection{Attributions more aligned with the data manifold are more perceptually-aligned}
\label{sec:user_study}

\begin{wraptable}[15]{r}{0.55\textwidth}
\footnotesize
\vspace{-13pt}
\caption{User study. The first column is the task. The second column (N) is the number of times the task was presented to the participants. Columns three (A) and four (B) show the number of times that the participants chose an image from group (A) or (B), respectively. Columns five and six show the average fraction in tangent space for the images in group A and B, respectively. The last column is the p-value that A<B (t-test).}
\label{tab:human_experiment}
\vspace{-11pt}
\begin{tabular}{cccccccc}\\\toprule  
Task & N & A & B & $\mathcal{T}^A$ & $\mathcal{T}^B$  & p-value  \\\midrule
CIFAR 1 & 300 & 36 & 217 & 0.27 & 0.43 & < 0.01 \\  \midrule
MNIST 1 & 600 & 0 & 580 & 0 & 1 & < 0.01 \\  \midrule
MNIST 2 & 600 & 143 & 315 &  0.32 & 0.38 & < 0.01 \\  \bottomrule
\end{tabular}
\end{wraptable} 

To assess whether attributions that are more aligned with the data manifold are also more perceptually-aligned, we conducted a user study (Table \ref{tab:human_experiment}). In this study, we did not tell the participants about explanations, the manifold hypothesis or feature attribution methods. We simply asked them to compare images according to different criteria. The study consisted of three different tasks: CIFAR 1, MNIST 1 and MNIST 2. Each task took the form of an A/B-test where the participants were repeatedly shown two corresponding images from group A and group B, and asked to make a choice. A complete documentation of the user study, including screenshots, is in appendix \ref{apx:user_study}.

The results of the user study are depicted in Table \ref{tab:human_experiment}. In the CIAFR 1 task, participants ware asked whether Input $\times$ Gradient attributions (group B) better highlighted the object in the image than the gradient (group A). The participants decided that Input $\times$ Gradient attributions significantly better highlighted the object in the image than the gradient (p-value < 0.01). In the MNIST 1 task, participants decided that the components of an attribution in tangent space are more visually structured than orthogonal components (p-value < 0.01). In the MNIST2 task, participants decided that among two randomly chosen attributions for the same image, the ones with the larger fraction in tangent space are more visually structured (p-value < 0.01).

\begin{figure*}[t]
         \centering
         \includegraphics[width=0.875\textwidth]{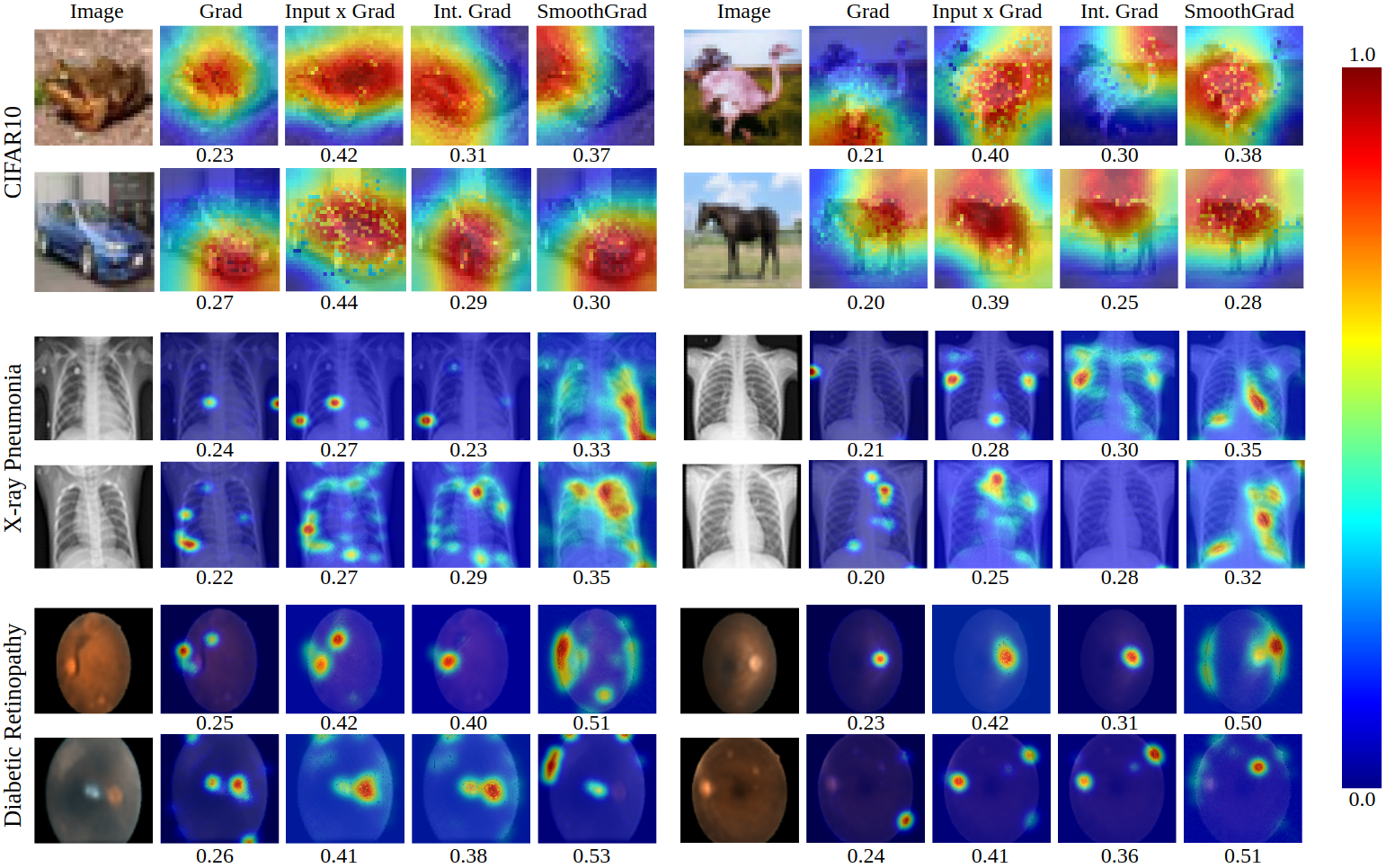}
         \caption{Feature attributions that are more aligned with the data manifold are more explanatory. (Top row) CIFAR10, (Middle row) X-Ray Pneumonia and (Bottom row) Diabetic Retinopathy. The number below an attribution depicts the fraction of the attribution in tangent space.}
         \label{fig:all_qq}
\end{figure*}

In conclusion, we find that humans perceive attributions with a larger fraction in tangent space as more perceptually-aligned, providing strong evidence for our hypothesis.

As additional qualitative evidence, Figure~\ref{fig:all_qq} depicts examples from the CIFAR10, Pneumonia and Retinopathy Detection dataset. The four CIFAR10 examples illustrate that Input $\times$ Gradient, the method most strongly aligned with the data manifold, also provides the most explanatory feature attributions. For the upper-left image of the frog, for example, Input $\times$ Gradient focuses on the central region covering the entire frog, while other methods seem to focus on only some parts of frog along with the background. For pneumonia, the qualitative examples on the left indicate that SmoothGrad focuses on the relevant region near the lungs to make the predictions, whereas raw gradients do not seem to focus on the lungs. For retinopathy, SmoothGrad seems to focus on regions away from the pupil to explain the diagnosis whereas other methods, such as the gradient, wrongly focus on the pupil. We also note that is a literature which demonstrates the utility of SmoothGrad and Integrated Gradients for pneumonia and diabetic retinopathy detection \citep{sayres2019using, van2020systematic} %

\subsection{The tangent space gives rise to a notion of feature importance}$\,$\\[12pt]
\label{sec:roar}

\begin{wrapfigure}[20]{r}{0.4\textwidth}
    \vspace{7pt}
    \includegraphics[width=0.4\textwidth]{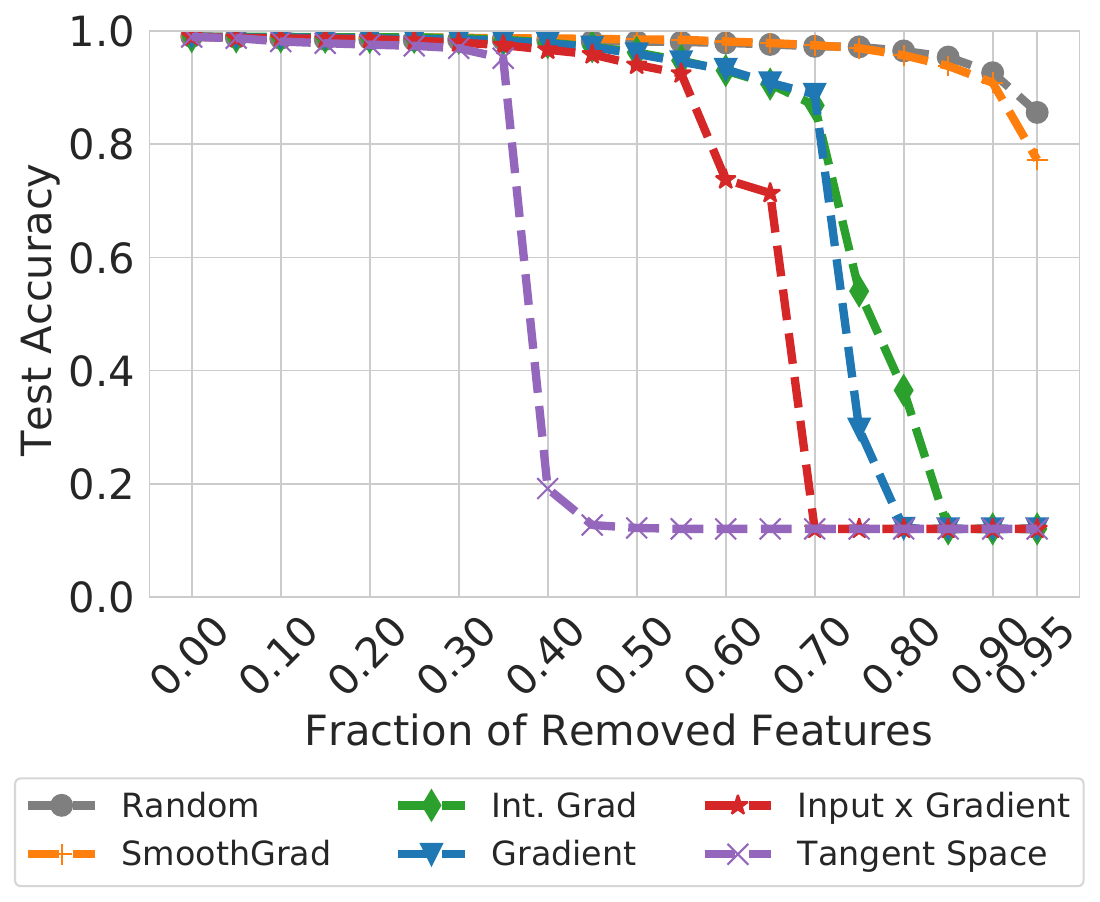}
    \vspace{-12pt}
    \caption{The tangent space gives rise to a notion of feature importance. Figure shows the ROAR benchmark on MNIST32. Additional figures for other datasets are in appendix \ref{apx:figures}.}
    \label{fig:roar}
\end{wrapfigure}

The tangent space gives rise to a notion of feature importance.\footnote{By identifying a feature with a unit vector for which we compute the fraction in tangent space.} This allows us to illustrate a connection between the tangent space and the remove-and-retrain (ROAR) benchmark \citep{hooker2019benchmark}. Figure \ref{fig:roar} shows the results of the ROAR benchmark on MNIST32. Input $\times$ Gradient, the method with the highest average fraction
in tangent space, also provides the most accurate measure of feature importance. Why is this the case? On this dataset, {\it the tangent space itself provides a highly accurate notion of feature importance} (purple curve in Figure \ref{fig:roar}). According to the ROAR metric, the tangent space even provides a more accurate notion of feature
importance than any of the explanation methods. While the tangent space provides an accurate measure of feature importance on MNIST32, it is actually simple to construct examples where the tangent space does {\it not} provide an accurate measure of feature importance. In fact, this will be the case whenever the image contains additional objects that are not relevant to the classification decision (as in the BlockMNIST task in \citep{shah2021input}). This already highlights an important point that we continue to discuss in Sec. \ref{sec:model_and_data}: Even if the manifold hypothesis holds, it cannot replace other metrics for explanations such as ROAR and sanity checks \citep{adebayo2018sanity}.

\section{Consequences of the manifold hypothesis for gradient-based explanations}
\label{sec:implications}
In the previous section we provided different kinds of evidence for our manifold hypothesis. In this section we ask: If the hypothesis were true, what would be the consequences? First, it would be desirable to train DNNs such that input gradients are as aligned with the data manifold as possible \citep{chang2018explaining}. A step in this direction is adversarial training (Sec. \ref{sec:adversarial}). However, perhaps we can also hope that the alignment of model gradients with the data manifold arises as a side effect of good generalization? Unfortunately this it not the case (Sec. \ref{sec:theory}). Finally, we ask if the alignment of feature attributions with the tangent space can also serve as sufficient criterion for explanations. The answer is no, as becomes clear by a comparison with sanity checks for explanations (Sec. \ref{sec:model_and_data}). 

\subsection{It is desirable to use adversarial training to$\,$\\align the gradient with the data manifold}
\label{sec:adversarial}

\begin{wrapfigure}[14]{r}{0.35\textwidth}
    \vspace{-65pt}
  \begin{center}
    \includegraphics[width=\linewidth]{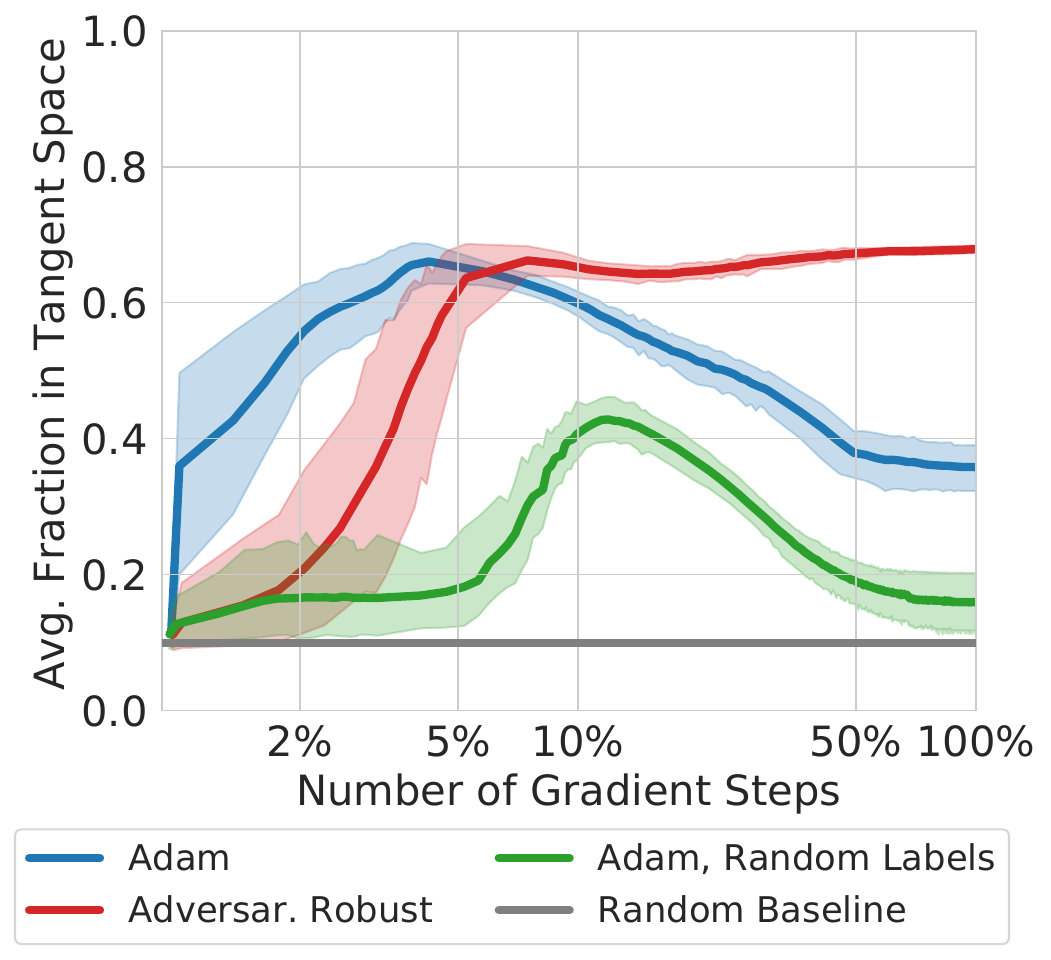}
  \end{center}
  \vspace{-12pt}
  \caption{Fraction of gradient in tangent space evolving over the course of training on MNIST32. Mean and 90\% confidence bounds.}
  \label{fig:training_evolution}
\end{wrapfigure}
\begin{figure}[t]
  \begin{center}
    \includegraphics[height=0.21\textwidth]{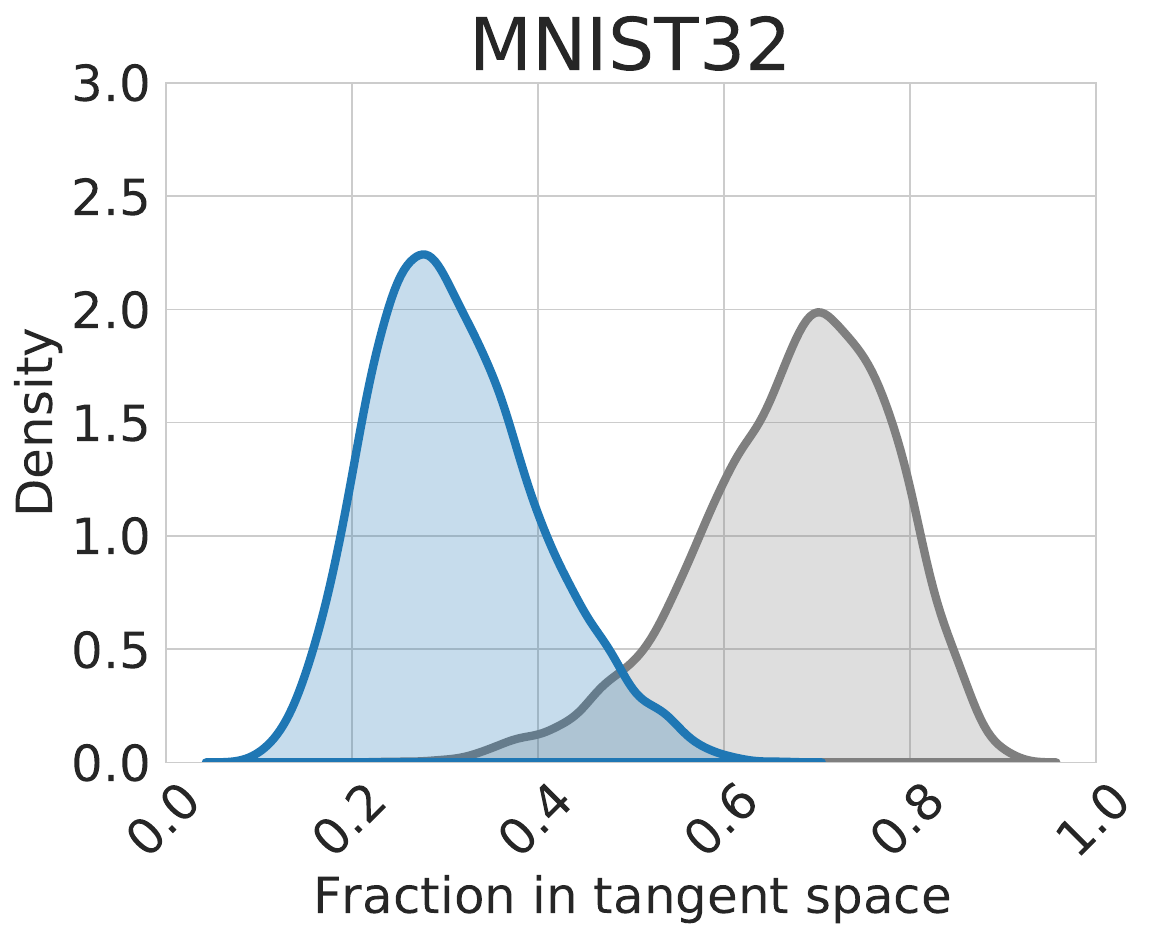}
    \hfill
    \includegraphics[height=0.21\textwidth]{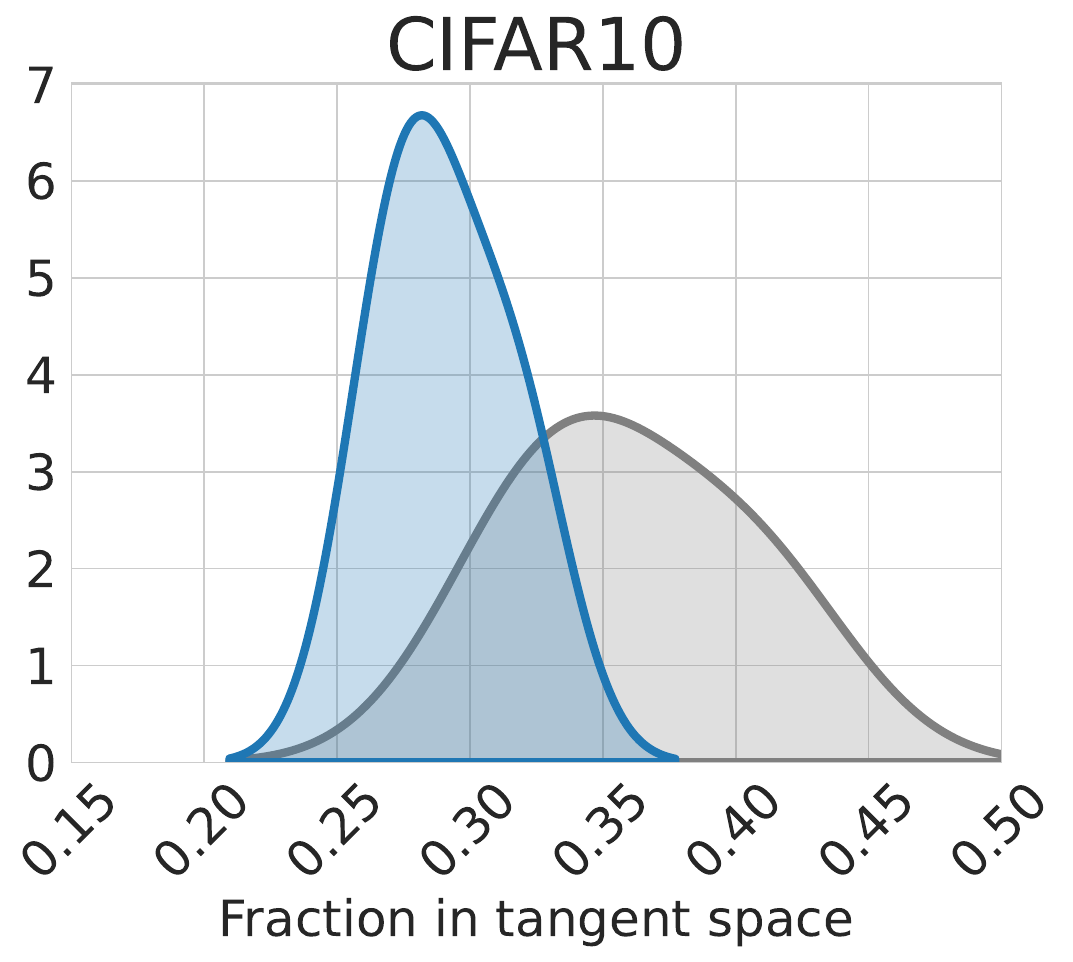}
    \hfill
    \includegraphics[height=0.21\textwidth]{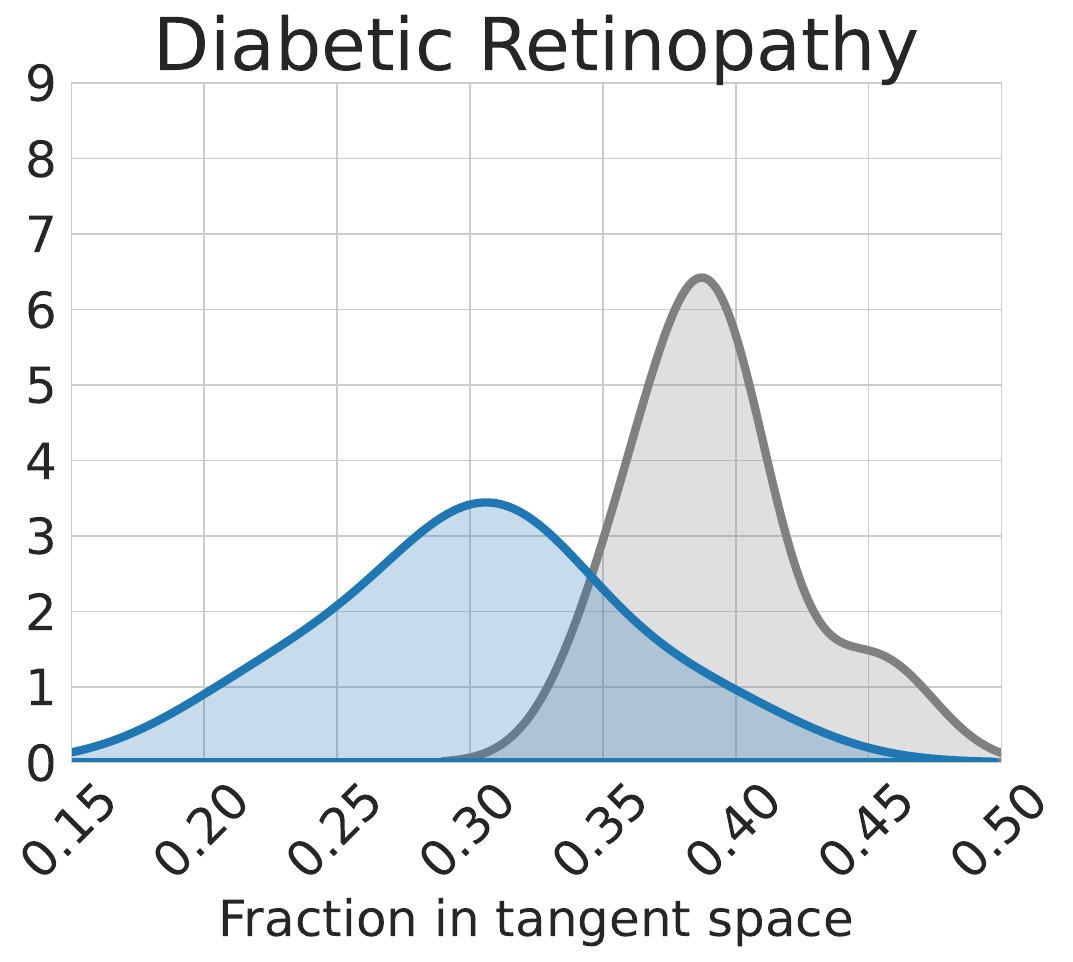}
    \hfill
    \includegraphics[height=0.21\textwidth]{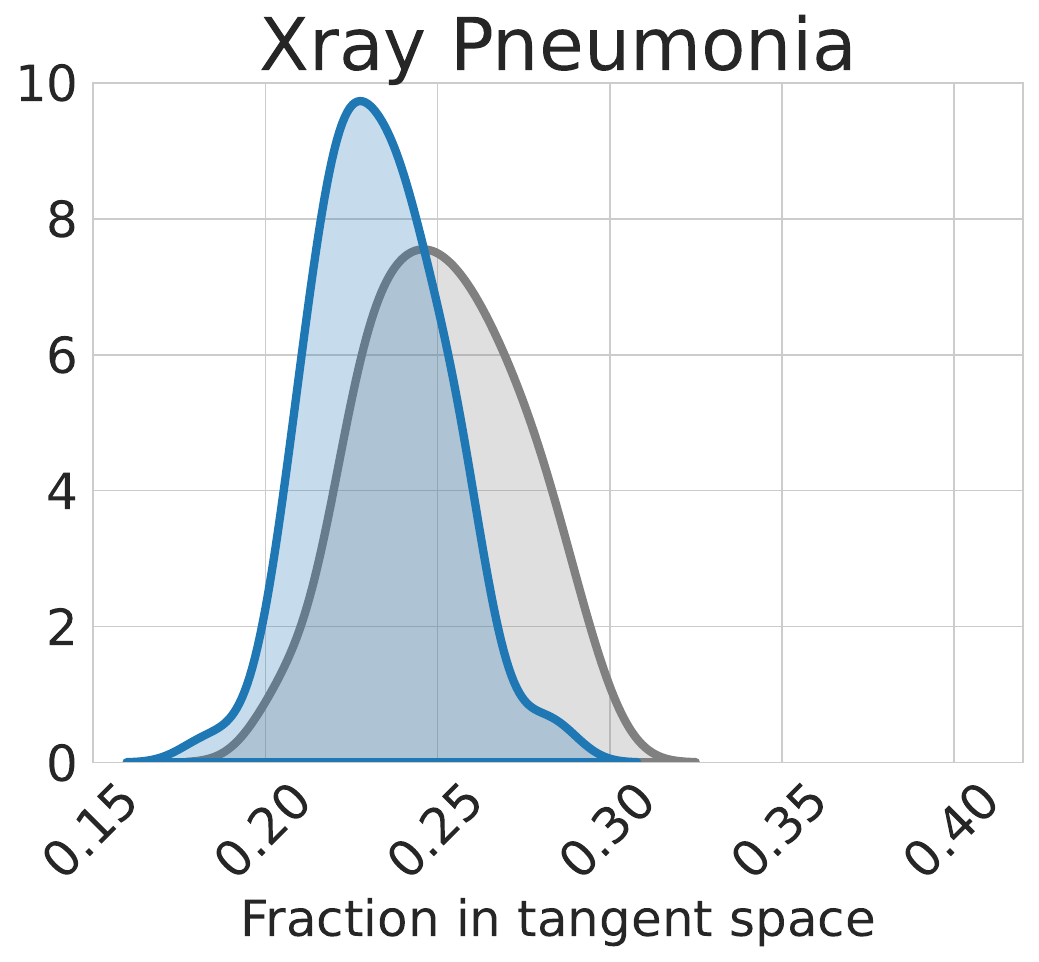}\\
    \vspace{3pt}
    \includegraphics[width=0.325\textwidth]{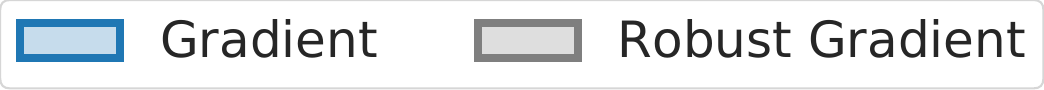}
  \end{center}
  \vspace{-5pt}
  \caption{Adversarial training improves the alignment of model gradients with the data manifold. Figure shows the fraction of standard- and $l_2$-robust gradients in tangent space across four datasets.}
  \label{fig:mnist32_robust_gradients}
 \vspace{-2pt}
\end{figure}

Previous work has shown that adversarial examples of robust networks lie closer to the image manifold, and suggested that this is because adversarial training restricts loss gradients to the image manifold \cite{stutz2019disentangling,kim2019bridging}. There is also existing evidence that the gradients of adversarially trained models provide better explanations \citep{tsipras2018robustness,kim2019bridging,shah2021input}. %
We now quantify how well adversarial training aligns model gradients with the image manifold. Figure \ref{fig:mnist32_robust_gradients} depicts the fraction of model gradients in tangent space, both for standard gradients (Sec. \ref{sec:post-hoc-alignment}), and for the robust gradients of a model trained with projected gradient descent (PGD) against an $l_2$-adversary \citep{madry2017towards}. It turns out that adversarial training significantly aligns the gradient with the data manifold. On MNIST32, the mean fraction of robust gradients in tangent space is $0.68$, compared with $0.31$ for the standard model, and $0.40$ for Input $\times$ Gradient (Figure \ref{fig:methods_fractions}). Moreover, $l_2$-adversarial training improves the alignment of model gradients with the data manifold across all tasks. Details regarding the adversarial training procedure are in Appendix \ref{apx:experiments}.

Why does adversarial training improve the relation of gradients with the data manifold? Consider Figure \ref{fig:training_evolution}, which depict the evolution of the fraction of model gradients in tangent space over the course of training. At initialization, the relation between model gradients and the tangent space of the data is as good as random. During the early steps of training with Adam \citep{kingma2014adam}, model gradients become rapidly aligned with the data manifold. However, the relation between model gradients and the data manifold deteriorates as the model increasingly fits the labels. This effect is avoided by $l_2$-adversarial training. As an additional experiment, training with random labels demonstrates that some properties of the data manifold are learned in a truly unsupervised way and not implicitly through the labels. More detailed figures, including the test error, are in appendix \ref{apx:figures}.

\subsection{Generalization it not enough since it does not align the gradient}
\label{sec:theory}

\begin{wrapfigure}[12]{r}{0.22\textwidth}
  \begin{center}
    \vspace{-13pt}
    \includegraphics[width=\linewidth]{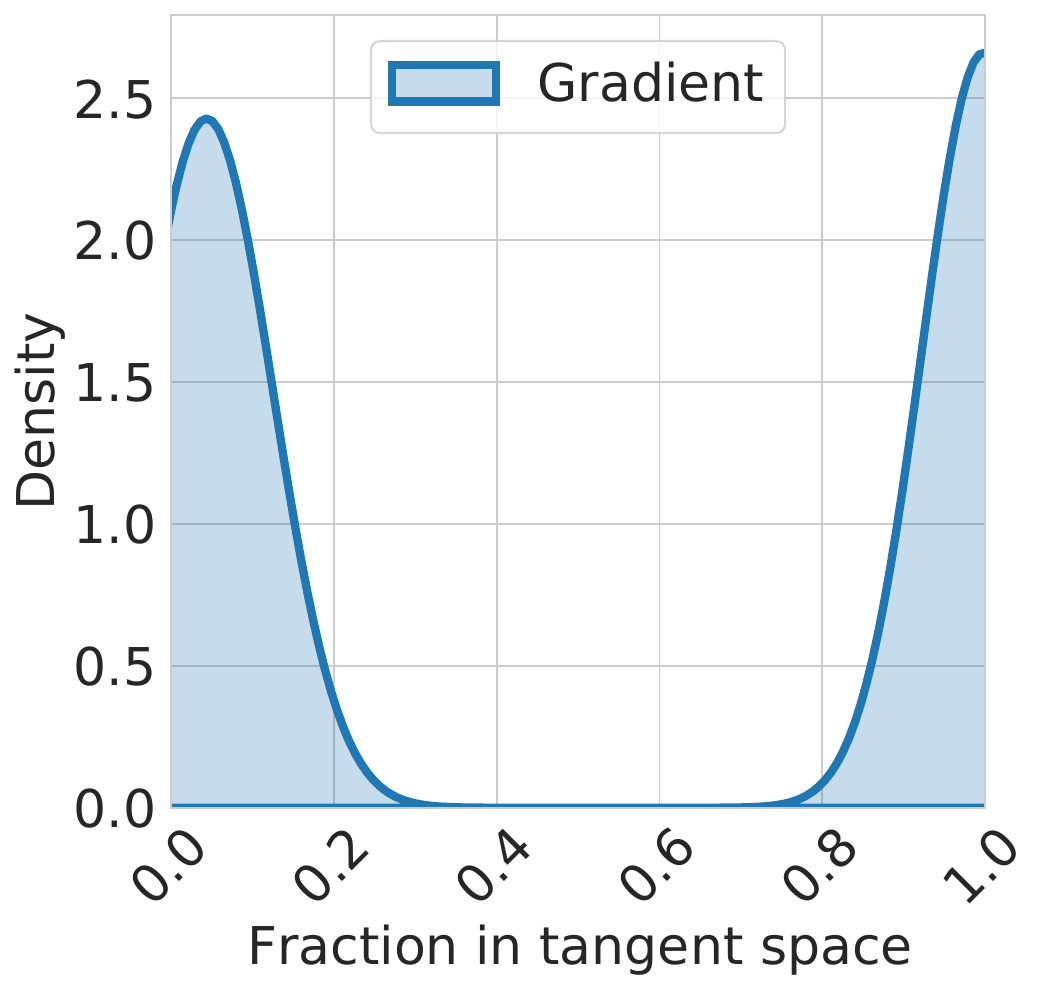}
  \end{center}
  \vspace{-8pt}
  \caption{Simulation.}
  \label{fig:theory_simulation}
\end{wrapfigure}
Can we hope that the alignment of model gradients with the data manifold arises as a side effect of generalization? Unfortunately this is not the case. In fact, a neural network that achieves a test accuracy of $100\%$ can exhibit an arbitrary amount of variation between its gradients and the data manifold (Theorem \ref{thm:1}). To see this, we construct a classification problem where (a) for 50\% of observations, model gradients lie within the tangent space of the data manifold, and (b) for the remaining 50\% of observations, model gradients are orthogonal to the tangent space of the data manifold. Figure \ref{fig:theory_simulation} depicts the simulation results for a two-layer neural network trained to solve this classification problem. To formally prove this result, we leverage the recently demonstrated connections between the training dynamics of infinite width neural networks and Wasserstein gradient flow  \citep{chizat2018global,chizat2020implicit}. The proof is in appendix \ref{apx:proof} and follows \citep{shah2021input}. 

\begin{theorem}[{\bf Generalization does not imply alignment of gradients with the data manifold}] \label{thm:1}
For every dimension $d>1$, there exists a manifold $\mathcal{M}_d\subset\mathbb{R}^d$, a probability distribution $\mathcal{D}$ on $\mathcal{M}_d\times\{-1,1\}$ and a maximum-margin classifier with zero test error given given by\begin{equation} \label{eq:infinite_wdith_max_margin_clasifier}
    \nu^\star=\argmax_{\nu\in\mathcal{P}(\mathbb{S}^{d+1})}\min_{(x,y)\in\mathcal{D}}y\cdot f(\nu,x),\quad f(\nu,x)=\mathbb{E}_{(w,a,b)\sim\nu}\,w\cdot \max(\langle\,a,x\rangle+b, 0)
\end{equation} such that\\[-4pt]\begin{equation*}
    \mathbb{P}_{(x,y)\sim\mathcal{D}}\left(\frac{\partial f(\nu^\star,x)}{\partial x}\in\mathcal{T}_x\right)>0.49\qquad\text{and}\qquad{\mathbb{P}_{(x,y)\sim\mathcal{D}}\left(\frac{\partial f(\nu^\star,x)}{\partial x}\in\mathcal{T}_x^\perp\right)>0.49}.
\end{equation*}
\end{theorem}

\subsection{Explanations need to be respect {\it both} the model and the data}
\label{sec:model_and_data}

Can the alignment of feature attributions with the tangent space of the data manifold also serve as a sufficient criterion for explanations? To that this is not the case, consider the explanation algorithm that returns a random feature attribution from the tangent space. Indeed, a random feature attribution that lies in the tangent space is perceptually-aligned. However, it does not correspond to a direction that is particularly relevant for the class of the image (examples are depicted in appendix Figure \ref{fig:apx_mnist_32_random}). In addition, a random feature attribution is {\it completely unrelated to the model}. It consequently fails sanity checks that assess the relationship between the explanation and the model, such as the parameter randomization test proposed in \cite{adebayo2018sanity}. For these reasons, the alignment of a feature attribution with the tangent space of the data manifold can only be a necessary criterion for explanations. At the same time, the central idea behind the manifold hypothesis is that attributions without any meaningful relationship to the data (such as the orthogonal components depicted in the third row of Figure \ref{fig:mnist_gradients}) are also not explanations, even if they were highly salient to the model. Thus, we find that an attribution needs to fulfill {\it two different kinds of criteria} in order to be an explanation: It needs to be related to the model (at the minimum, pass the sanity checks) {\it and} to the structure of the data (our hypothesis: it must lie in the tangent space of the image manifold).

\section{Conclusion}
\label{sec:discussion}

In this work, we focus on a particular aspect of feature attributions: whether they are aligned with the tangent space of the data manifold. The objective of this paper is {\it not} to claim that the gradients of existing models provide good explanations, or that any particular post-hoc explanation method works especially well. Instead, we would like to contribute to a line of work that, independently of particular algorithms, develops criteria by which explanations can be judged.
As we demonstrate in Sections \ref{sec:results} and \ref{sec:theory}, the question of whether an attribution is aligned with the data manifold is amendable to empirical and theoretical analysis. %
While current models and algorithms provide only imperfect alignment, it is an open question whether this is due to the fact that we have not yet found the right model architecture or algorithm, or because the problem is more difficult than classification alone. To the best of our knowledge, the question of how model gradients can be aligned with the data manifold is essentially unexplored in the machine learning literature. Although we are, to the best of our knowledge, the first to conduct a systematic evaluation of the manifold hypothesis, aspects of it are implicit in previous works \citep{kim2019bridging,anders2020fairwashing,chang2018explaining,stutz2019disentangling,frye2020shapley}.

\section*{Acknowledgements}

The authors would like to thank Suraj Srinivas and anonymous reviewers for their helpful comments. This work has been supported by the German Research Foundation through the Cluster of Excellence “Machine Learning – New Perspectives for Science" (EXC 2064/1 number 390727645), the BMBF Tübingen AI Center (FKZ: 01IS18039A), and the International Max Planck Research School for Intelligent Systems (IMPRS-IS).

\bibliography{references}

\newpage
\appendix

\section{Model Architectures and Training Details}
\label{apx:experiments}

\subsection{MNIST32}
\label{apx:mnist_construction}

We first describe the creation of the MNIST32 dataset. We autoencoded the original MNIST dataset with a $\beta$-TCVAE \citep{chen2018isolating} and the same architecture as in \cite{burgess2018understanding}. The hyperparameters were $\alpha=\gamma=1$, $\beta=6$. We use \url{https://github.com/YannDubs/disentangling-vae} (MIT License). On the reconstructed images, we trained a SimpleNet-V1 to replicate the original labels \citep{hasanpour2016lets}. Training with Adam and a learning rate of $1e-3$ allowed to replicated the labels of the test images with an accuracy of 96\%. To increase the quality of the generated images, we additionally applied rejection sampling based on the the softmax score of the class predicted by the SimpleNet. Every sample from the autoencoder was accepted with probability $p_\text{softmax}^2$. Random samples from the MNIST32 dataset are depicted in Figure \ref{fig:apx_random_samples}. 

On the MNIST32 dataset, we trained the default model architecture from \url{https://github.com/pytorch/examples/tree/master/mnist}. We trained for 50 epochs with Adam, an initial learning rate of $1e-4$ and learning rate decay of $1e-1$ after 10 epochs each.

Adversarially robust training on MNIST32 was perfomed as follows. We trained the same model architecture against an $l_2$-adversary with projected gradient descent (PGD). For each gradient step, the size of the adversarial perturbation $\epsilon$ was randomly chosen from $[1, 4, 8]$ and we took $100$ iterations with a step size of $\alpha=2.5 \epsilon / 100$ each \citep{madry2017towards}.

To overfit the MNIST32 dataset with random labels, we disabled the dropout layers of the neural network. We then trained for 3000 epochs with Adam, and intial learning rate of $1e-4$ and learning rate decay of $1e-1$ after 1000 epochs each.

\subsection{MNIST256}

\begin{figure}[h]
\centering
     \begin{subfigure}{0.49\linewidth}
         \centering
            \includegraphics[width=\linewidth]{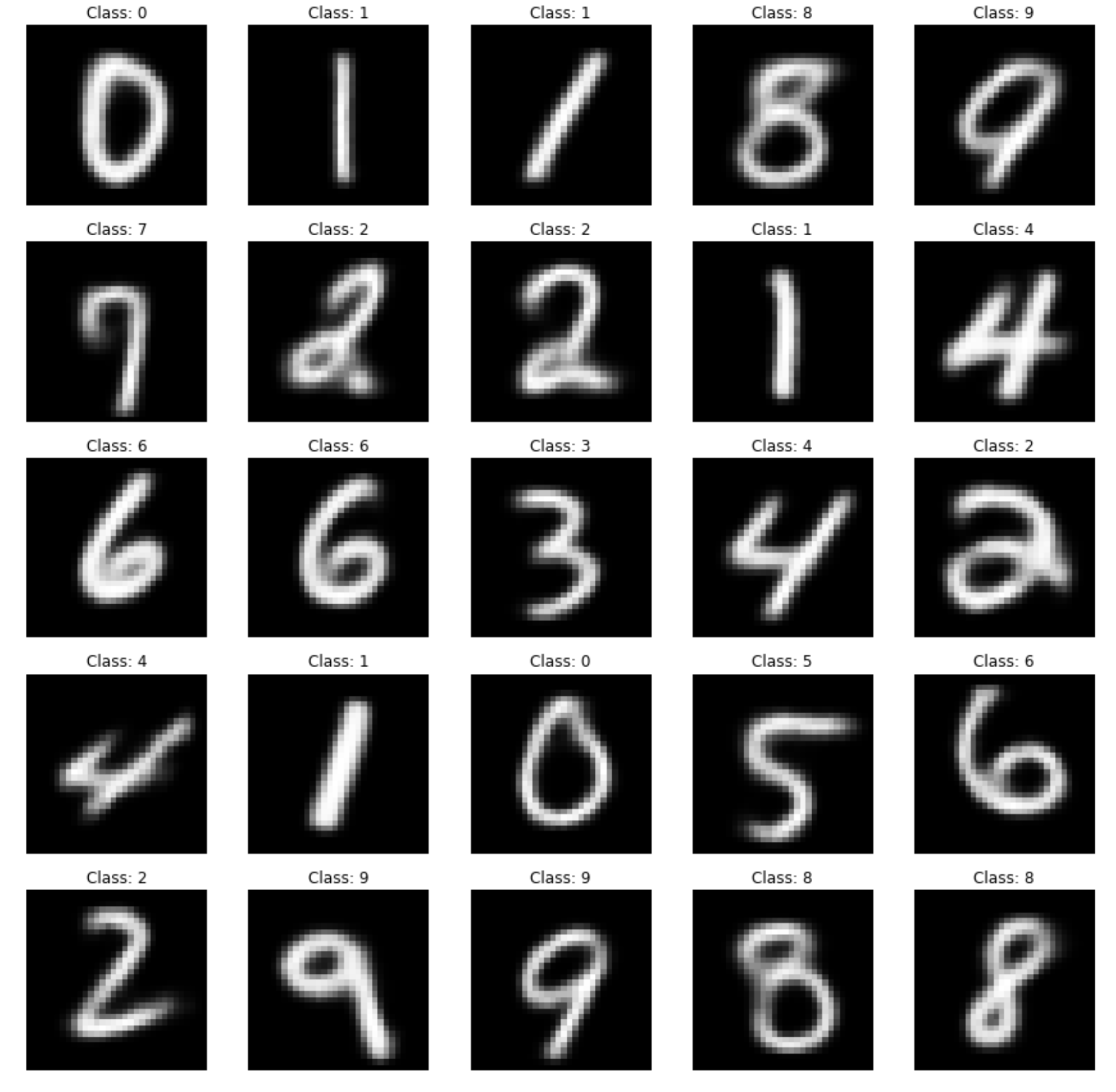}
     \end{subfigure}
     \hfill
     \begin{subfigure}{0.49\linewidth}
         \centering
    \includegraphics[width=\linewidth]{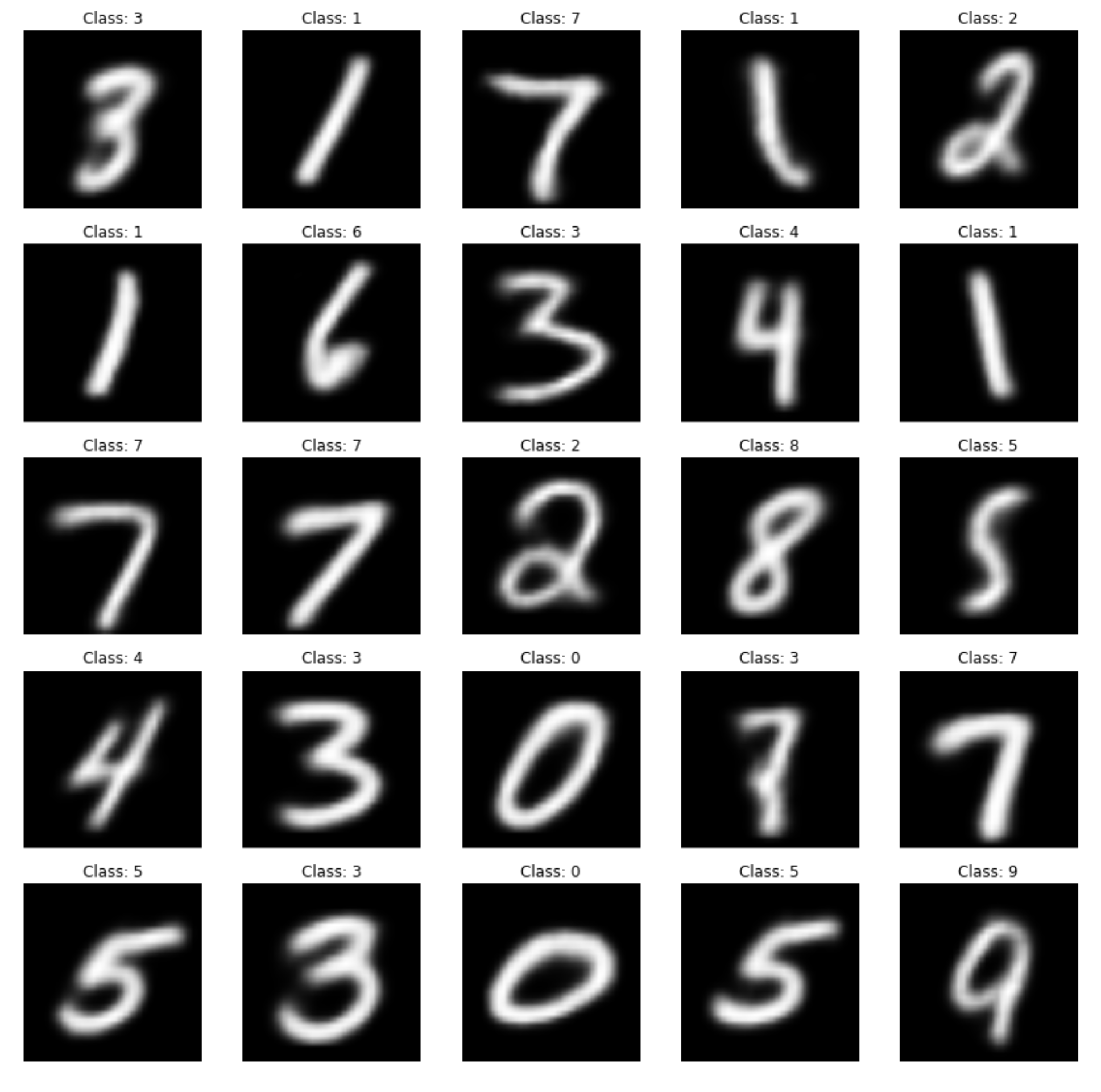}
     \end{subfigure}
    \caption{Random samples from the generated datasets. Left: MNIST32. Right: MNIST256.}
    \label{fig:apx_random_samples}
\end{figure}

To create the MNIST256 dataset, we appended a bilinear upsampling layer to the decoder that was used to generate the MNIST32 dataset. Note that bilinear upsampling is differentiable, which is required to compute the tangent spaces. Random samples from the MNIST256 dataset are depicted in Figure \ref{fig:apx_random_samples}.

On the MNIST256 dataset, we trained a ResNet18 for 50 epochs with Adam, an initial learning rate of $1e-2$ and a learning rate decay of $1e-1$ after 10 epochs each.

\subsection{EMNIST128}
\label{apx:emnist}
The EMNIST dataset is a set of handwritten character digits derived from the NIST Special Database 19  and converted to a $28\times28$ pixel image format and dataset structure that directly matches the MNIST dataset. We used the dataset as available from PyTorch \url{https://pytorch.org/vision/stable/datasets.html#emnist}.
The images were resized to $128\times128$ to make it a high-dimensional problem and we used a subset consisting of 60 classes (in contrast to other experiments where number of classes are typically low).
We trained an autoencoder using the reconstruction approach and encoder-decoder architecture using Adam optimizer with learning rate set to 1e-4, decayed over 200 epochs using cosine annealing. We then train a VGG network to perform the classification, in a similar manner.

\subsection{CIFAR10}
\label{apx:cifar10}
The CIFAR-10 dataset consists of 60000 32x32 colour images in 10 classes, with 6000 images per class. There are 50000 training images and 10000 test images.
We use the dataset available directly from the PyTorch dataloaders as described here \url{https://pytorch.org/vision/stable/datasets.html#cifar}. To learn the manifold, we use the reconstruction approach using an autoencoder with the latent dimension set to be 144 with $\sqrt{k/d} \approx 0.20$.
We use the Adam optimizer with learning rate set to 1e-4 decayed using cosine annealing over 200 epochs to learn the autoencoder.
We then trained a VGG16 classifier using Adam with an initial learning rate of 1e-4, again decayed using cosine annealing. The classifier achieved a test accuracy of 94.1\%.

\subsection{Pneumonia Detection}
\label{apx:pneumonia}
The original dataset at \url{https://www.kaggle.com/paultimothymooney/chest-xray-pneumonia} contains high-resolution chest X-ray images with 2 classes: Normal and Pneumonia (with pneumonia being of two types viral and bacterial, but within class distinction of pneumonia is not considered in this problem). The problem is posed as a binary classification problem to decide between the normal and an abnormal class (pneumonia).
The images were resized to $1\times256\times224$ (i.e., 57344 dimensional data) and the autoencoder is used to learn the manifold of the images where the latent dimension is reduced to $8\times28\times32$ (i.e., 7168) with $\sqrt{k/d} = 0.20$, we then fine tune a Resnet18 model (previously trained on ImageNet) to perform the classification using a learning rate of 1e-4, decayed with cosine annealing over 200 epochs and using Adam optimizer. The classifier achieved a test accuracy of 89\%.

\subsection{Diabetic Retinopathy Detection}
\label{apx:retinopathy}

The original 3 channel (RGB) fundus image dataset at\\ \url{https://www.kaggle.com/c/diabetic-retinopathy-detection} contains 5 classes with varying degrees of diabetic retinopathy. We posed the problem as a binary classification problem to decide between the normal and an abnormal class.
The images were resized to $3\times224\times224$ (i.e., 150528 dimensional data) and the autoencoder is used to learn the manifold of the images where the latent dimension is reduced to $8\times28\times28$ (i.e., 6272) with $\sqrt{k/d} = 0.20$, we then fine tune a Resnet18 model (previously trained on ImageNet) to perform the classification using a learning rate of 2e-4, decayed with cosine annealing over 150 epochs and using Adam optimizer. The classifier achieved a test accuracy of 92\%.

\subsection{Hardware}

All models were trained on NVIDIA GeForce RTX 2080 Ti GPUs, using an internal cluster. The total amount of compute required by this project was less than 1 GPU-year.

\begin{figure}
\centering
\begin{minipage}{\linewidth}
  \begin{algorithm}[H]
        \caption{The generative approach} \label{the_algorithm}
        \begin{algorithmic}
        \Require Dataset $X=(x_i,y_i)_{i=1}^n$.
        \Require Dimension of latent space $k\in[d]$.
        \State Train a variational autoencoder $q_\phi(p_\theta(x))$ on $X$ with latent dimension $k$.
        \State Latent states $\hat z_i\sim p_\theta(x_i)$ and reconstructions $\hat x_i\sim q_\phi(\hat z_i)$.
        \State Let $c:\mathbb{R}^{k+d}\to[C]$ solve  $(\hat z_i,\hat x_i)\to y_i$.\Comment{The labeling function}
        \State Sample $n$-times from the prior $\tilde z_i\sim \mathcal{N}(0,\mathbf{I}_k)$. \Comment{Draw the dataset}
        \State The dataset is $\tilde x_i\sim q_\phi(\tilde z_i)$, $\tilde y_i=c(\tilde z_i, \tilde x_i)$.
        \For{$i\in[n]$}         \Comment{Compute tangent spaces}
            \For{$l\in[d]$}
                \For{$m\in[k]$}
                    \State $t_{i,l,m}=\frac{\partial (q_\phi)_{l}}{\partial z_m}(\tilde z_i)$
                \EndFor
            \EndFor
            \State $\mathcal{T}_{\tilde x_i}=\text{span}<\begin{pmatrix}
            t_{i,1,0} \\
            \hdots \\
        	t_{i,d,0}
        \end{pmatrix},\cdots, \begin{pmatrix}
            t_{i,1,k} \\
            \hdots \\
        	t_{i,d,k}
        \end{pmatrix}>$
        \EndFor
        \State \Return $\,\left(\tilde x_i, \mathcal{T}_{\tilde x_i}, \tilde y_i\right)_{i=1}^n$\Comment{Data points, Tangent Spaces, Labels}
        \end{algorithmic}
        \end{algorithm}
\label{fig:generative_approach}
\caption{The generative approach to create a dataset with a known manifold structure.}
\end{minipage}
\end{figure}

\newpage
\section{Connection with Srinivas and Fleuret \citep{srinivas2020rethinking}}
\label{apx:connetion_srinivas}

We now highlight the the connections of the criterion in Srinivas and Fleuret \citep{srinivas2020rethinking} with our manifold hypothesis. Let $p_{\text{data}}(x\,|\,y=i)$ be the ground truth class-conditional density model. Let $p_\theta(x\,|\,y=i)=\frac{\exp(f_i(x))}{Z(\theta)/C}$ be the density model implicitly given by the classifier $f$ (compare Section 3 in \cite{srinivas2020rethinking}). As in the original derivation, we assume equiprobable classes. Alignment of the implicit density model with the ground truth class-conditional density model implies that $\nabla_x \log p_{\text{data}}(x\,|\,y=i)=\nabla_x f_i(x)$. We now assume that the data concentrates around a low-dimensional manifold, and then show that that $\nabla_x \log p_{\text{data}}(x\,|\,y=i)$ lies within the tangent space of the manifold. 

We first show that this holds true if the data concentrates uniformly around the manifold. Let us specify what it means that the data concentrates around a low-dimensional manifold. Let $\mathcal{M}$ be a $k$-dimensional manifold.  Let $p_{\text{data},\mathcal{M}}(x\,|\,y=i)$ be the ground-truth class-conditional density model on the manifold. That is $p_{\text{data},\mathcal{M}}(x\,|\,y=i)$ is a function that lives on the manifold $\mathcal{M}$. Now, every point $x\in\mathbb{R}^d$ can be written as $x=x_{\mathcal{M}}+z$, where $x_{\mathcal{M}}\in\mathcal{M}$ is the point on the manifold that is closest to $x$, and $z=x_\mathcal{M}-x$ is orthogonal to the tangent space $\mathcal{T}_{x_{\mathcal{M}}}$. Concentration of the data around the manifold means that the ground-truth class-conditional density of the data concentrates around the manifold. We assume that this can be written as\footnote{This is of course a simplifying assumption.}
\begin{equation}
\label{eq:manifold_concentration}
    p_{\text{data}}(x\,|\,y=i)=h(||x-x_\mathcal{M}||_2)\cdot p_{\text{data},\mathcal{M}}(x_{\mathcal{M}}\,|\,y=i)\quad\forall i.
\end{equation}
In words, the class-conditional density at $x$ is given by the class-conditional density of closest point on the manifold, times a term that accounts for the distance of $x$ to the manifold. 

By uniform concentration we mean that there exists a band of finite width $\epsilon$ around the data manifold, and that data points occur uniformly within this band. Formally,
\begin{equation}
h(r)=D\cdot1_{[0, \epsilon)}(r)
\end{equation}
where $1_A(r)$ denotes the indicator function of the set $A$ and $D$ is a normalization constant. Consequently,
\begin{equation}
\label{eq:uniform_on_manifold}
    p_{\text{data}}(x\,|\,y=i)=D\cdot 1_{\{||x-x_\mathcal{M}||_2<\epsilon\}} \cdot p_{\text{data},\mathcal{M}}(x_{\mathcal{M}}\,|\,y=i).
\end{equation}  
Under this assumption, alignment of the implicit density model with the ground truth class-conditional density model implies that the gradient $\nabla_x f_i(x)$ is aligned with the tangent space of the data manifold. To see this, first note that
\begin{equation*}
    \log p_{\text{data}}(x\,|\,y=i) = \log p_{\text{data},\mathcal{M}}(x_{\mathcal{M}}\,|\,y=i)+\log\left(D \right)
\end{equation*} 
for every point that is observed under the data distribution. Now, let  $t_1,\dots,t_k$ be an orthonormal basis of $\mathcal{T}_{x_\mathcal{M}}$, and let $v_1,\dots,v_{d-k}$ be an orthonormal basis of $\mathcal{T}_{x_\mathcal{M}}^\perp$. Since these vectors form an orthonormal basis of $\mathbb{R}^d$, the gradient of $\log p_{\text{data}}(x\,|\,y=i)$ can be written as
\begin{equation*}
   \nabla_x \log p_{\text{data}}(x\,|\,y=i) = \sum_{j=1}^k t_j \partial_{t_j}\log p_{\text{data}}(x\,|\,y=i) + \sum_{j=1}^{d-k} v_j \partial_{v_j}\log p_{\text{data}}(x\,|\,y=i)
\end{equation*}
where $\partial_{v}f$ denotes the directional derivative of $f$ in direction $v$. By definition of the directional derivative and \eqref{eq:uniform_on_manifold}, for all directions $v_j$ orthogonal to the data manifold,
\begin{equation}
\label{eq:partial_derivative}
\begin{split}
    \partial_{v_j}\log p_{\text{data}}( x\,|\,y=i)&=\lim_{\delta\to 0}\frac{\log p_{\text{data}}( x+\delta v_j\,|\,y=i)-\log p_{\text{data}}( x\,|\,y=i)}{\delta}\\
    &=\lim_{\delta\to 0}\frac{\log p_{\text{data},\mathcal{M}}(x_{\mathcal{M}}\,|\,y=i)+\log(D)-\log p_{\text{data},\mathcal{M}}(x_{\mathcal{M}}\,|\,y=i)-\log(D)}{\delta}\\[2pt]
    &=0.  
\end{split}
\end{equation}
Where we additionally assumed that the point $x_\mathcal{M}$ does not change if we move along a direction that is orthogonal to the tangent space $\mathcal{T}_{x_\mathcal{M}}$ (which is subject to a mild regularity condition on the manifold). Consequently, 
\begin{equation*}
   \nabla_x \log p_{\text{data}}(x\,|\,y=i) = \sum_{j=1}^k t_j \partial_{t_j}\log p_{\text{data}}(x\,|\,y=i)
\end{equation*}
which lies by definition in the tangent space of the data manifold.

While this clearly demonstrates that there are interesting connections between the work of Srinivas and Fleuret \citep{srinivas2020rethinking} and our manifold hypothesis, the assumption that the data concentrates uniformly around the manifold might be seen as unrealistic. Instead of (\ref{eq:uniform_on_manifold}), we might want to assume that the density decays as move away from the data manifold, for example according to 
\begin{equation}
    \label{eq:normal_decay}
    h(r)= D\cdot\exp(-r^2/2).
\end{equation}
Note that this approximately corresponds to the sampling process where we first sample a point on the data manifold and then add i.i.d. normal noise. Under this assumption, alignment of the implicit density model with the ground truth class-conditional density model still implies that the model gradient is aligned with the tangent space of the data manifold {\it for all data points that lie exactly on the manifold}. To see this, we compute again (\ref{eq:partial_derivative}) which now gives
\begin{equation*}
\begin{split}
\partial_{v_j}\log p_{\text{data}}( x\,|\,y=i)&=\lim_{\delta\to 0}\frac{\log p_{\text{data}}( x+\delta v_j\,|\,y=i)-\log p_{\text{data}}( x\,|\,y=i)}{\delta}\\
&=\lim_{\delta\to 0}\frac{\log(\exp(-||\delta v_j||_2^2)) - \log(\exp(-||0||_2^2))}{\delta}\\
&=\lim_{\delta\to 0}\frac{\delta^2}{\delta}=0.\\
\end{split}
\end{equation*}
In this computation, we assumed that $x=x_\mathcal{M}$. If this is note that case, that is if we move away from the data manifold, we have instead
\begin{equation}
\label{eq:non-zero-partial}
\begin{split}
\partial_{v_j}\log p_{\text{data}}( x\,|\,y=i)&=\lim_{\delta\to 0}\frac{\log p_{\text{data}}( x+\delta v_j\,|\,y=i)-\log p_{\text{data}}( x\,|\,y=i)}{\delta}\\
&=-\lim_{\delta\to 0}\frac{||x-x_\mathcal{M}+\delta v_j||_2^2-||x-x_\mathcal{M}||_2^2}{\delta}\\
&=-\partial_{v_j}||x-x_\mathcal{M}||_2^2\\
&>0.
\end{split}
\end{equation}
Note that this term is determined solely by the distance of the point to the data manifold. In particular, it does not depend on the class $i$. Moreover, it can become quite large: The gradient of the ground truth class-conditional density model can be dominated by directions of quick decay of the overall probability density as we move away from the low-dimensional manifold around which the data concentrates. For this reason, we propose the following normalization: Instead of being aligned with $\log p_{\text{data}}(x\,|\,y=i)$, the implicit density model should be aligned (up to a constant factor) with
\begin{equation}
\label{eq:modified_alignment}
\log\frac{p_{\text{data}}(x\,|\,y=i)}{p_\text{data}(x)}.
\end{equation}
If the overall data distribution is relatively uniform, this normalization does not matter for the derivative. However, if the data tightly concentrates around a low dimensional manifold, for example according to \eqref{eq:normal_decay}, then $\nabla_x f_i(x)=\nabla_x \log \left( p_{\text{data}}(x\,|\,y=i)/p(x)\right )$ again implies that the gradient is aligned with the tangent space of the data manifold. In fact, if the data distribution {\it on the manifold} is close to uniform, that is if $p(x)\approx p(y)$ for all $x,y\in\mathcal{M}$, then alignment of the implicit density model with (\ref{eq:modified_alignment}) implies that the implicit density model is aligned with the ground-truth class conditional density model {\it on the manifold}. To see this, first note that
\begin{equation*}
\begin{split}
        p_\text{data}(x)&=\sum_{i=1}^C p_\text{data}(x\,|\,y=i)\\
        &=\sum_{i=1}^C h(||x-x_\mathcal{M}||)\cdot p_\text{data}(x_\mathcal{M}\,|\,y=i)\\[6pt]
        &=h(||x-x_\mathcal{M}||)\cdot p_\text{data}(x_\mathcal{M})\\
\end{split}
\end{equation*}
and consequently
\begin{equation*}
    \log\frac{p_{\text{data}}(x\,|\,y=i)}{p_\text{data}(x)}=\log \frac{p_{\text{data}}( x_\mathcal{M}\,|\,y=i)}{p_{\text{data}}( x_\mathcal{M})}.
\end{equation*}
By the same argument as above,
\begin{equation*}
\partial_{v_j}\log \frac{p_{\text{data}}( x_\mathcal{M}\,|\,y=i)}{p_{\text{data}}( x_\mathcal{M})}=0
\end{equation*}
and thus
\begin{equation*}
   \nabla_x \log \frac{p_{\text{data}}( x\,|\,y=i)}{p_{\text{data}}(x)} = \sum_{j=1}^k t_j \partial_{t_j}\log p_{\text{data}}(x\,|\,y=i)+\sum_{j=1}^k t_j \partial_{t_j}\log p_{\text{data}}(x)
\end{equation*}
which lies in $\mathcal{T}_{x_\mathcal{M}}$. In addition, since $\partial_{t_j}\log p_{\text{data}}(x)=\partial_{t_j}\log p_{\text{data}}(x_\mathcal{M})$, the second term vanishes if $p_\text{data}(x)$ on the manifold is close to uniform.

\section{Proof of Theorem \ref{thm:1}}
\label{apx:proof}

\begin{proof}
Let $d>1$. We begin by defining the manifold. Let\begin{equation}
t_\text{max} = \begin{cases}
    4(d-2) \quad \text{if } d \text{ is even}\\
    4(d-1) \quad \text{if } d \text{ is odd}.
 \end{cases}
\end{equation}
For $t\in[0,t_\text{max}]$, consider the continuous curve $f(t)$ that walks along the edges of the shifted hypercube, alternating between the first and other dimensions
\begin{equation}
f(t) = \begin{cases}
\left(-1/2+(t-\floor{t}),\smash{\underbrace{1,\dots,1}_{\floor{t/2}}},\smash{\underbrace{0,\dots,0}_{d-1-\floor{t/2}}}\right)^t & \quad \text{if } \floor{t}\mod 4=0\\[25pt]
    \left(1/2,\smash{\underbrace{1,\dots,1}_{\floor{t/2}}},t-\floor{t},\smash{\underbrace{0,\dots,0}_{d-2-\floor{t/2}}}\right)^t & \quad \text{if } \floor{t}\mod 4=1\\[25pt]
    \left(1/2-(t-\floor{t}),\smash{\underbrace{1,\dots,1}_{\floor{t/2}}},\smash{\underbrace{0,\dots,0}_{d-1-\floor{t/2}}}\right)^t & \quad \text{if } \floor{t}\mod 4=2\\[25pt]
    \left(-1/2,\smash{\underbrace{1,\dots,1}_{\floor{t/2}}},t-\floor{t},\smash{\underbrace{0,\dots,0}_{d-2-\floor{t/2}}}\right)^t & \quad \text{if } \floor{t}\mod 4=3.\\[20pt]
  \end{cases}
\end{equation}
In all dimensions, $f(t)$ starts at $(-1/2,0,\dots,0)^t$. If $d$ is even, $f(t)$ ends at $(-1/2,1,\dots,1,0)^t$. If $d$ is odd, $f(t)$ ends at $(-1/2,1,\dots,1)^t$. In even dimensions, connect the endpoint $(-1/2,1,\dots,1,0)^t$ to the starting point via straight lines to the corner points $(\sqrt{d-1}/2,1,\dots,1,0)^t$, $(\sqrt{d-1}/2,1,\dots,1,1)^t$, and $(-1/2,1,\dots,1,1)^t$ In odd dimensions, connect the endpoint $(-1/2,1,\dots,1)^t$ to the starting point via straight lines to the corner points $(-1/2,2/3,\dots,2/3)^t$, $(-1/2+\sqrt{d-1}/2,2/3,\dots,2/3)^t$, $(-1/2+\sqrt{d-1}/2,1/3,\dots,1/3)^t$, $(-1/2,1/3,\dots,1/3)^t$. The whole point of this construction is to obtain a closed connected curve that does not lie in any proper subspace and that walks exactly half of the time along the first coordinate, and the rest of the time orthogonal to it. By smoothing the corners of this connected curve, we obtain a smooth connected manifold $\mathcal{M}$.

Let $U_\mathcal{M}$ be the uniform distribution on $\mathcal{M}$. Let $\mathcal{D}_x$ be given by $\mathcal{D}_x(A)=U_\mathcal{M}(A/M)/(1-U_\mathcal{M}(M))$ where $M=\{x\in\mathcal{M}:|x_1|<\epsilon\}$. Let the label be given by $y=\sign x_1$. The separating hyperplane with maximum margin is $x_1=0$. We claim that $\nu^\star=\frac{1}{2}\delta_{\theta_0^\star}+\frac{1}{2}\delta_{\theta_1^\star}$, $\theta_0^\star=\left(\frac{1}{\sqrt{2}},\frac{1}{\sqrt{2(1+\epsilon^2)}},0,\dots,0,\frac{\epsilon}{\sqrt{2(1+\epsilon^2)}}\right)^t$, $\theta_1^\star=\left(\frac{-1}{\sqrt{2}},\frac{-1}{\sqrt{2(1+\epsilon^2)}},0,\dots,0,\frac{\epsilon}{\sqrt{2(1+\epsilon^2)}}\right)^t$, is a maximizer of \begin{equation}
\argmax_{\nu\in\mathcal{P}(\mathbb{S}^{d+1})}\min_{(x,y)\in\mathcal{D}}y\cdot f(\nu,x).
\end{equation}
By Proposition 12 in \citep{chizat2020implicit}, we have to show that there exists a measure $p^\star$ on $\mathcal{M}$ (the support vectors) such that\begin{equation} \label{first_condition}
\text{Support}(\nu^\star)\in\argmax_{(w,a,b)\in\mathbb{S}^{d+1}}\mathbb{E}_{(x,y)\sim p^\star}\left(y\cdot w\phi(\langle\,a,x\rangle+b)\right)
\end{equation}
and
\begin{equation} \label{second_condition}
\text{Support}(p^\star)\in\argmin_{(x,y)\in\mathcal{D}}\mathbb{E}_{(w,a,b)\sim \nu^\star}\left(y\cdot w\phi(\langle\,a,x\rangle+b)\right).
\end{equation}
We claim that $p^\star$ is given by\begin{equation}
    p^\star=\frac{1}{2}\delta_{(-\epsilon, 0, \dots, 0)^t}+\frac{1}{2}\delta_{(\epsilon, 0, \dots, 0)^t}.
\end{equation}

We first show (\ref{first_condition}). It holds that\begin{equation*}
\mathbb{E}_{(x,y)\sim p^\star}\left(y\cdot w\phi(\langle\,a,x\rangle+b)\right)=\frac{w}{2}\left(\phi(a_1 \epsilon+b)-\phi(-a_1 \epsilon+b)\right).
\end{equation*}
We differentiate two cases. Note that $\theta_0^\star$ achieves an objective larger than zero, hence $a_1\neq 0$. 

{\bf Case 1, $a_1>0$.} If $a_1>0$, then $\phi(a_1\epsilon+b)>\phi(-a_1\epsilon+b)$. This implies $b\geq 0$ and $b\leq a_1\epsilon$. The maximization problem can then be written as
\begin{equation*}
\begin{aligned}
& \underset{w,a_1,b}{\text{max}}
& & \frac{w}{2}a_1\epsilon+\frac{w}{2}b\\
& \text{subject to}
& & w^2+a_1^2+b^2=1\\
&&& 0\leq b \leq a_1\epsilon\\
&&& a_1 > 0.
\end{aligned}
\end{equation*}
For $\epsilon$ small enough, the unique solution is given by $b=a_1\epsilon$, $w=\frac{1}{\sqrt{2}}$ and $a_1=\frac{1}{\sqrt{2(1+\epsilon^2)}}$, i.e. by $\theta_0^\star$. The objective is $1/(2\sqrt{1+\epsilon^2})$.

{\bf Case 2, $a_1<0$.} If $a_1<0$, then $\phi(a_1\epsilon+b)<\phi(-a_1\epsilon+b)$. This implies $b\geq 0$ and $b\leq -a_1\epsilon$. The maximization problem can now be written as
\begin{equation*}
\begin{aligned}
& \underset{w,a_1,b}{\text{max}}
& & -\frac{w}{2}a_1\epsilon+\frac{w}{2}b\\
& \text{subject to}
& & w^2+a_1^2+b^2=1\\
&&& 0\leq b \leq -a_1\epsilon\\
&&& a_1 < 0.
\end{aligned}
\end{equation*}
For $\epsilon$ small enough, the unique solution is given by $b=-a_1\epsilon$, $w=\frac{1}{\sqrt{2}}$, and $a_1=\frac{1}{\sqrt{2(1+\epsilon^2)}}$, i.e. by $\theta_1^\star$.  The objective is again $1/(2\sqrt{1+\epsilon^2})$. This shows (\ref{first_condition}).

We now show (\ref{second_condition}). Explicit computation shows\begin{equation*}
\mathbb{E}_{(w,a,b)\sim \nu^\star}\left(y\cdot w\phi(\langle\,a,x\rangle+b)\right)=\frac{y}{\sqrt{2}}\phi\left(\frac{x_1+\epsilon}{\sqrt{2(1+\epsilon^2)}}\right)-\frac{y}{\sqrt{2}}\phi\left(\frac{-x_1+\epsilon}{\sqrt{2(1+\epsilon^2)}}\right).
\end{equation*}
For $y=1$, $x_1\geq\epsilon$ and the second term vanishes. The minimum is then attained iff $x_1=\epsilon$. For $y=-1$, $x_1\leq\epsilon$ and the first term vanishes. The minimum is then attained iff $x_1=-\epsilon$. This proves (\ref{second_condition}).

We now compute the gradient of $f$. We have \begin{equation}
f(\nu^\star,x)=\frac{1}{\sqrt{2}}\phi\left(\frac{x_1+\epsilon}{\sqrt{2(1+\epsilon^2)}}\right)-\frac{1}{\sqrt{2}}\phi\left(\frac{-x_1+\epsilon}{\sqrt{2(1+\epsilon^2)}}\right).
\end{equation}
Thus, for $i>1$,\begin{equation}
\frac{\partial f(\nu^\star,x)}{\partial x_i}=0.
\end{equation}
For $i=1$,\begin{equation}
\frac{\partial f(\nu^\star,x)}{\partial x_1}=\frac{1}{2\sqrt{1+\epsilon^2}}.
\end{equation}
Thus, the gradient of $f$ is constant and a multiple of $(1,0,\dots,0)^t$.

Except at the corners, the tangent space of $\mathcal{M}$ is either given by $\text{span}\langle\,(1,0,\dots,0)^t\rangle$ or orthogonal to $(1,0,\dots,0)^t$. The proof is completed by noting that it is orthogonal to $(1,0,\dots,0)^t$ with probability 0.5, that we can smooth the corners in regions of arbitrarily small measure, and by choosing $\epsilon$ arbitrarily small.
\end{proof}

\newpage
\section{Plots and Figures}
\label{apx:figures}

\subsection{Figure Creation Details}

Figure \ref{fig:mnist_gradients} was created by normalizing both vectors to unit norm and trimming the pixels with the largest 0.005\% absolute values. Images within the same column lie within the same color space, i.e. equal colors imply equal values along the respective coordinates. 

\subsection{Additional Plots and Figures}

\begin{figure}[h]
    \centering
    \includegraphics[width=\textwidth]{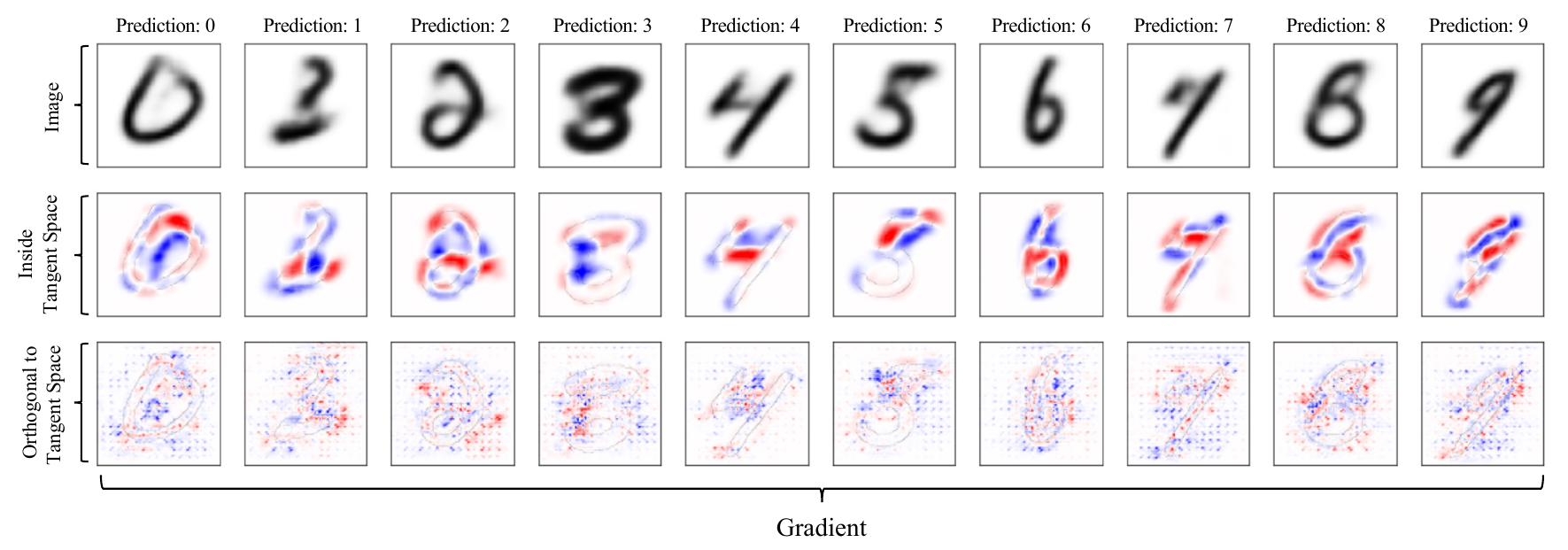}
    \caption{The part of an attribution that lies in tangent space is perceptually-aligned, whereas the part that is orthogonal to the tangent space is not. (First row) Images from the test set of MNIST256. (Second row) The part of the attribution that lies in tangent space. (Third row) The part of attribution that is orthogonal to the tangent space. Red corresponds to positive, blue to negative attribution.}
    \label{fig:apx_mnist_256_images}
\end{figure}

\newpage

\begin{figure*}
    \centering
    \includegraphics[width=\textwidth]{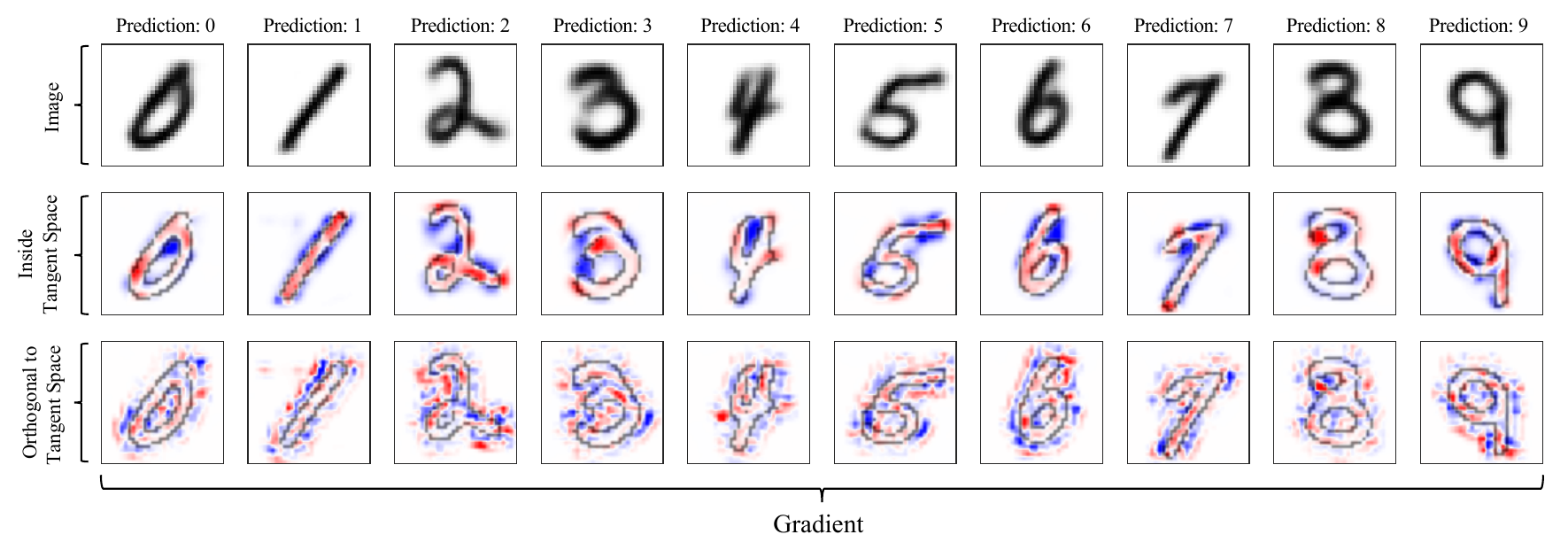}
    \includegraphics[width=\textwidth]{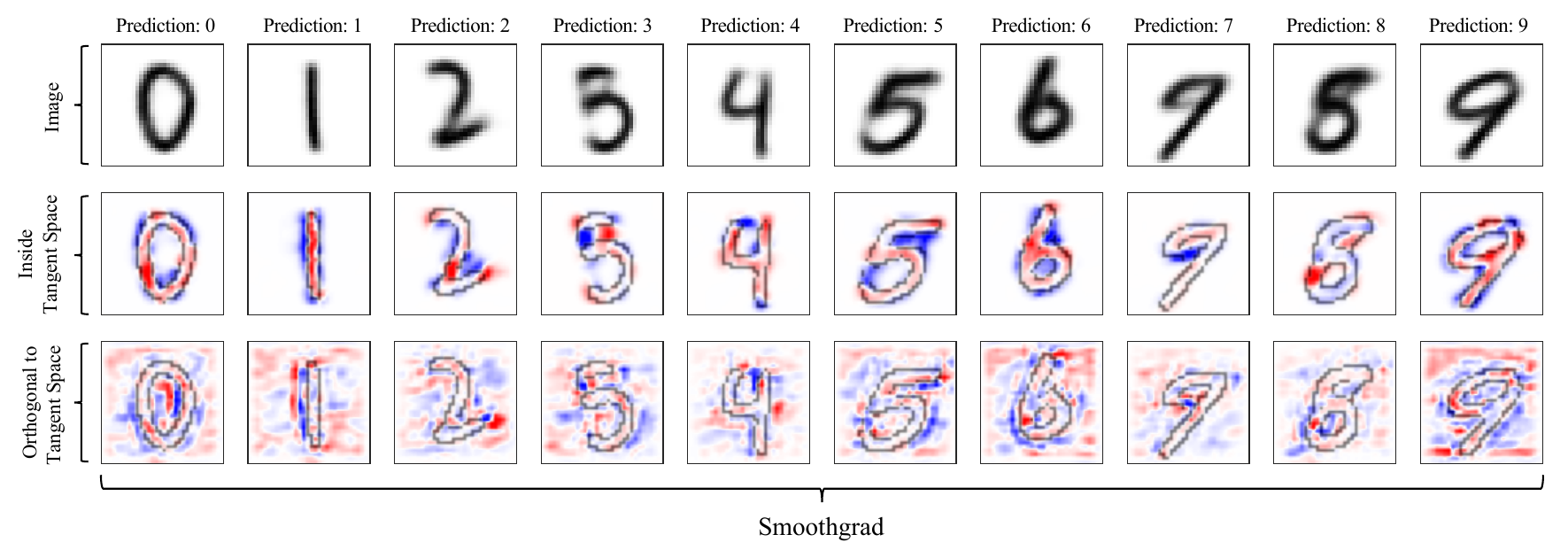}
    \includegraphics[width=\textwidth]{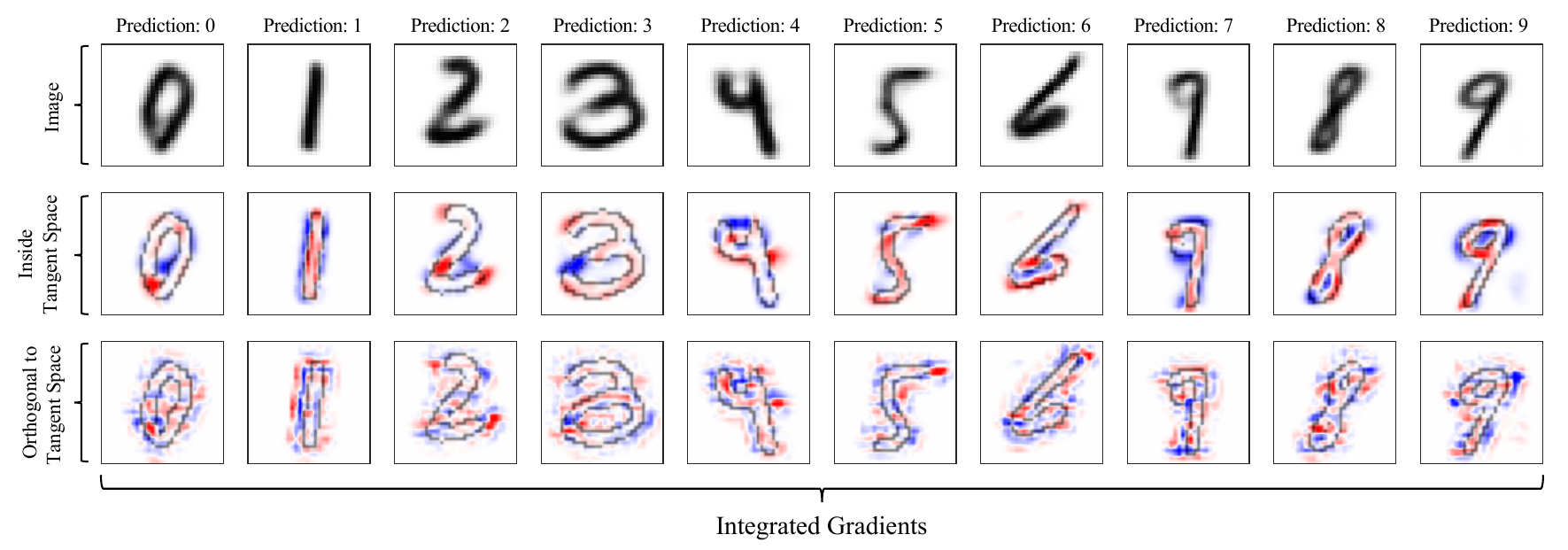}
    \caption{The part of an attribution that lies in tangent space is perceptually-aligned, whereas the part that is orthogonal to the tangent space is not. (First row) Images from the test set of MNIST32. (Second row) The part of the attribution that lies in tangent space. (Third row) The part of attribution that is orthogonal to the tangent space. Red corresponds to positive, blue to negative attribution.}
    \label{fig:apx_mnist_32_additional_attributions}
\end{figure*}

\clearpage
\newpage

\begin{figure}[ht!]
    \centering
    \includegraphics[width=\textwidth]{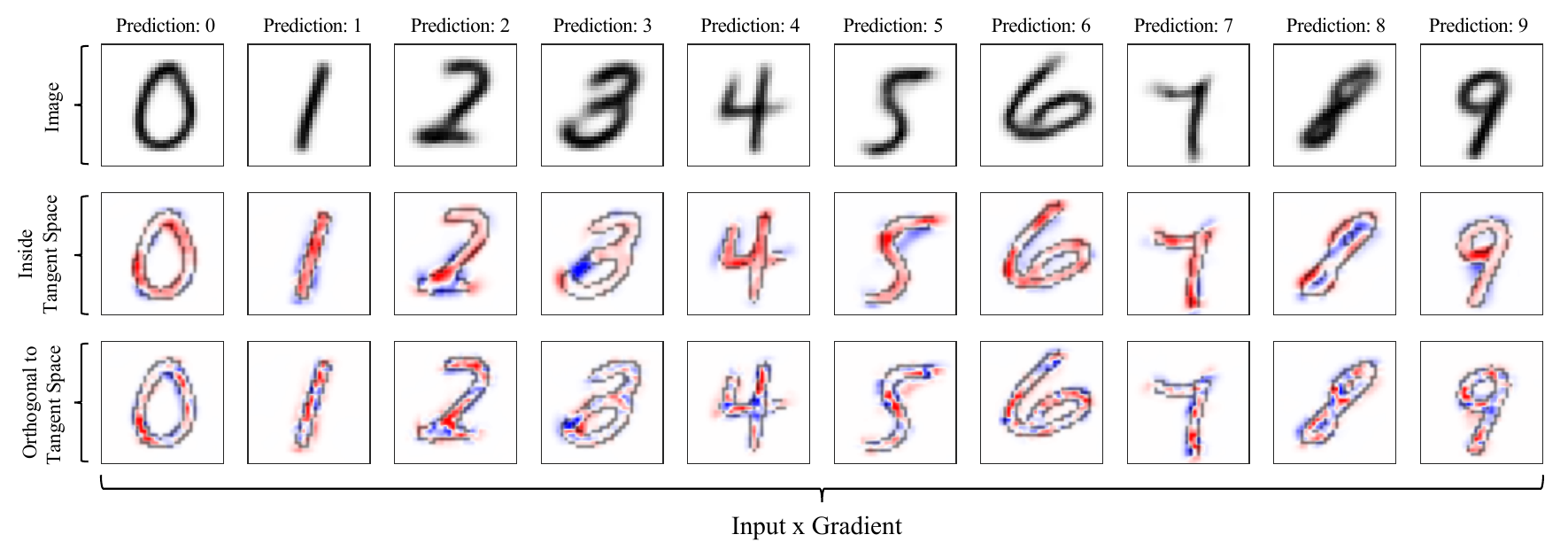}
    \includegraphics[width=\textwidth]{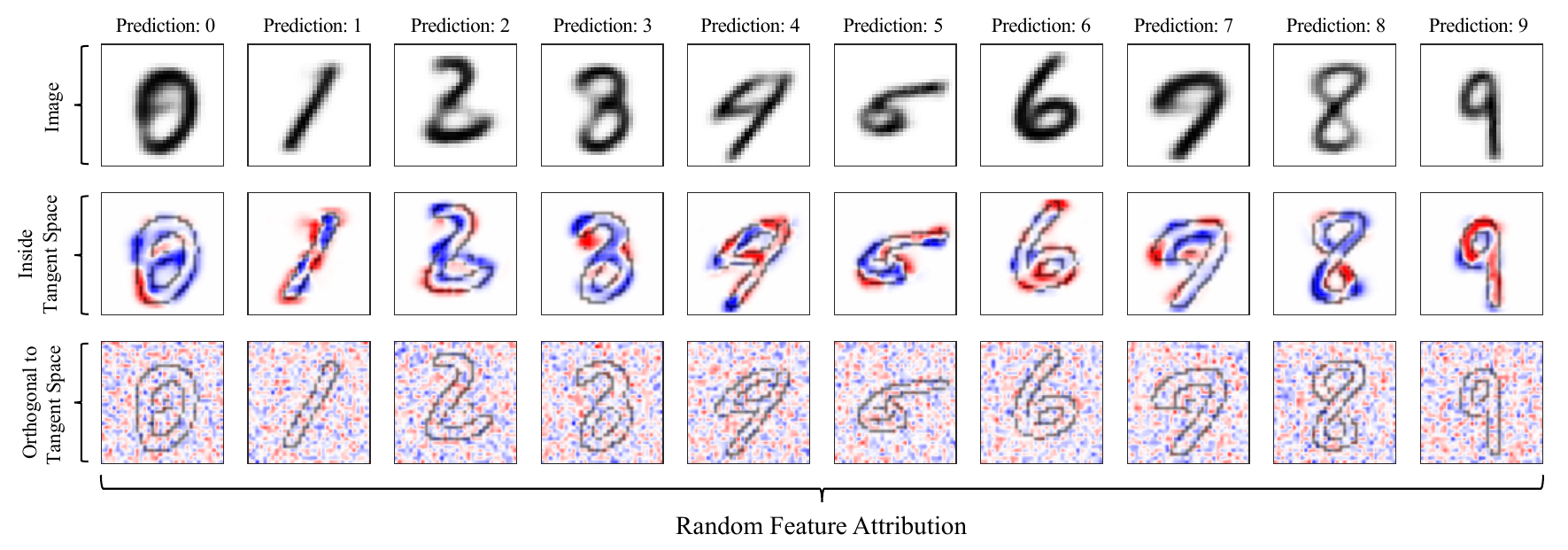}
    \caption{The part of an attribution that lies in tangent space is perceptually-aligned, whereas the part that is orthogonal to the tangent space is not. (First row) Images from the test set of MNIST32. (Second row) The part of the attribution that lies in tangent space. (Third row) The part of attribution that is orthogonal to the tangent space. Red corresponds to positive, blue to negative attribution.}
    \label{fig:apx_mnist_32_random}
\end{figure}

\clearpage
\newpage

\begin{figure}[ht!]
\centering
     \begin{subfigure}{0.32\linewidth}
         \centering
            \includegraphics[width=\linewidth]{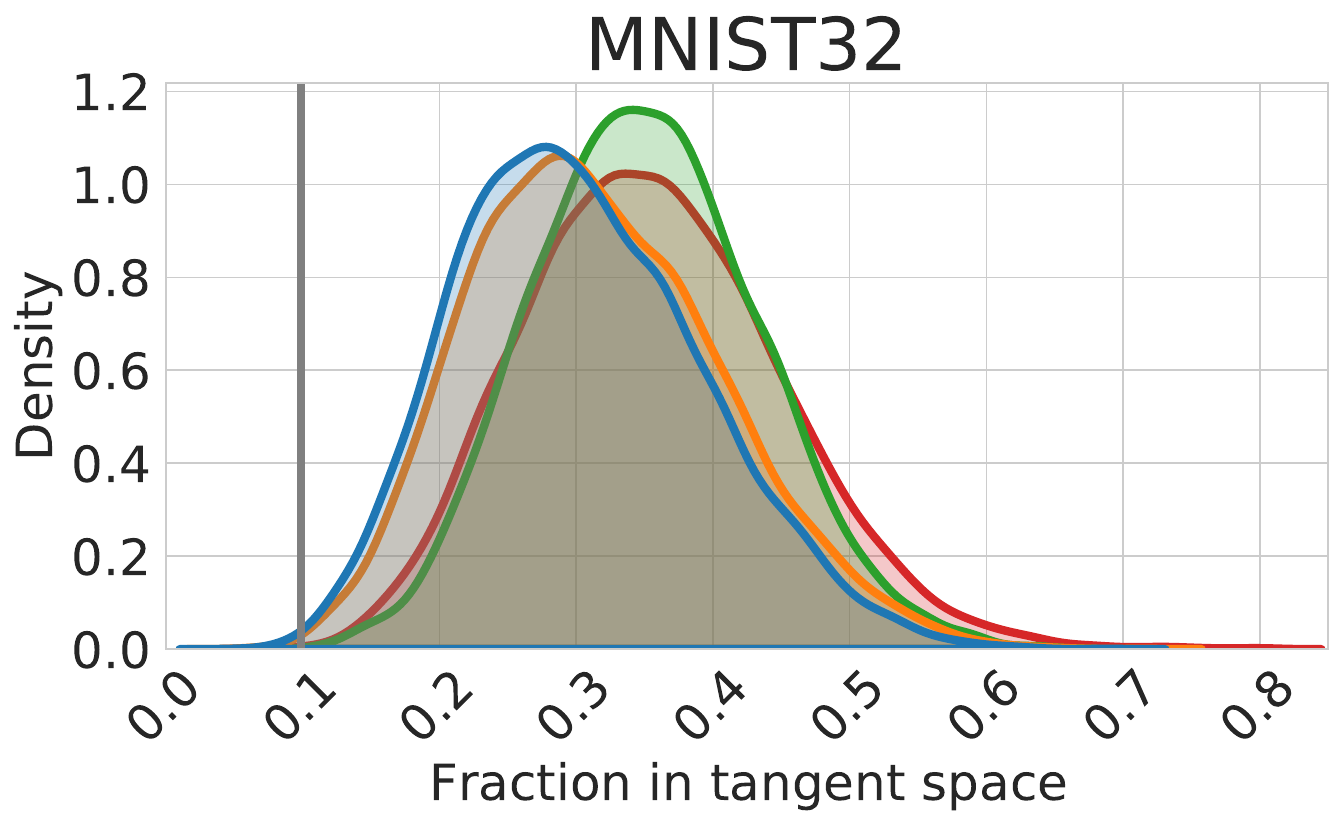}
     \end{subfigure}
     \hfill
     \begin{subfigure}{0.32\linewidth}
         \centering
            \includegraphics[width=\linewidth]{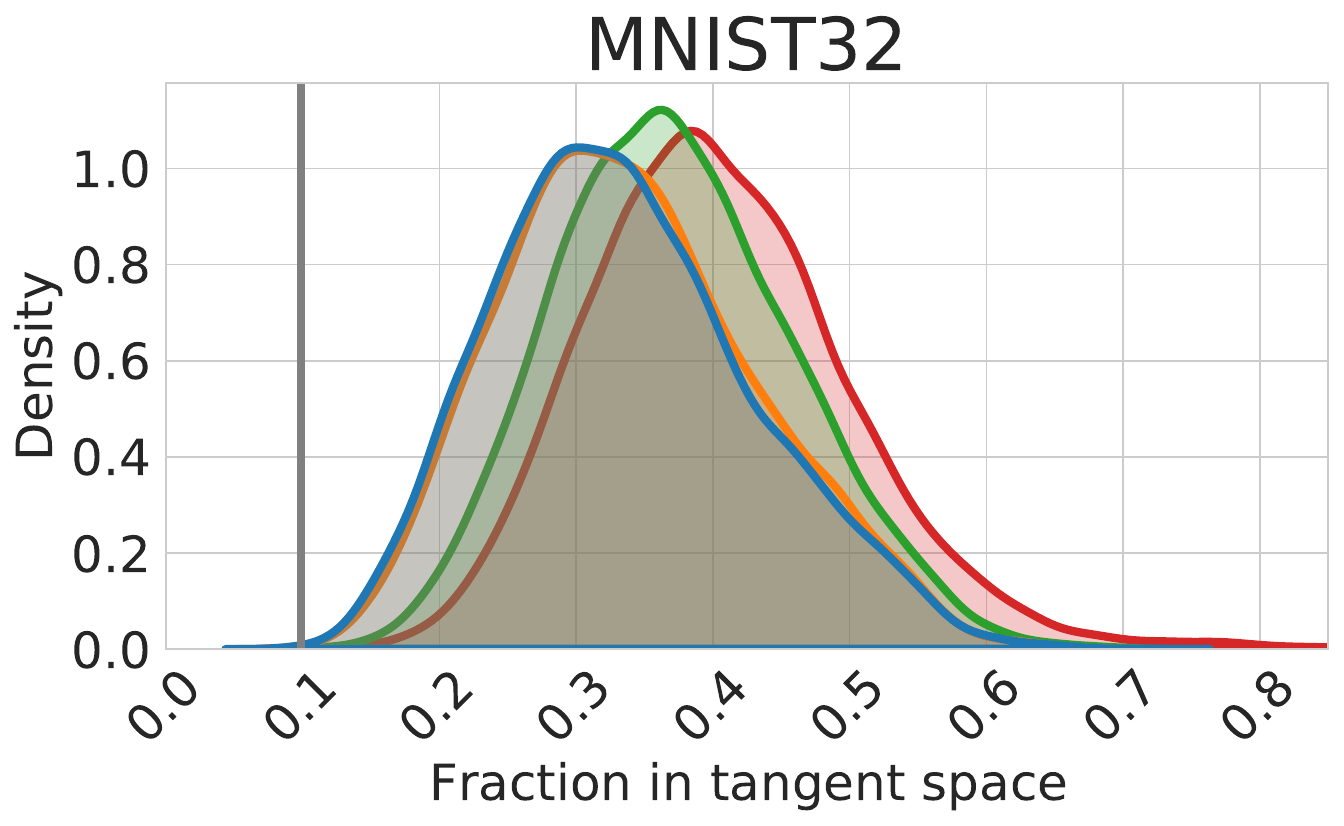}
     \end{subfigure}
     \hfill     \begin{subfigure}{0.32\linewidth}
         \centering
            \includegraphics[width=\linewidth]{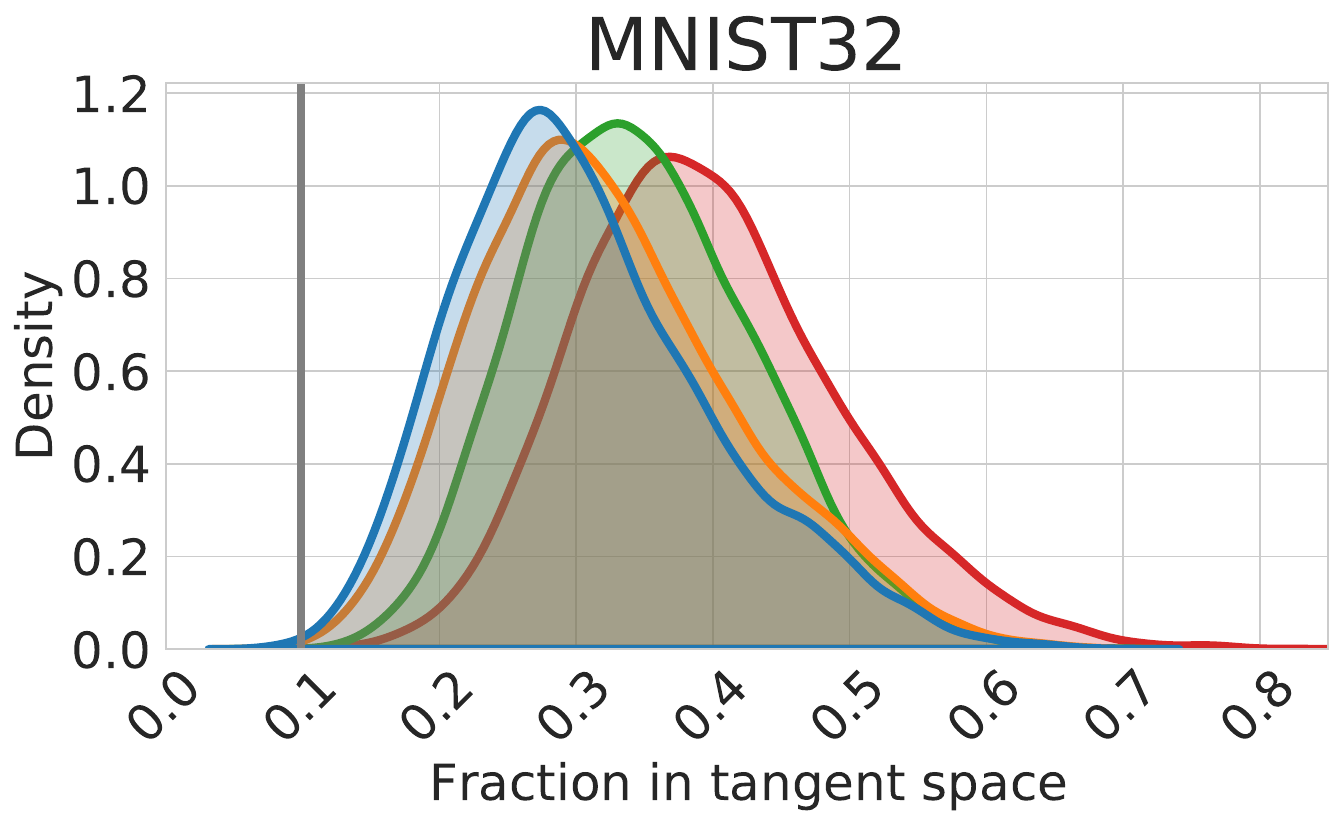}
     \end{subfigure}
     \hfill     \begin{subfigure}{0.32\linewidth}
         \centering
            \includegraphics[width=\linewidth]{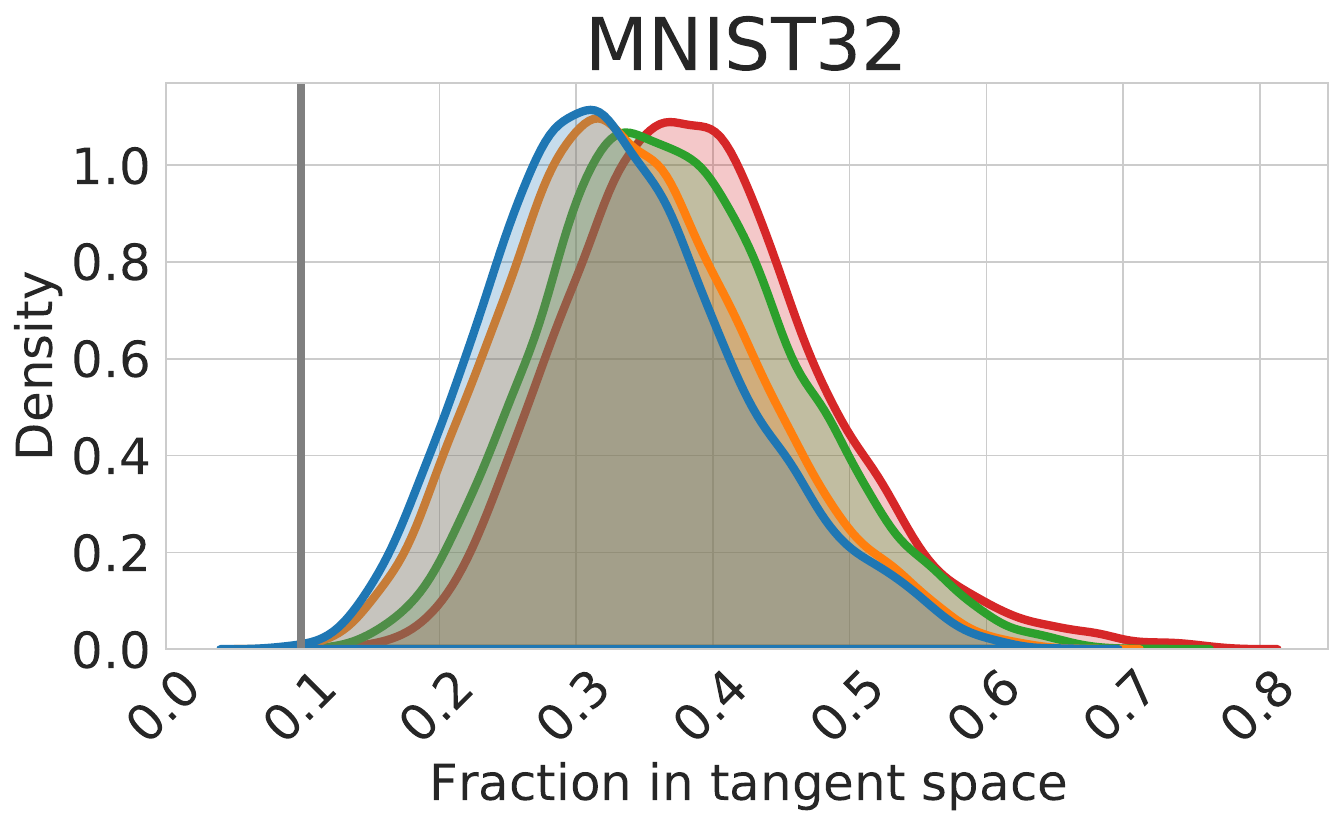}
     \end{subfigure}
     \hfill     \begin{subfigure}{0.32\linewidth}
         \centering
            \includegraphics[width=\linewidth]{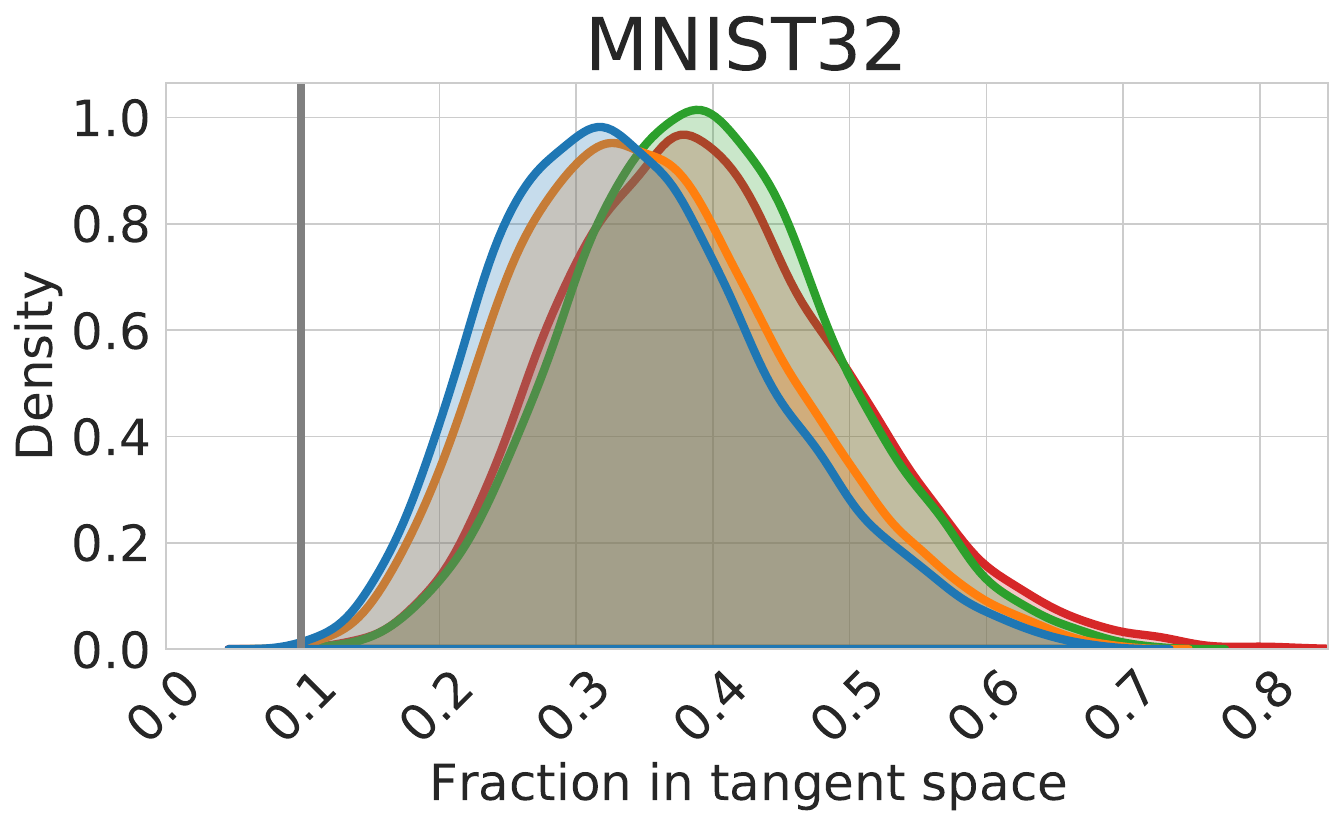}
     \end{subfigure}
     \hfill     \begin{subfigure}{0.32\linewidth}
         \centering
            \includegraphics[width=\linewidth]{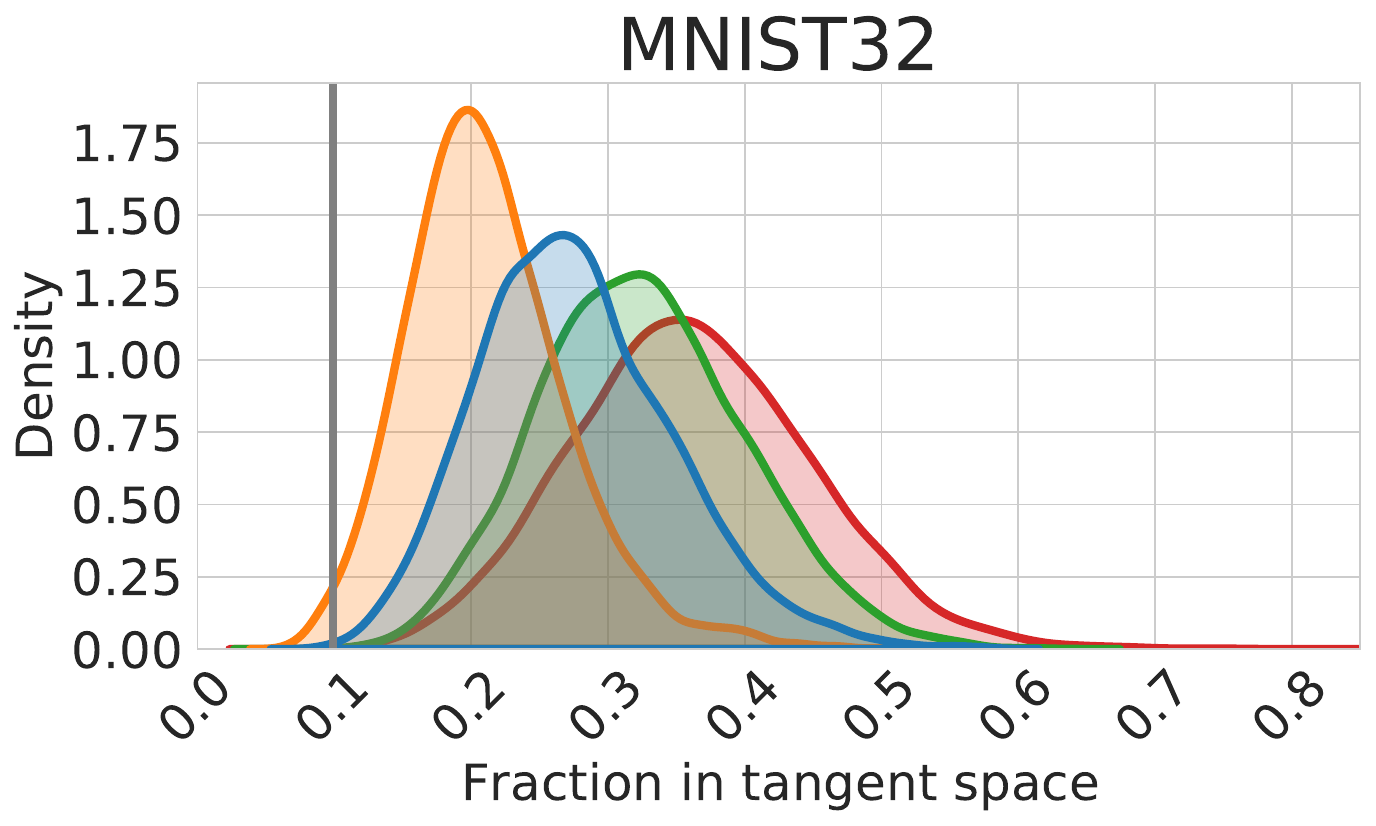}
     \end{subfigure}
     \hfill     \begin{subfigure}{0.32\linewidth}
         \centering
            \includegraphics[width=\linewidth]{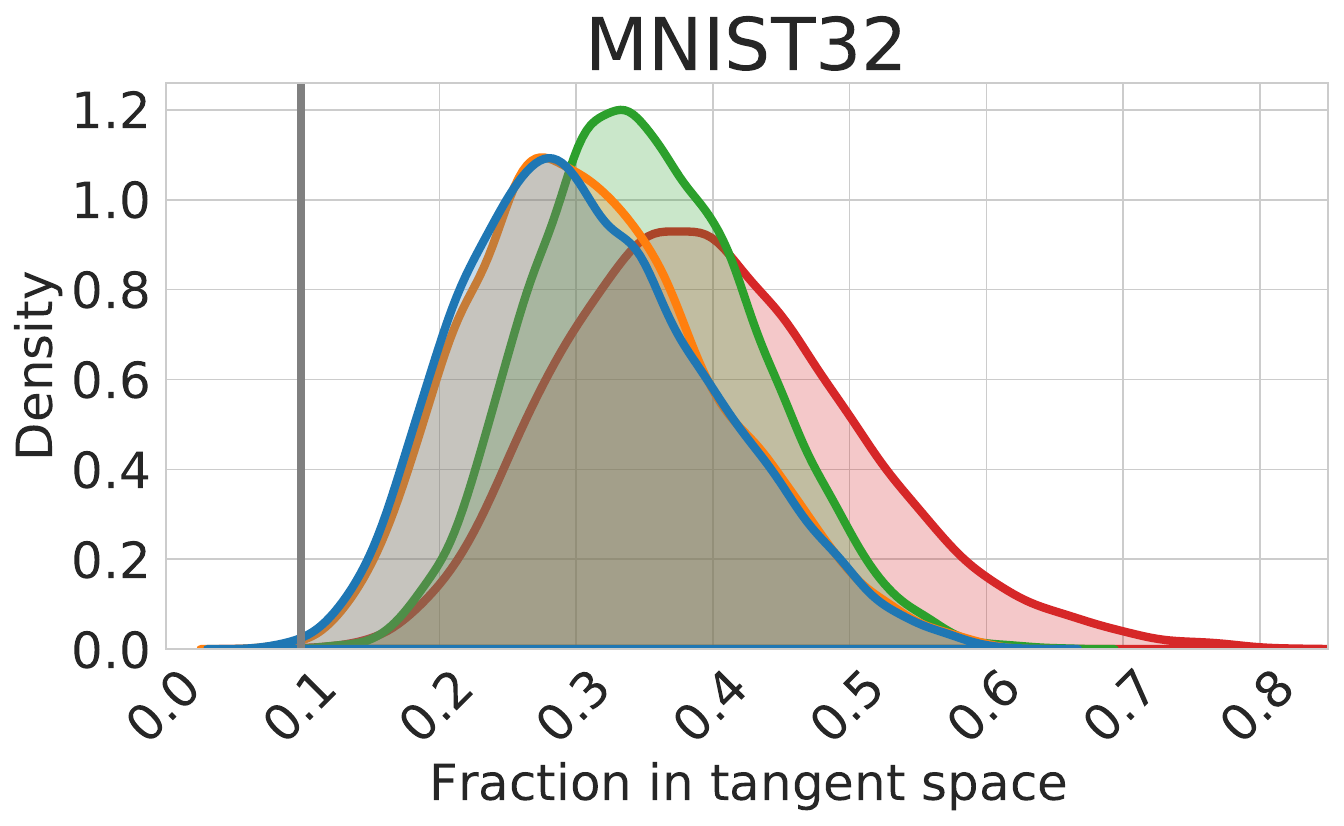}
     \end{subfigure}
     \hfill     \begin{subfigure}{0.32\linewidth}
         \centering
            \includegraphics[width=\linewidth]{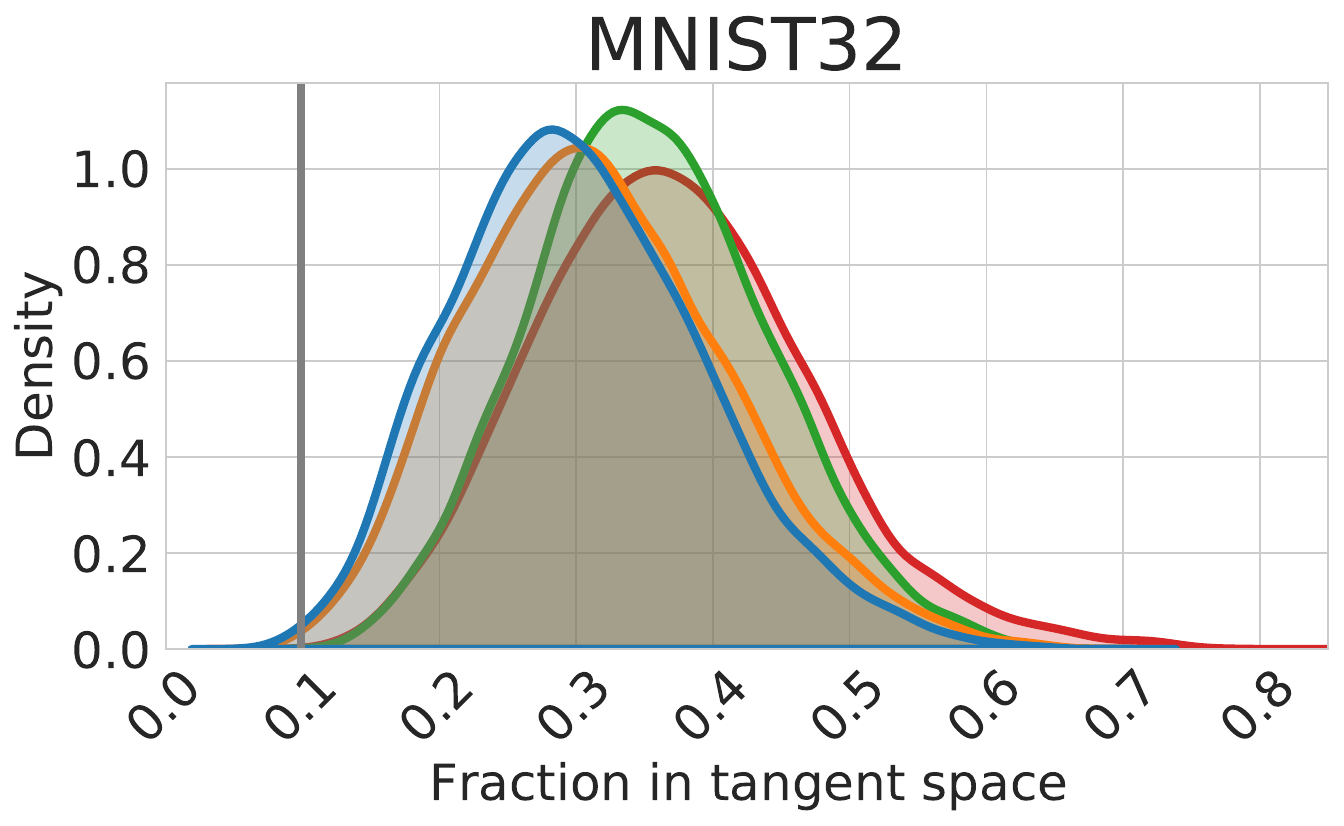}
     \end{subfigure}
     \hfill     \begin{subfigure}{0.32\linewidth}
         \centering
            \includegraphics[width=\linewidth]{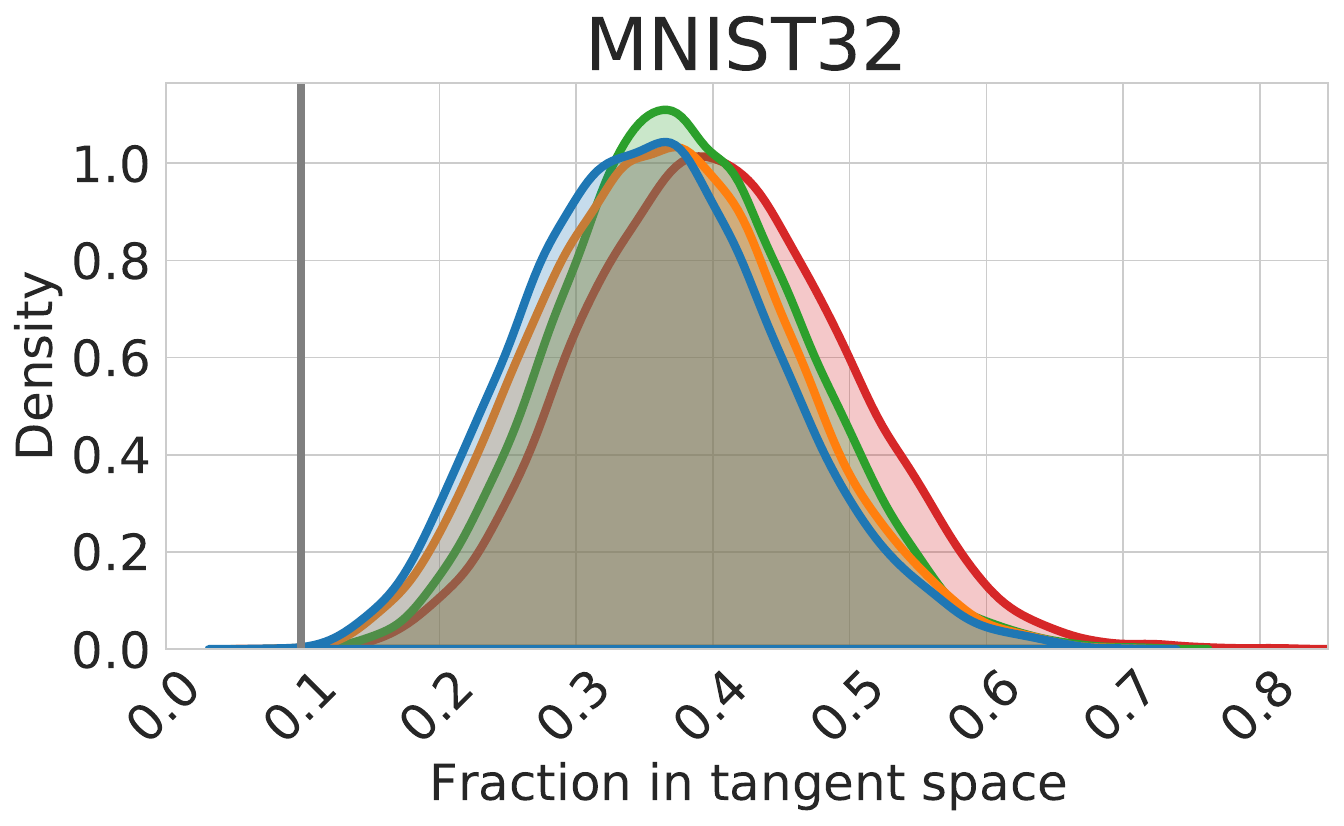}
     \end{subfigure}
     \hfill     \begin{subfigure}{0.32\linewidth}
         \centering
            \includegraphics[width=\linewidth]{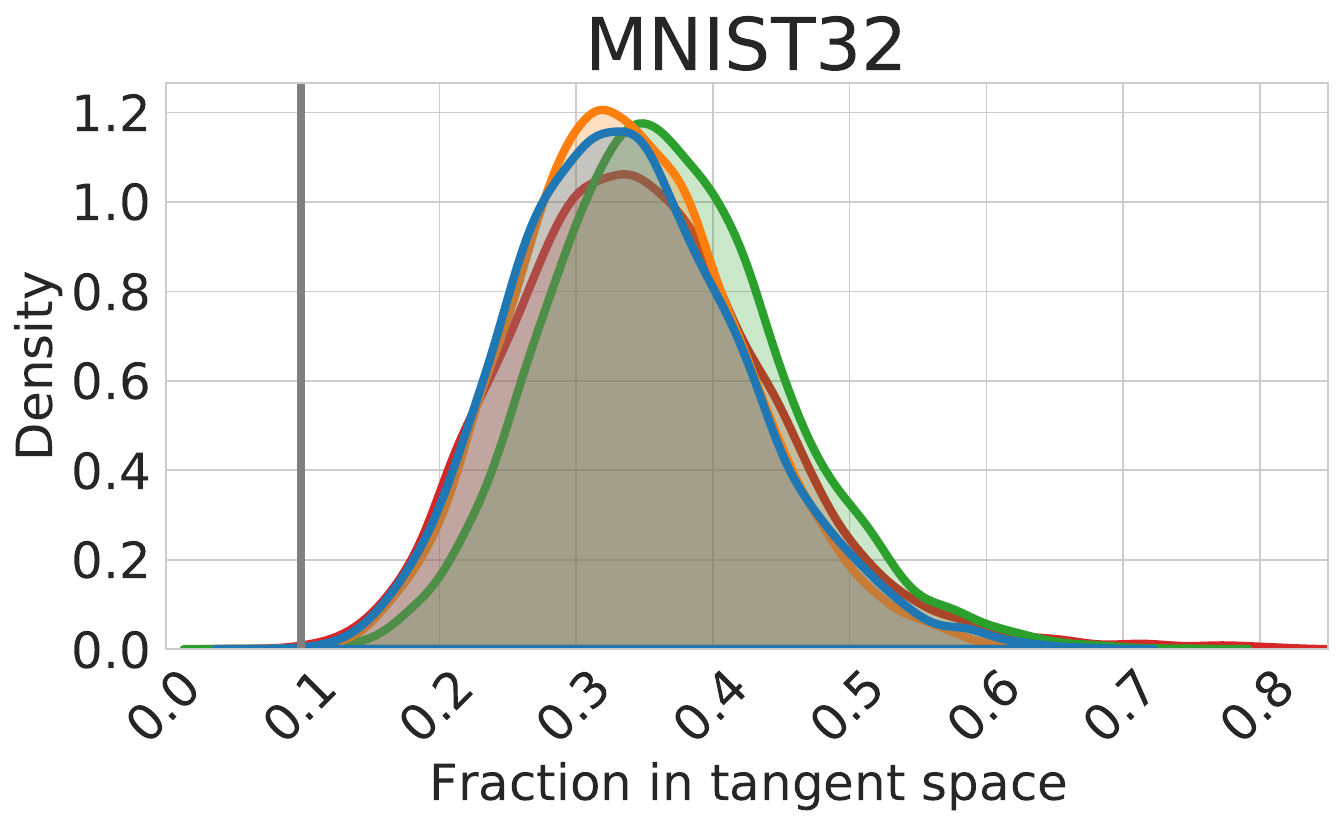}
     \end{subfigure}
     \hfill     
    \caption{Replication of Figure \ref{fig:methods_fractions} in the main paper. Figure shows the result of replicating the entire process that led to Figure \ref{fig:methods_fractions} 10 times (compare Appendix Section \ref{apx:mnist_construction}). First, we re-trained the autoencoder with a different random seed. Then we trained a new labelling function and sampled a new dataset. Finally, we trained a model on the new dataset, computed the different feature attributions and projected them into the tangent space. Except for the last replication where not method seems to improve upon the gradient, the relative ordering between the different feature attribution methods is the same as in Figure \ref{fig:methods_fractions} in the main paper. Note that we would not expect the replications to look exactly same: When we re-train the autoencoder with a different random seed, the generated samples follow a different distribution. Similarly, Figure \ref{fig:training_evolution} in the main paper shows that re-training the image classifier with a different random seed slightly effects the fraction of model gradients in tangent space.}
    \label{fig:apx_mnist_replication}
\end{figure}

\clearpage

\begin{figure}[t]
  \begin{center}
    \includegraphics[width=0.3\textwidth]{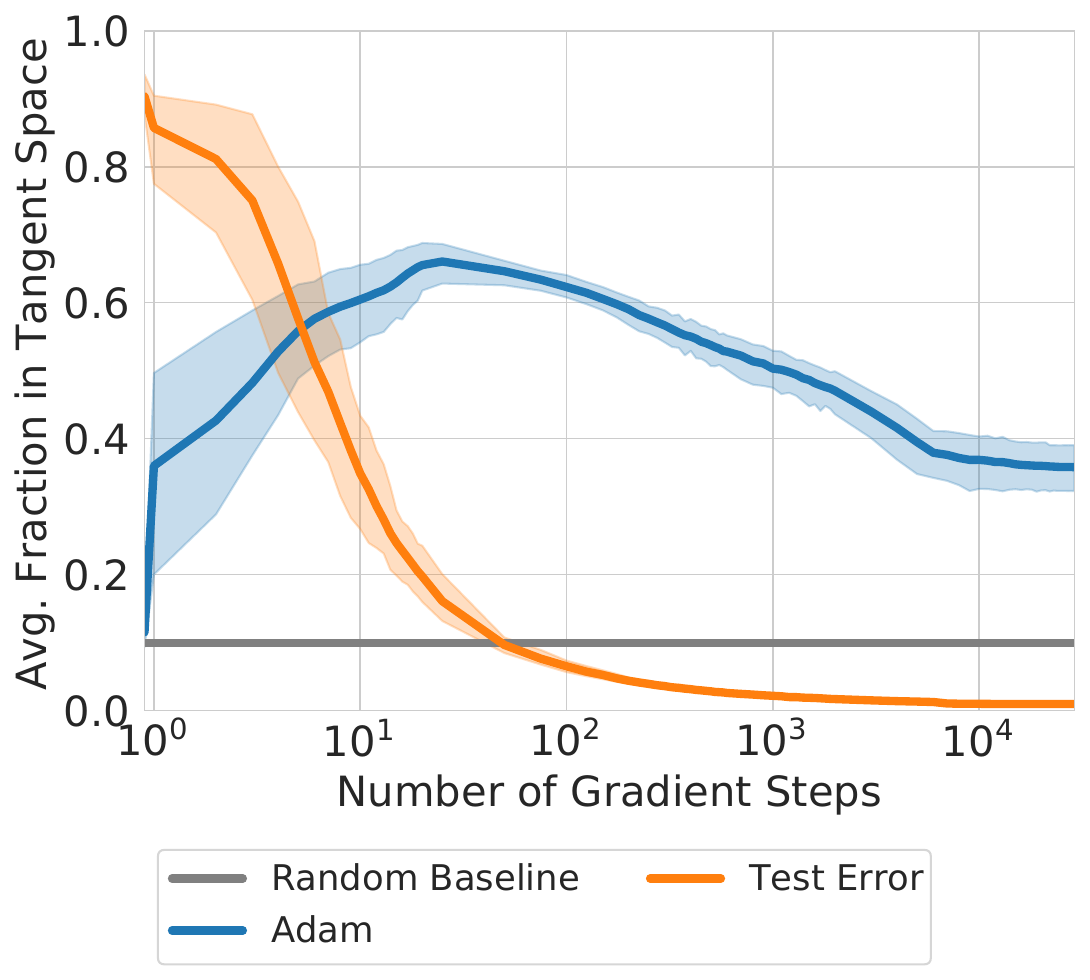}
    \hfill
    \includegraphics[width=0.3\textwidth]{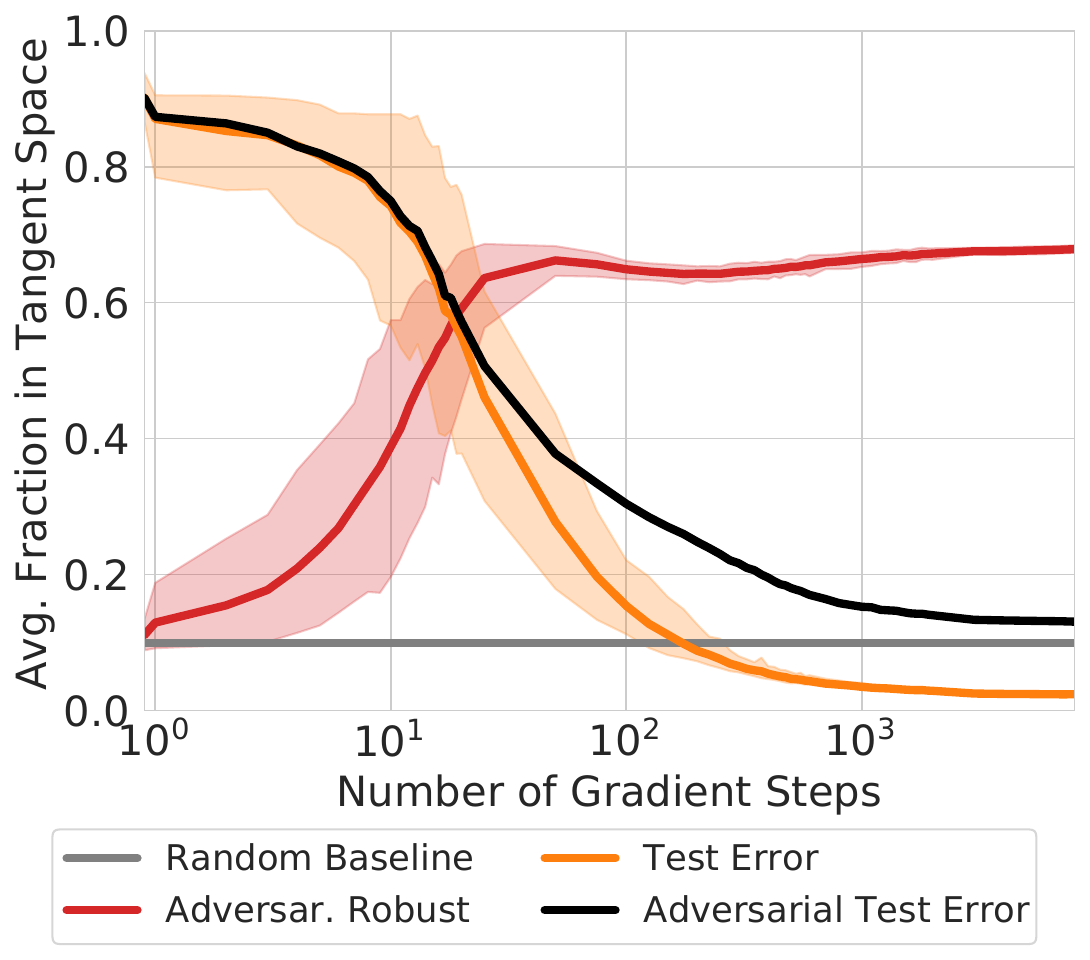}
    \hfill
    \includegraphics[width=0.3\textwidth]{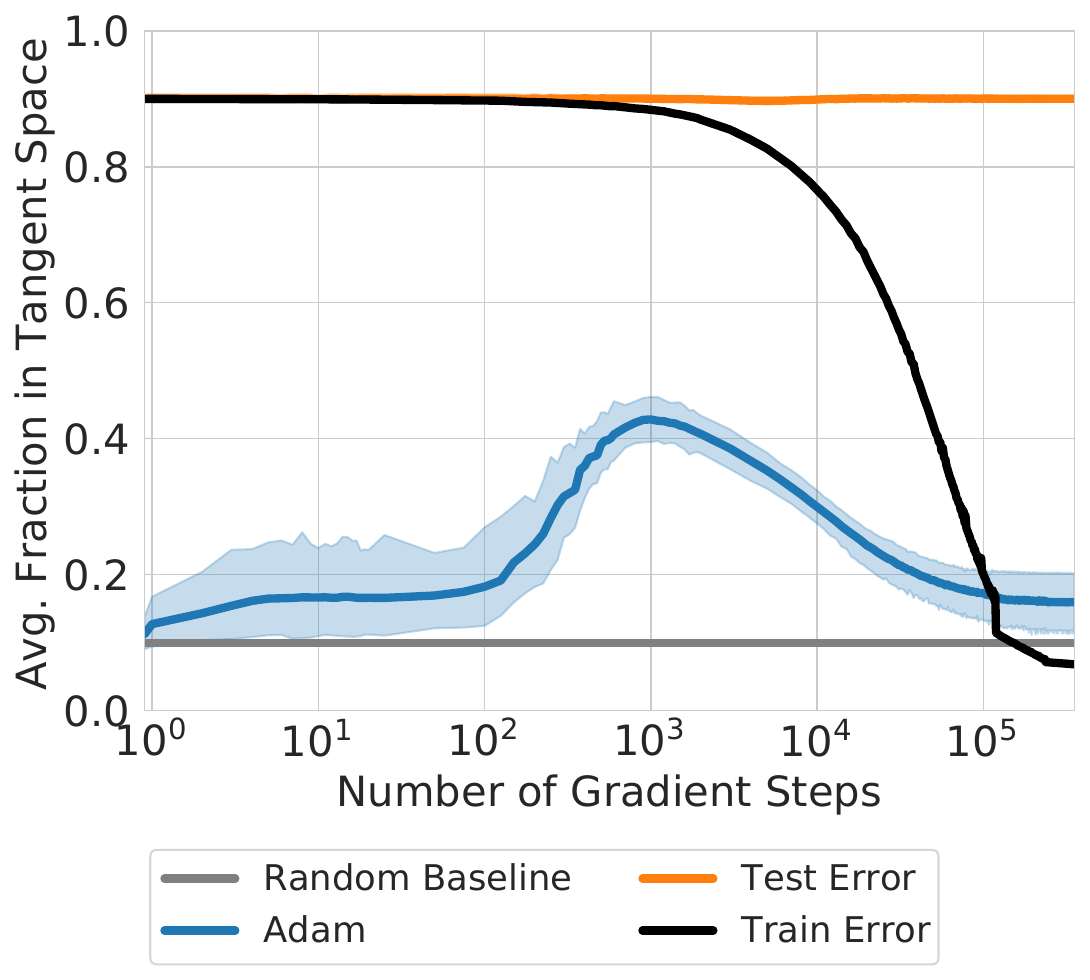}
  \end{center}
  \caption{
  Fraction of gradient in tangent space evolving over the course of training. Mean and 90\% confidence bounds. Gray line shows the expected fraction of a random vector in tangent space. (Left) Training with Adam. (Center) PGD Adv. Robust Training. (Right) Training with Adam and random labels.}
  \label{fig:supplement_evolution}
\end{figure}

\begin{figure}
  \begin{center}
    \includegraphics[width=\linewidth]{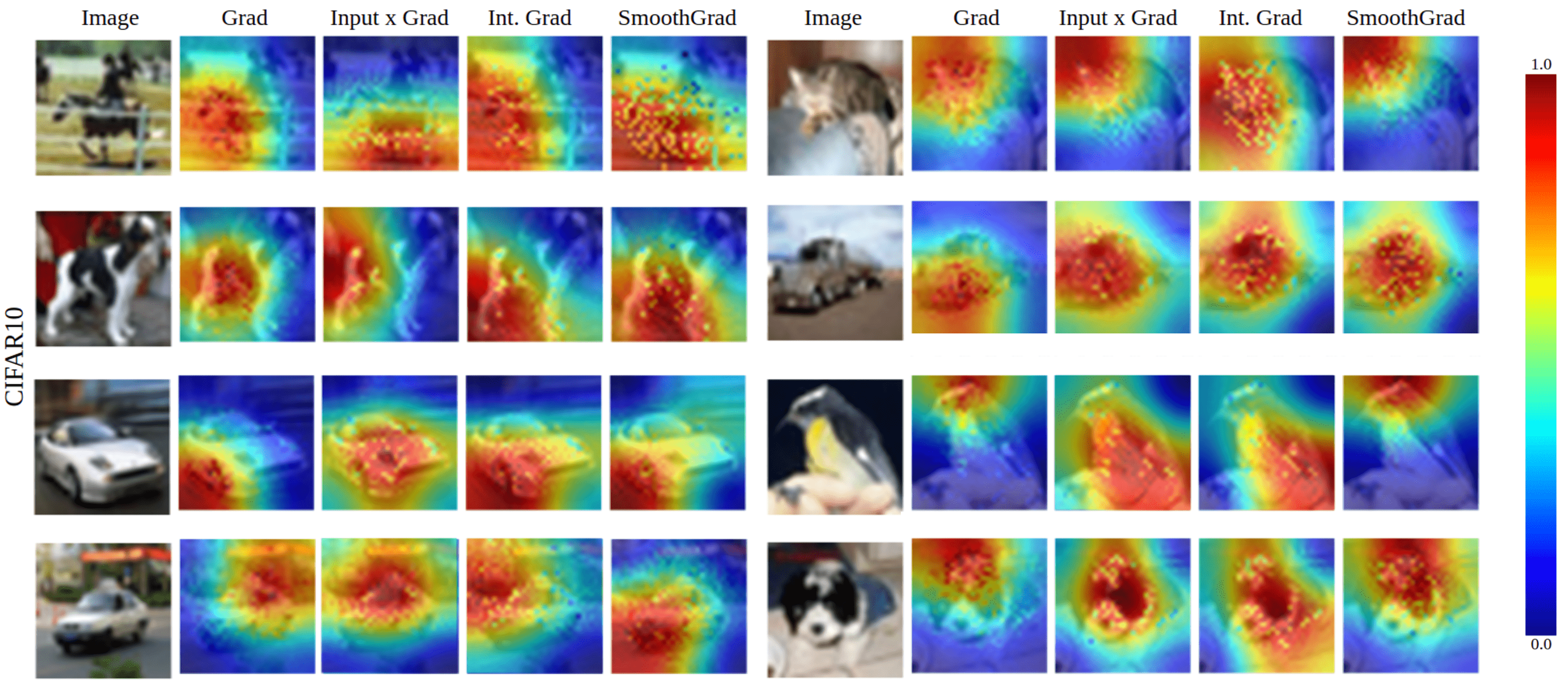}
  \end{center}
  \caption{Additional examples from the CIFAR10 dataset.}
\end{figure}

\begin{figure}
  \begin{center}
    \includegraphics[width=\linewidth]{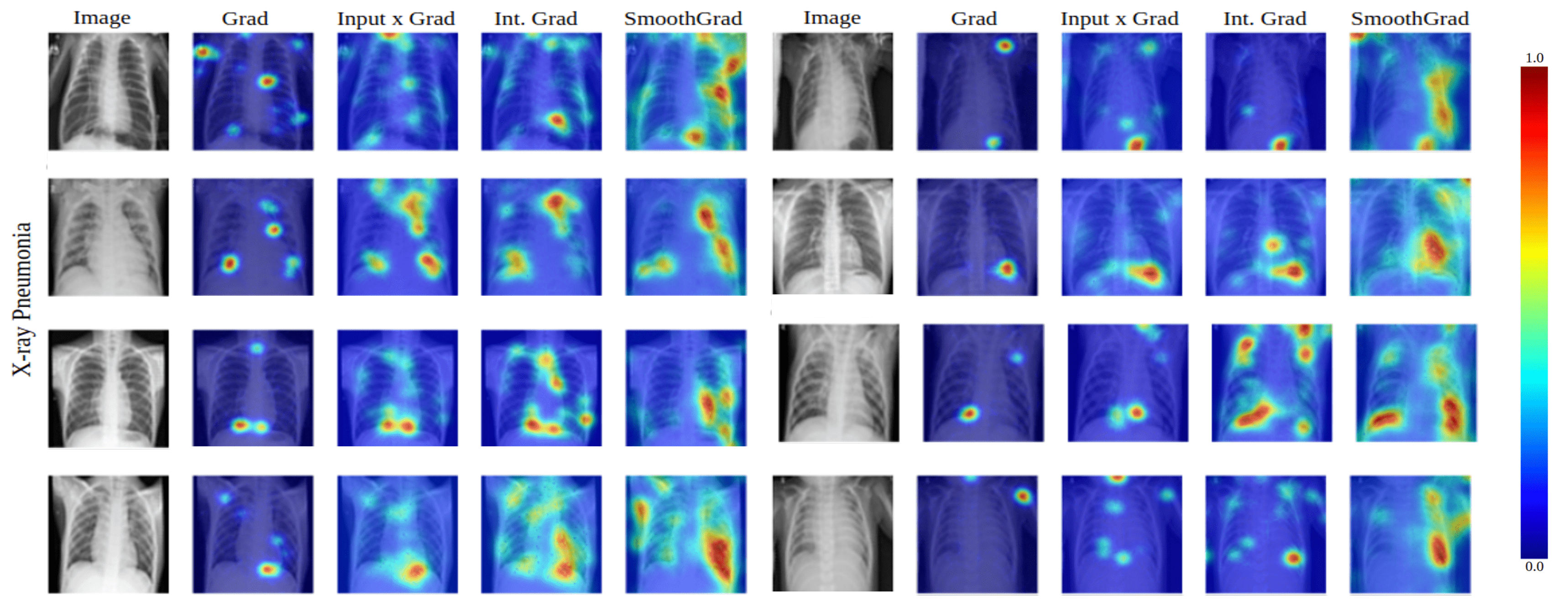}
  \end{center}
  \caption{Additional examples from the X-Ray Pneumonia dataset.}
\end{figure}

\clearpage
\newpage

\begin{figure}[ht!]
  \begin{center}
    \includegraphics[width=\linewidth]{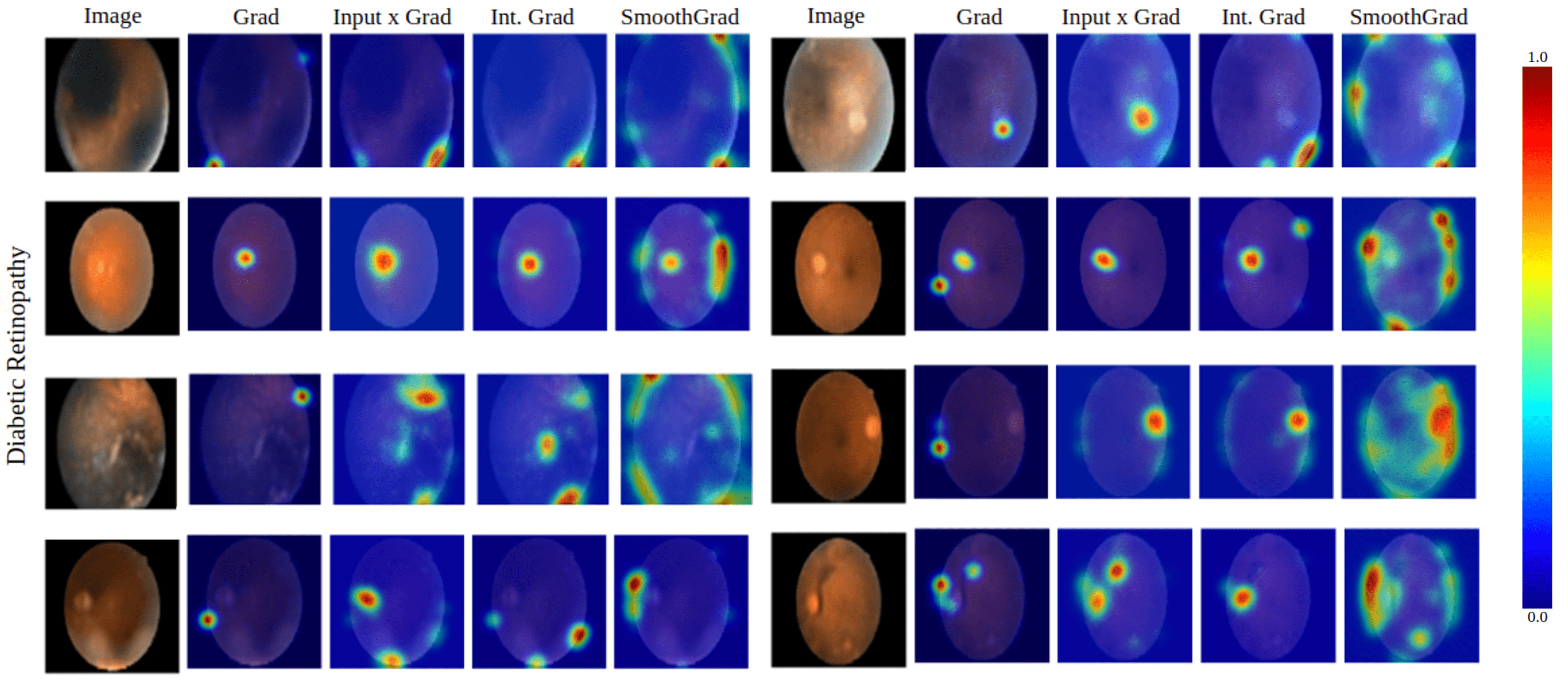}
  \end{center}
  \caption{Additional examples from the Diabetic Retinopathy dataset.}
\end{figure}

\clearpage
\newpage

\newpage
\section{User Study Documentation}
\label{apx:user_study}

\subsection{Outline}

The study consisted of 3 tasks. Each participant completed all 3 tasks. The CIFAR 1 task consisted of 10 decisions per participant, randomly selected from a pool of 15 cases. The MNIST 1 task consisted of 20 decisions per participant, randomly selected from a pool of 100 cases. The MNIST 2 task consisted of 20 decisions per participant, randomly selected from a pool of 200 cases. See the screenshots below for the way in which we introduced the respective tasks to the participants. 

\subsection{Participants}

The participants of the user study were contacted via email. We contacted students, most of whom did not have any prior experience with machine learning. Participation in the study was voluntary and anonymous (see screenshot of consent form below). The entire study took less than 5 minutes and the participants were not compensated. Overall, 30 students participated in the study. Because the study was voluntary and anonymous, we did not collect any personal data and there were no foreseeable risks to the participants, we did not apply for ethics approval. 

\subsection{Selection of images}

The images presented in the study were randomly sampled from the test sets of the respective datasets.%

\subsection{Screenshots}

\begin{figure*}[h]
    \centering
    \includegraphics[width=\textwidth]{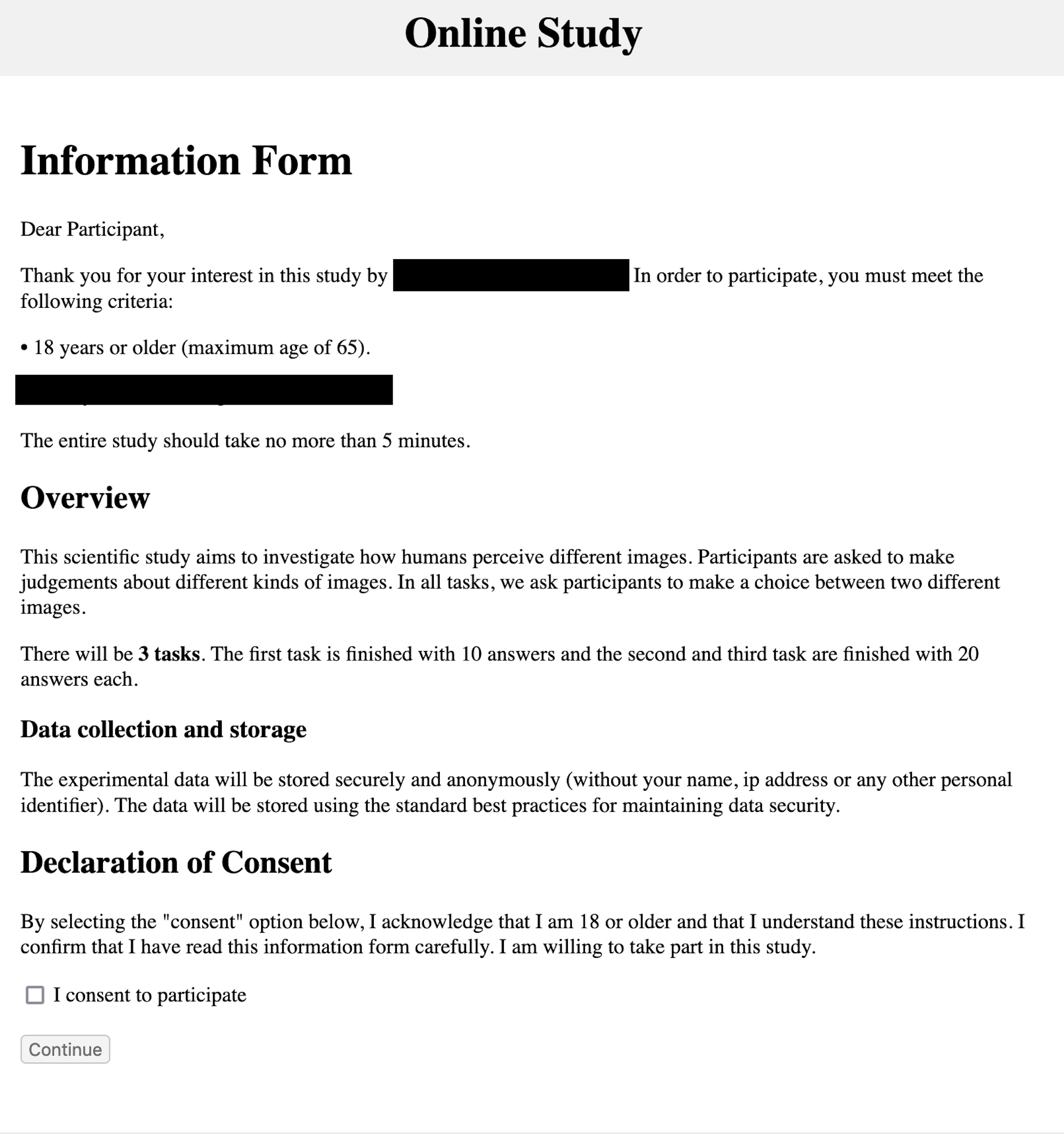}
    \caption{Information form and consent.}
\end{figure*}

\newpage
\begin{figure}
    \centering
    \includegraphics[width=0.9\textwidth]{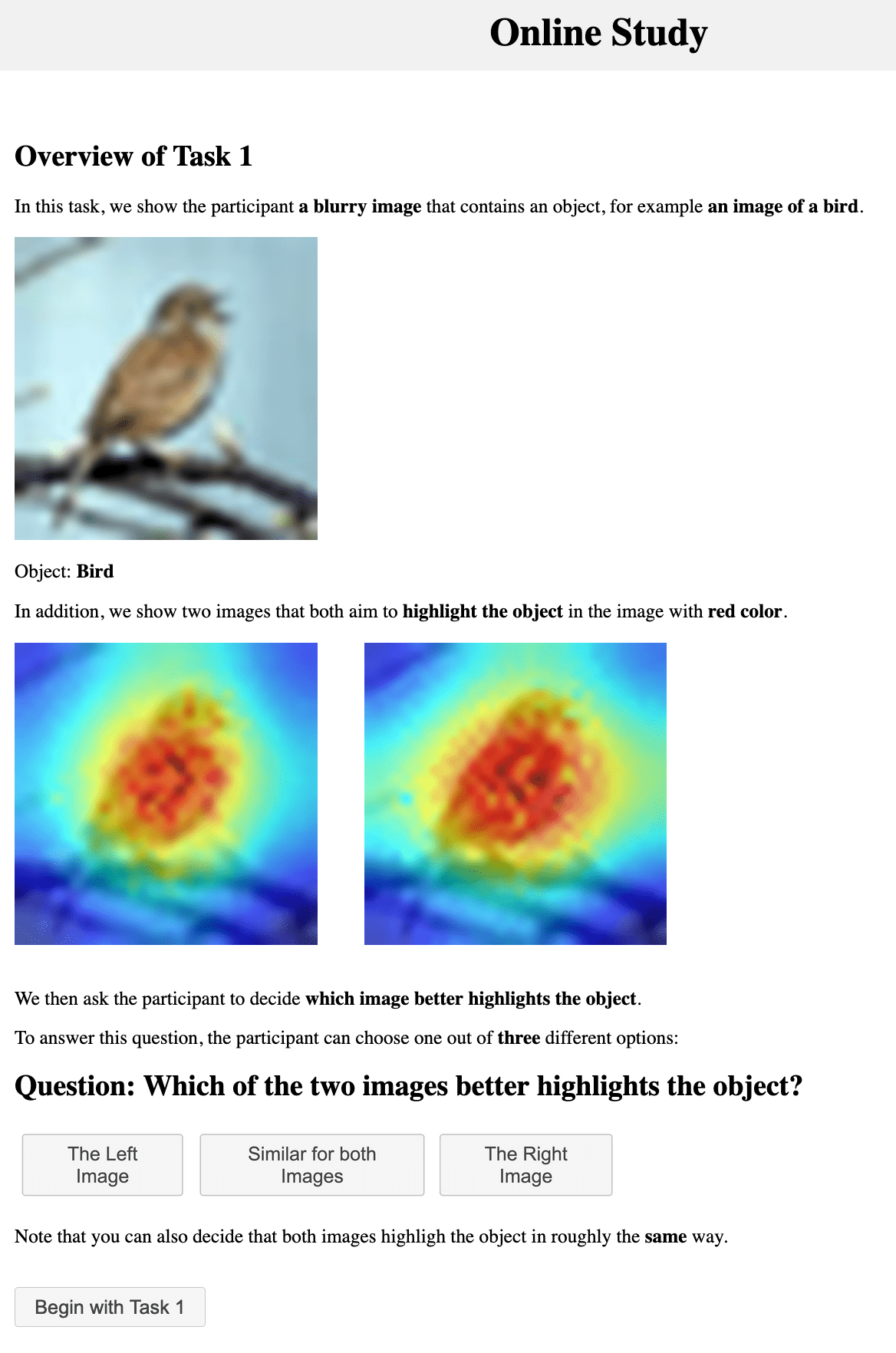}
    \caption{Description of the CIFAR task.}
\end{figure}

\newpage
\begin{figure}
    \centering
    \includegraphics[width=\textwidth]{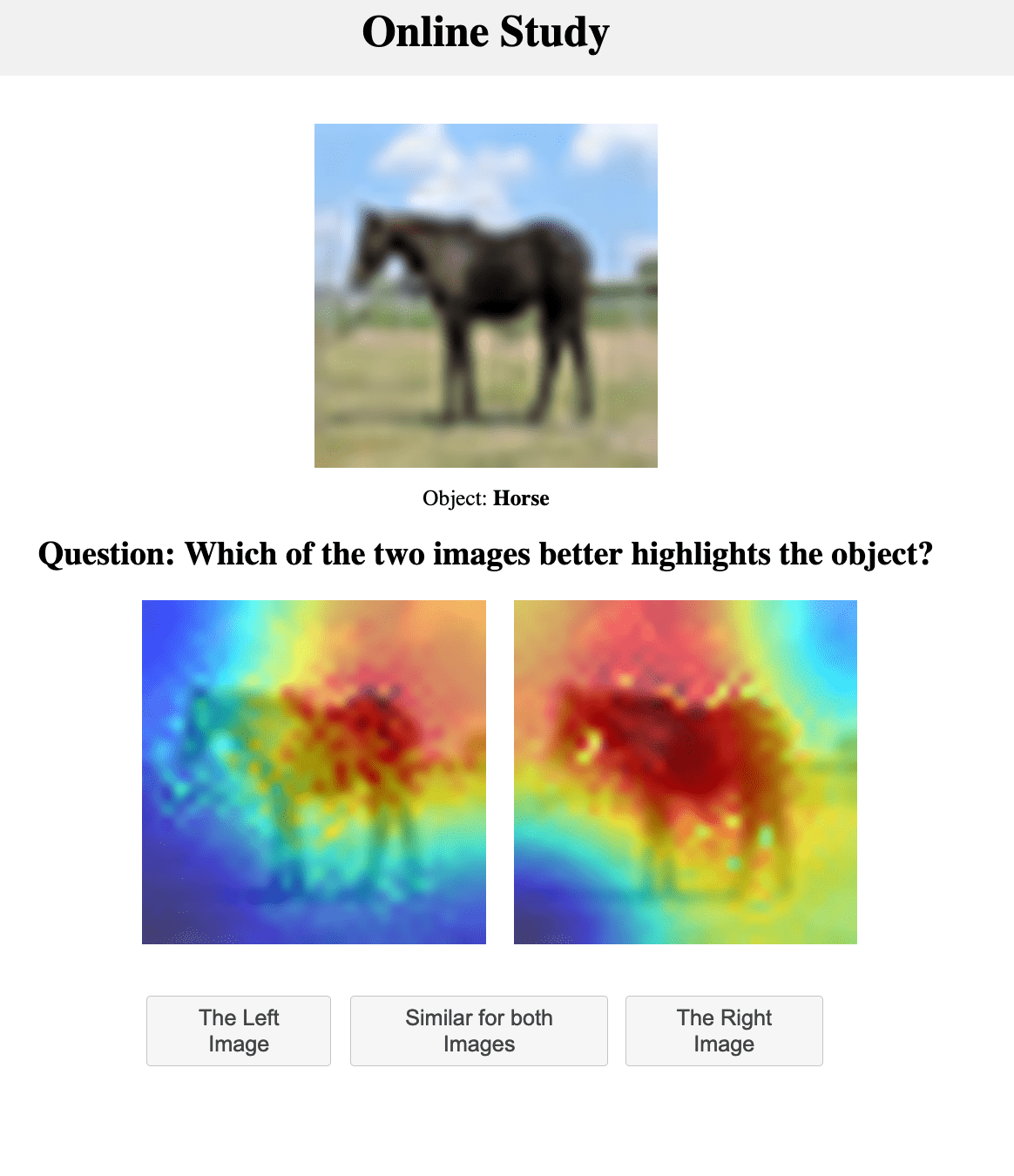}
    \caption{An example from the CIFAR task. Each participant completed 10 instances of this task.}
\end{figure}

\newpage
\begin{figure}
    \centering
    \includegraphics[width=\textwidth]{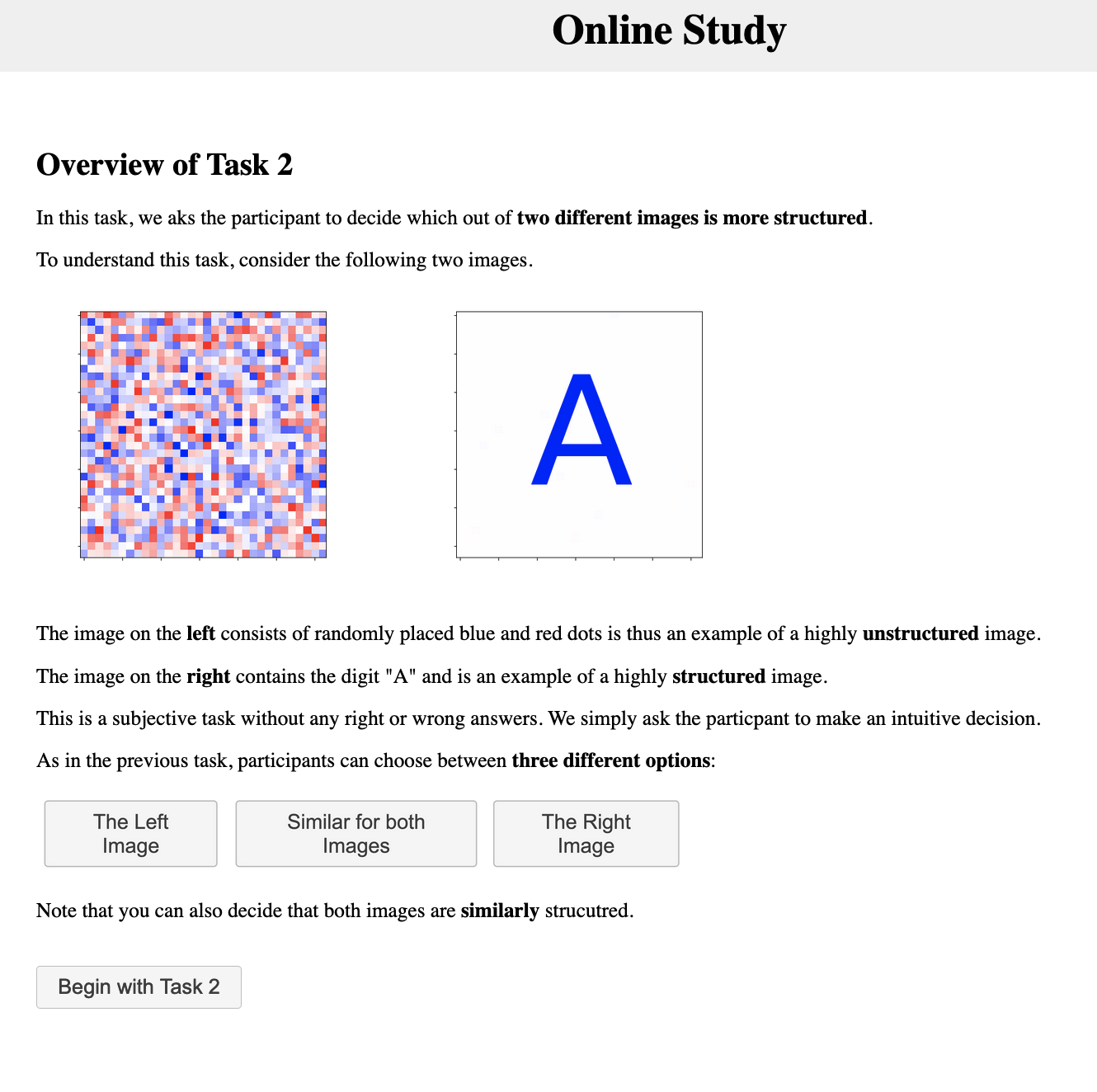}
    \caption{Description of the MNIST tasks.}
\end{figure}

\newpage
\begin{figure}
    \centering
    \includegraphics[width=\textwidth]{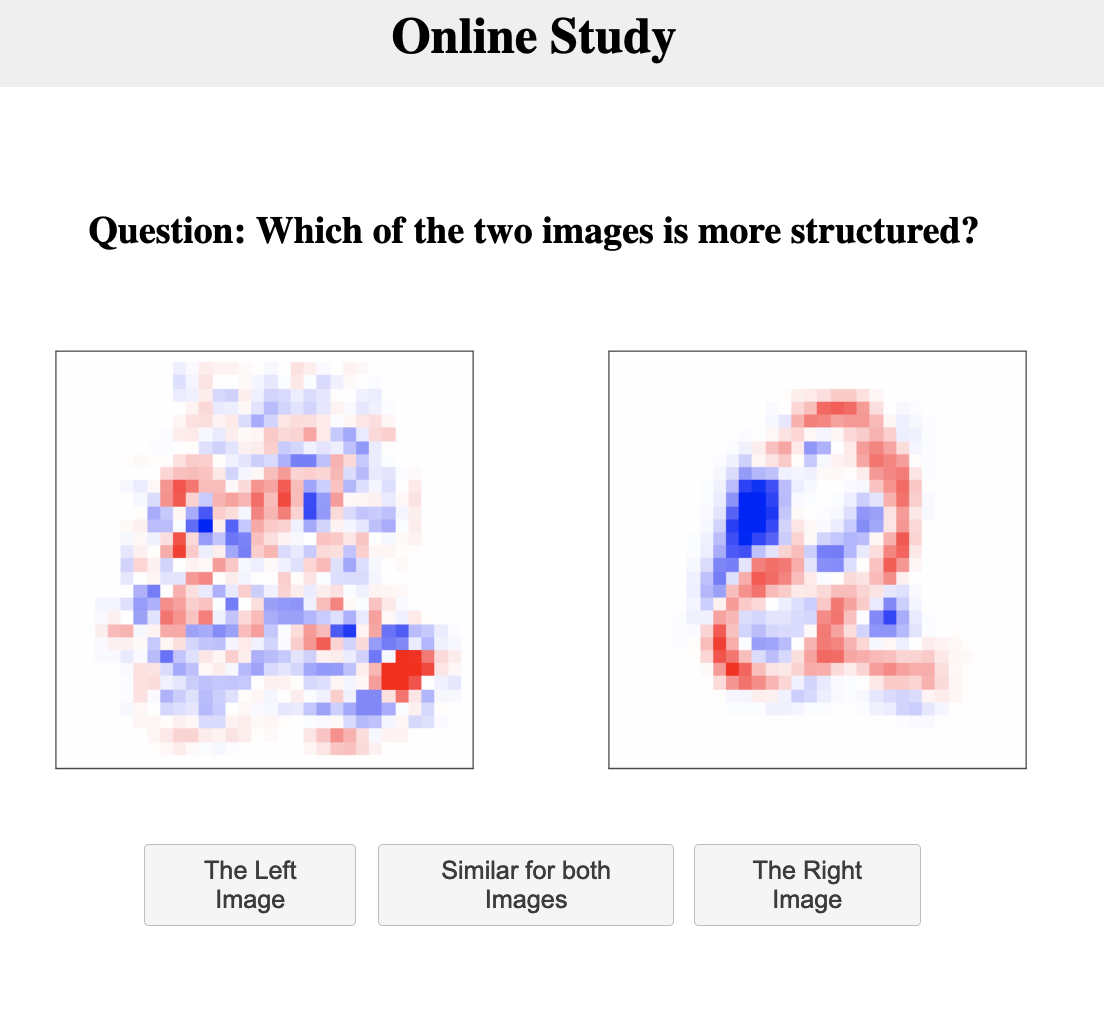}
    \caption{A example from the first MNIST task. Each participant completed 20 instances of this task.}
\end{figure}

\newpage
\begin{figure}
    \centering
    \includegraphics[width=\textwidth]{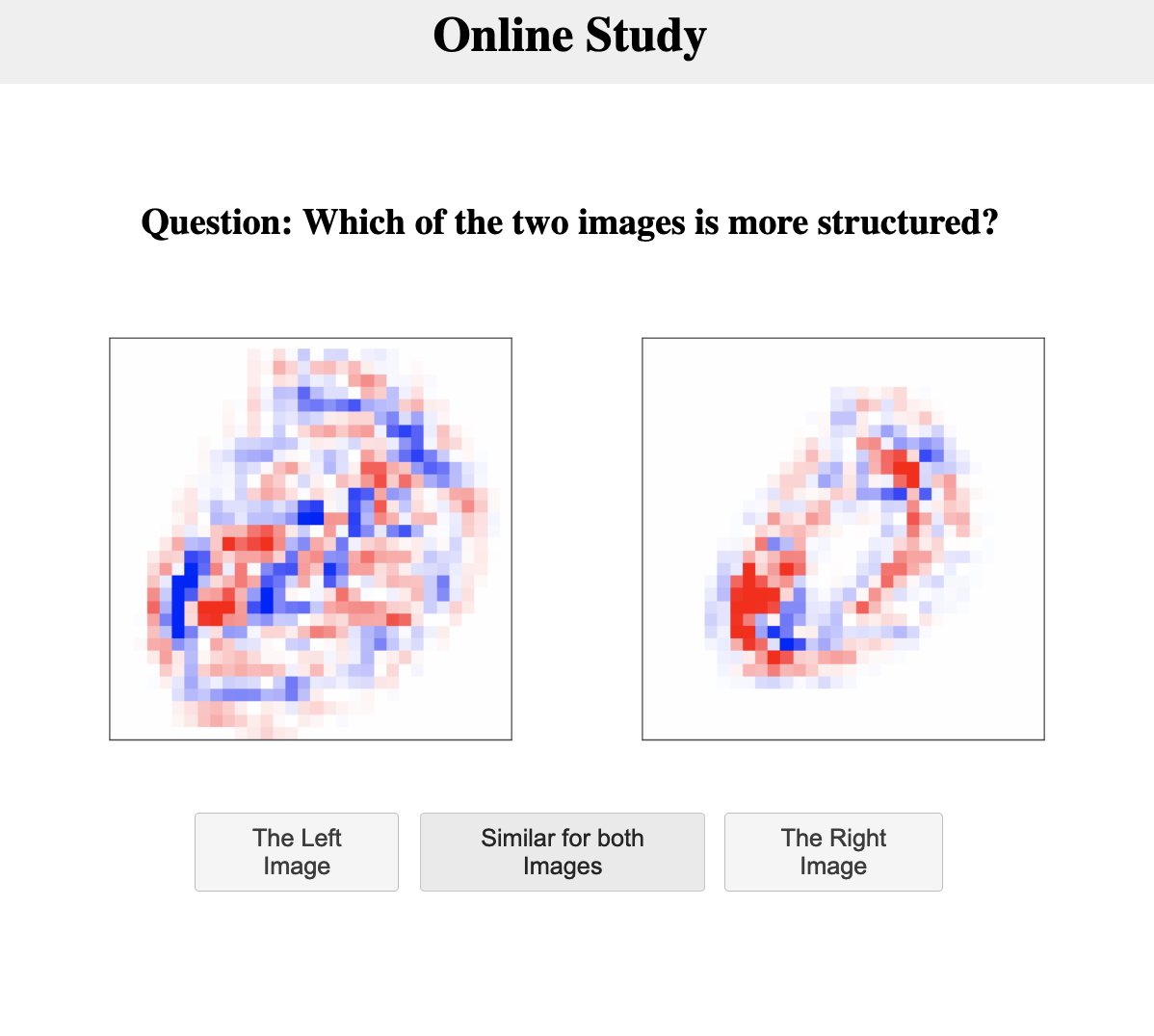}
    \caption{An example from the second MNIST task. Each participant completed 20 instances of this task.}
    \label{fig:my_label}
\end{figure}

\end{document}